\pdfoutput=1
\documentclass[11pt]{article}

\usepackage{EMNLP2022}

\usepackage{times}
\usepackage{latexsym}

\usepackage[T1]{fontenc}
\usepackage[utf8]{inputenc}

\usepackage{microtype}

\usepackage{inconsolata}
\usepackage{booktabs}
\usepackage{graphicx}
\usepackage{tabularx}
\usepackage{xtab}
\usepackage{afterpage}
\usepackage{placeins}
\usepackage{flafter}
\usepackage{longtable}
\usepackage[font=small,labelfont=bf]{caption}
\usepackage[caption=false]{subfig}
\usepackage{enumitem}
\usepackage{soul}

\newcommand\ourslong{Recursive Reprompting and Revision}
\newcommand\ourslongbold{\textbf{Re}cursive \textbf{Re}prompting and \textbf{Re}vision}
\newcommand\oursabstract{Re$^3$}
\newcommand\ours{\textsc{re$^3$}}
\newcommand\oursshortlength{\textsc{re$^3$-short}}
\newcommand\ourslonglength{\textsc{re$^3$-long}}

\newcommand\rolling{\textsc{rolling}}
\newcommand\rollingfinetune{\textsc{rolling-ft}}
\newcommand\noplanner{\textsc{draft-rewrite-edit}}

\newcommand\norerank{\textsc{plan-draft-edit}}
\newcommand\noeditor{\textsc{plan-draft-rewrite}}
\newcommand\entailment{\textsc{entailment}}
\newcommand\entailmentdpr{\textsc{entailment-dpr}}
\newcommand\structured{\textsc{structured-detect}}
\newcommand\plan{Plan}
\newcommand\draft{Draft}
\newcommand\rewrite{Rewrite}
\newcommand\edit{Edit}

\usepackage{todonotes}

\title{\oursabstract{}: Generating Longer Stories With Recursive Reprompting and Revision}

  \author{{\bf Kevin Yang}$^{1}$\ \ \ \ 
    {\bf Yuandong Tian}$^2$\ \ \ \ 
  {\bf Nanyun Peng}$^3$\ \ \ \ 
  {\bf Dan Klein}$^1$\\
    $^1$UC Berkeley, $^2$Meta AI, $^3$UCLA \\
     \texttt{\{yangk,klein\}@berkeley.edu,yuandong@meta.com,violetpeng@cs.ucla.edu}
     }

\begin{document}
\maketitle
\begin{abstract}
  We consider the problem of automatically generating longer stories of over two thousand words. Compared to prior work on shorter stories, long-range plot coherence and relevance are more central challenges here. 
  %Compared to prior settings which focus on much shorter stories, in our task, maintaining overall plot coherence as well as relevance to an initial premise becomes a major challenge. 
  We propose the \ourslong{} framework (\oursabstract{}) to address these challenges by (a)
  %by repeatedly re-prompting a general-purpose language model, with a carefully constructed prompt which includes information from both a high-level plan as well as the current state of the story. 
  prompting a general-purpose language model to construct a structured overarching plan, and (b) generating story passages by repeatedly injecting contextual information from both the plan and current story state into a language model prompt.
%   (\em{a}) generating a structured overarching plan and (\em{b}) repeatedly reprompting a general-purpose language model using a carefully designed prompt involving both the plan and current story state. 
  We then revise by (c) reranking different continuations for plot coherence and premise relevance, and finally (d) editing the best continuation for factual consistency. Compared to similar-length stories generated directly from the same base model, 
  %human evaluators judged our stories as substantially superior in plot coherence (by 14\% absolute gain) and relevance to a given initial premise (by 20\%).% ; moreover, evaluators believed up to 83.3\% of our generated stories to be written by humans. 
  human evaluators judged substantially more of \oursabstract{}'s stories as having a coherent overarching plot (by 14\% absolute increase), and relevant to the given initial premise (by 20\%). 
% TODO something about how we hope this can inspire future work on long-form story generation?
  % Inspired by how humans are taught to write, we propose a plan-write-edit framework for maintaining long-term plot coherence in automated story generation for stories up to X thousand words long. Rather than having a language model generate a story directly, our planning module first generates a story's setup and outline, followed by expanding each outline section to yield a story with a coherent overarching plot. Meanwhile, our editing module reranks outputs for correspondence to the outline in addition to overall plot coherence, and then iteratively improves the selected output if we detect factual discontinuities. Compared to similar-length stories generated directly from GPT3~\cite{brown2020language}, human evaluators judged our stories as substantially superior in plot coherence as well as relevance to a given initial premise; moreover, evaluators believed up to X\% of generated stories to be written by humans. 
%   \yuandong{We should mention ``with the help of large pre-trained language models'' somewhere in the abstract.}
\end{abstract}

\section{Introduction}

\begin{figure}[t!]
\centering
\includegraphics[width=0.98\linewidth]{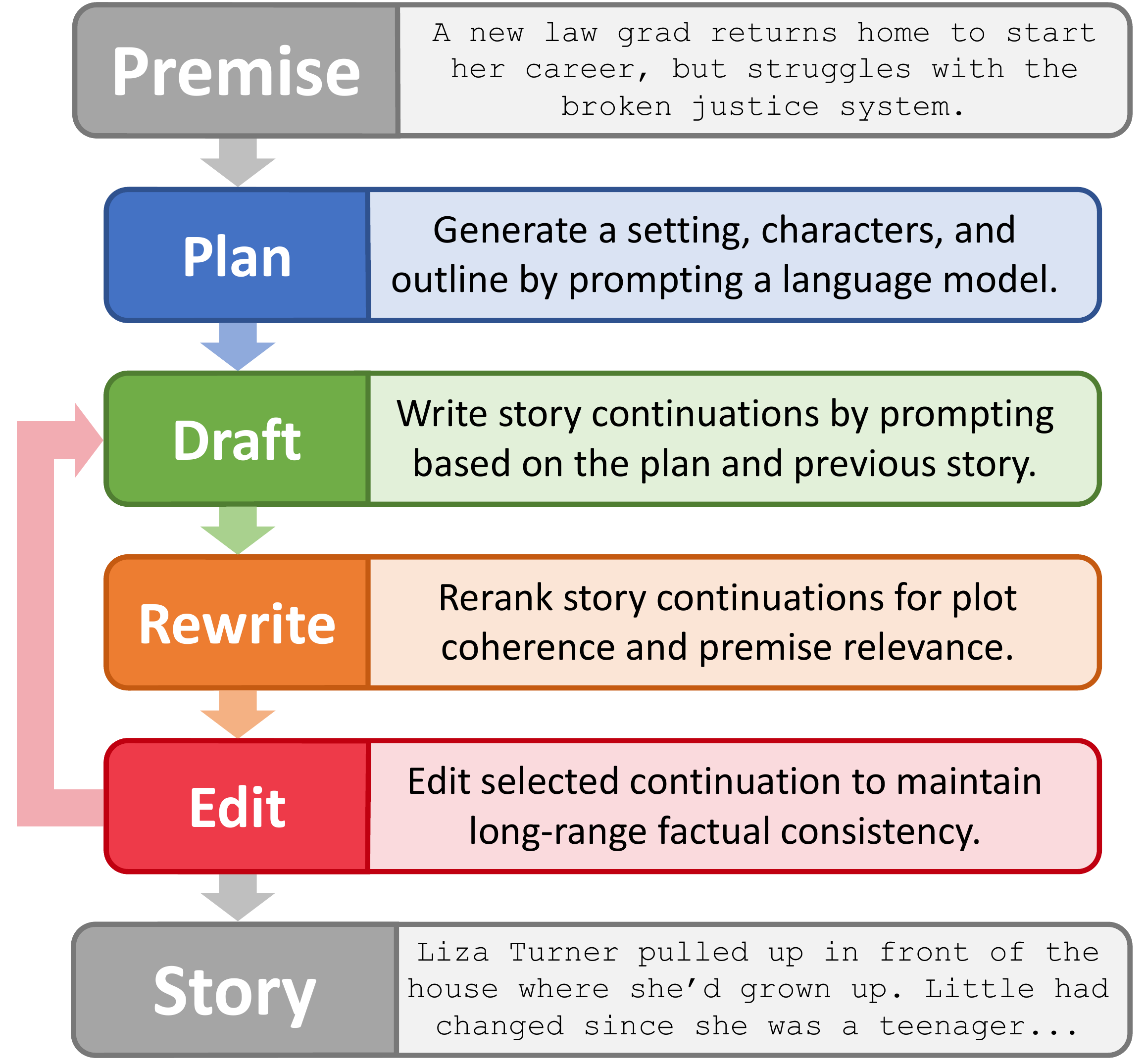}
\caption{High-level overview of \oursabstract{}.}
\label{fig:main}
\vspace{-1em}
\end{figure}

% \begin{figure*}[t!]
%     \centering
    
%     \includegraphics[width=1.0\textwidth]{EMNLP 2022/figures/method12.pdf}
%     \caption{\small Illustration of the \oursabstract{} framework. Italics indicate prompt components and boxes indicate output text or data structures. \textbf{(a) \plan{}:} Given a premise, the \plan{} module prompts a language model for a setting, character descriptions, and outline. Each part is conditioned on all previous parts. \textbf{(b) \draft{}:} To generate each next passage, our recursive reprompting procedure selects relevant parts of the plan from (a) in addition to parts of the previously generated story. The different components are concatenated in the order shown, and then used to prompt a language model. \textbf{(c) \rewrite{}:} Generated story continuations are reranked by coherence and outline relevance. \textbf{(d) \edit{}:} For each character, the \edit{} module infers natural language facts and turns them into attribute-value pairs. New values (blue) are added, while contradictory values (red) are corrected.
%     %\yuandong{Maybe it is better to show concrete examples of the prompt sent to GPT-3, and closely labels where each part comes from and which part uses story dataset? Right now I am a bit confused e.g., how the outline is constructed from the given premise and how many GPT-3 API calls are needed.}
%     }
%     \vspace{-0.5em}
%     \label{fig:method}
% \end{figure*}

Generating long-term coherent stories is a longstanding challenge for artificial intelligence, requiring a comprehensive grasp of linguistic, world, and commonsense knowledge~\cite{charniak1972toward,turner1994creative}. Recently, many works have automatically generated short stories ranging in length from five sentences to one or two paragraphs~\cite{fan2018hierarchical,yao2019plan,goldfarb2020content,rashkin2020plotmachines,han2022go}. While stories of such length serve as a good test bed for text generation, they are much shorter than typical short stories meant for human consumption, which are often several pages in length.

% However, humans may envision a short story to be several pages in length, as is typically the case when human authors publish short stories.

% However, this definition of short stories contrasts with the typical human notion of a short story

% However, the typical human notion of a short story can be several pages long

% However, when human authors publish short stories, 
% a single story can be several pages long. \violet{I wonder can we add more motivation than simply say that humans can write longer stories. There are many things humans do but AI may not need to do. Can we say something about necessity of generating longer stories.} %Thus there is a disconnect between prior machine-written short stories compared to the typical human definition. %\yuandong{I think the consistency is the key. The discrepancy in definition of ``short story'' may be a secondary effect.}

In this work, we aim to bridge some of this gap by generating much longer ``short'' stories: 
the final generated stories in our experiments are 2000-2500 words. We are the first to automatically generate plot-coherent stories of such length, with further length increases limited primarily by evaluation rather than technical issues.\footnote{We generate a 7500-word story in Appendix \ref{appendix:longer_story}.} 
Generating stories of such length faces qualitatively new challenges compared to prior work on shorter stories.
First, the system must maintain a coherent overarching plot over thousands of words. 
Given an initial premise, it should maintain relevance to this premise over thousands of words as well.
%The system should also be controllable to some degree; for example, it should follow an initial premise if given. 
Additional challenges
include preservation of narration style and avoiding factual contradictions over a very long horizon. 

Of course, recent years have also witnessed a dramatic rise in the capabilities of general-purpose (non-finetuned) large pretrained language models. Of particular note are their strong zero-shot capabilities, especially when given clever prompts~\cite{brown2020language,kojima2022large}. Yet despite recent improvements, even the best models to date may still struggle with complex long-form generation, such as in our story generation task (Section \ref{sec:evaluation}). %\yuandong{Could we generate one consistent story of 10000 words long and show in arXiv/submission? This can be setup as a regular NLP challenge later. Not too long for human to read, and not too short so that regular language models can cheat.} 
% Generating stories TODO major problems and why harder than prior work. coherence, and relatedly staying on topic for outline point or initial premise. 

% TODO discuss the main structure of our method, intuition with human writing process
In contrast, human writers successfully navigate the myriad challenges of long-form generation on a regular basis.
%Following prior human-in-the-loop approaches~\cite{goldfarb2019plan,coenen2021wordcraft,lee2022coauthor}, \violet{I'll probably move this statement until after we explain our approach. The current way sounds unnecessarily incremental. We can maybe introduce our process, and say that while there are prior works also explore this direction, we are different because of x, y, z.} 
We observe that a human writer does not simply write a long document in one shot. Rather, he or she may (a) create a detailed plan, then (b) draft each next passage of the document according to that plan. He or she may then revise by (c) rewriting passages entirely, and/or (d) post-editing for finer details.% such as long-range factual inconsistencies. %For long stories with many characters, the writer may even (\em{f}) keep a summary for the traits and personality of characters, their roles in the story and interactions so far.  
%first creates a plan or outline before proceeding with the main text, meanwhile rewriting and editing as needed. 

Motivated by this observation, we propose the \ourslongbold{} framework (\oursabstract{}, Figure \ref{fig:main}) to generate longer stories. While based on the human writing process, \oursabstract{} is a fully automatic system with no human intervention, unlike prior approaches which model the human writing process with a human in the loop~\cite{goldfarb2019plan,coenen2021wordcraft,lee2022coauthor}.
%In this work, we generate longer stories by mimicking the human writing process via a fully automatic system with no human intervention. We propose the \ourslongbold{} framework (\oursabstract{}), shown in Fig.\ref{fig:method}. 
First, (a) \oursabstract{}'s \plan{} module generates a plan by prompting GPT3~\cite{brown2020language} to augment a given premise with a setting, characters, and outline.
(b) \oursabstract{}'s \draft{} module then generates each next story continuation by \textit{recursively reprompting} GPT3 using a strategically crafted prompt, in a procedure which can be viewed as a generalization of chain-of-thought prompting~\cite{kojima2022large}. %based on both the initial plan and the story thus far. 
%This prompted generation is a subroutine for generating limited-length continuations: at each next step 
Specifically, our prompt is dynamically reconstructed at each step by selectively manifesting contextually relevant information from the initial plan---itself generated by prompting---and the story thus far. We then divide the revision process into (c) a
\rewrite{} module which emulates a full rewrite by reranking alternate continuations, and (d) an \edit{} module which makes smaller local edits to improve factual consistency with previous passages. 
% In addition to drafting, \oursabstract{} incorporates a revision system beginning with (\em{c}) our \rewrite{} module, which selects the best of several candidate
% continuations at each step using an ensemble of rerankers, emulating the rewriting process. 
% (d) Finally, \oursabstract{}'s \edit{} module makes smaller local edits if it detects
% factual inconsistencies with previous passages. %We emphasize that \oursabstract{} operates fully automatically, without a human in the loop as in some prior approaches~\cite{coenen2021wordcraft,lee2022coauthor}. %\yuandong{We can put Fig 1 here. Also we want to make sure it is clear which part of the process is done by human, and which parts are done manually. It is in Fig 1 but we should also mention in the main text.}

% Compared to prior works which model parts of the human writing process, 
As an additional contribution, our \plan{} and \draft{} modules are fully zero-shot rather than trained on %\violet{I added this because arguably, the pre-training could contrain existing story datasets.}
existing story datasets. %, leveraging the strong controllability of large language models via prompting. 
Thus not only does \oursabstract{} generate stories an order of magnitude longer than those of prior work, but it is not limited to any particular training domain.% \violet{I removed the controllable claim as it's too brief to understand what do you mean, plus we don't seem to have experimental results on this.} %Moreover, \oursabstract{} is itself potentially highly controllable. 

To evaluate \oursabstract{} for longer story generation, we compare its generated stories to similar-length stories from two GPT3-based ``rolling-window'' baselines (Section \ref{sec:evaluation}). %, all  on a given story premise.
% two GPT3-based ``rolling-window'' baselines:
% one using GPT3 out-of-the-box, and one finetuned on long stories from the WritingPrompts dataset~\cite{fan2018hierarchical}. 
In pairwise comparisons,
human evaluators rated stories from \oursabstract{} as significantly and substantially more coherent in overarching plot (up to 14\% absolute increase in the fraction deemed coherent), as well as relevant to the initial premise (up to 20\%). In fact, evaluators predicted up to 83\% of stories written by \oursabstract{} to be written by humans. 
% Additionally, human evaluators noted significantly more miscellaneous writing problems (e.g., changes in narration style, factual inconsistencies, highly confusing text)
% in baselines' stories compared to those of \oursabstract{}. 
The results indicate that \oursabstract{} can be
highly effective at improving long-range coherence and premise relevance in longer story generation.\footnote{All code and data available at \url{https://github.com/yangkevin2/emnlp22-re3-story-generation}.}% emphasize more in future work about this strategic prompting?

% Finally, in Sec. \ref{sec:analysis} we conduct an ablation analysis on the individual components of \oursabstract{}. We find that the reprompting (planning and drafting) and reranking components are crucial to overall performance, while our proof-of-concept editing system is useful in a controlled environment but does not significantly affect our main metrics. 

% We conclude by discussing limitations and directions for future exploration. We hope this work can inspire
% future methods using similar strategic-planning-and-revision frameworks for long-form generation, 
% and also spark further effort toward generating coherent long stories and eventually even novels. % TODO bother with the strategic planning revision? or just talk about longer stories? is this too land-grabby?

% TODO contributions section? 
% coherent long stories, emphasize more the strategic prompting stuff?

% TODO discuss experiments + results compared to the baseline

    % - [ ] we propose this plan, write, and edit system for story generation. 
    % - [ ] motivation: prior works generate much shorter stories. we show some of the different set of problems when you use much better models, but try to generate much longer stories. 
    % - [ ] motivation: the rolling window baseline sucks; identify some problems
    % - [ ] identify major problems: coherence to an overarching plot, and consistency of details. maybe coherence + controllability / relevance to premise are more important? since you don't have any real results on factual consistency. 
\section{Related Work}

\textbf{Automatic Story Generation.} Several previous works have modeled parts of our proposed writing process, usually one part at a time. 

Most similar to our \plan{} module are approaches using an outline or structured schema to maintain plot coherence~\cite{li2013story,fan2018hierarchical,yao2019plan,goldfarb2020content,rashkin2020plotmachines,tian2022sonnet}. Other methods for high-level planning include latent variables~\cite{miao2016language,wang2019t,wang2022language}, coarse-to-fine slot-filling~\cite{fan2019strategies}, and keywords and/or control codes~\cite{peng2018towards,ippolito2019unsupervised,xu2020megatron,lin2021plug}. 

Meanwhile, our \rewrite{} module uses rerankers similar to \citet{guan2020knowledge} and \citet{wang2020narrative}, although we model both coherence and premise relevance. \citet{yu2020draft} iteratively edits and improves the output like our \edit{} module, but we additionally \textit{detect} when edits are required.
% Meanwhile, our rerankers for the \rewrite{} module are trained similarly to those of \citet{guan2020knowledge} and \citet{wang2020narrative}, although we train both coherence and relevance rerankers. 
% While revision systems, especially systems like our \edit{} module, are less explored for automatic story generation

We emphasize again the length of stories we aim to generate. In prior studies, out-of-the-box language models struggled to generate even very short stories~\cite{holtzman2019curious,see2019massively}. 
% Leveraging the ROCStories dataset of five-sentence stories~\cite{mostafazadeh2016corpus}. 
Although there exist datasets of relatively longer stories, such as WritingPrompts~\cite{fan2018hierarchical} and STORIUM~\cite{akoury2020storium}, 
many works still only focus on stories of about five sentences~\cite{wang2019t,yao2019plan,qin2019counterfactual,wang2022language}, even when using language models with hundreds of billions of parameters~\cite{xu2020megatron}. %, leveraging the ROCStories dataset of five-sentence stories~\cite{mostafazadeh2016corpus}. 
% Even works using language models with hundreds of billions of parameters focus on few-sentence stories~\cite{xu2020megatron}. 
Some challenges of generating longer stories are apparent in \citet{wang2022language}: their method generates high-quality few-sentence stories, but their forced long text generations, while judged better than baselines', remain confusing and repetitive. 
%Generating multiple-thousand-word stories involves qualitatively new challenges compared to very short stories, particularly 
Moreover, maintaining long-range plot coherence, premise relevance, and factual consistency is substantially harder over multiple-thousand-word horizons.% over passages far longer than the context window of the language model used for generation. 

\medskip
\noindent\textbf{Human-In-The-Loop Story Generation}. In contrast to fully automatic approaches like \oursabstract{}, several recent works have proposed human-interactive methods to maintain quality in longer stories~\cite{coenen2021wordcraft,lee2022coauthor,chung2022talebrush}. Such works commonly combine both planning and revision systems~\cite{goldfarb2019plan,coenen2021wordcraft}. 
%Finally, some recent works have generated longer high-quality stories using large language models with a human in the loop~\cite{goldfarb2019plan,coenen2021wordcraft,lee2022coauthor,chung2022talebrush}. 
In principle, \oursabstract{} is also highly controllable via human interaction, as both our planning and revision systems operate nearly entirely in natural language space; however, we focus on fully automatic generation in this work. 

\medskip
\noindent\textbf{Prompting.} Numerous works have demonstrated general-purpose language models' strong zero-shot ability on a wide variety of tasks via prompting~\cite{brown2020language,zhong2021adapting,sanh2021multitask,ouyang2022training,wu2022autoformalization}. Careful prompt design can yield further gains~\cite{lee2021dialogue,liu2021pre,kojima2022large}.
% even on tasks for which general-purpose models may seem unsuited, such as reasoning ~\cite{}.
However, most prompting methods focus on shorter-answer tasks rather than long-form generation. Instead of generating the output in one shot, our recursive reprompting procedure treats prompting as a \textit{subroutine} to generate the final output in conjunction with our planning and revision infrastructure. Compared to chain-of-thought prompting approaches like \citet{kojima2022large}, \oursabstract{} goes a step further by repeatedly re-composing the prompt in modular fashion, dynamically recombining the most contextually relevant parts of both the high-level plan and the story thus far.

\section{\ourslong{}}

We now describe our \ourslong{} framework (\oursabstract{}), which decomposes the human writing process into our \plan{}, \draft{}, \rewrite{}, and \edit{} modules. See Appendix \ref{appendix:example_steps} for concrete examples of each component in practice. 

\subsection{\plan{} Module}\label{sec:setup_outline}

\begin{figure}[htbp]
\centering
\includegraphics[width=1.0\linewidth]{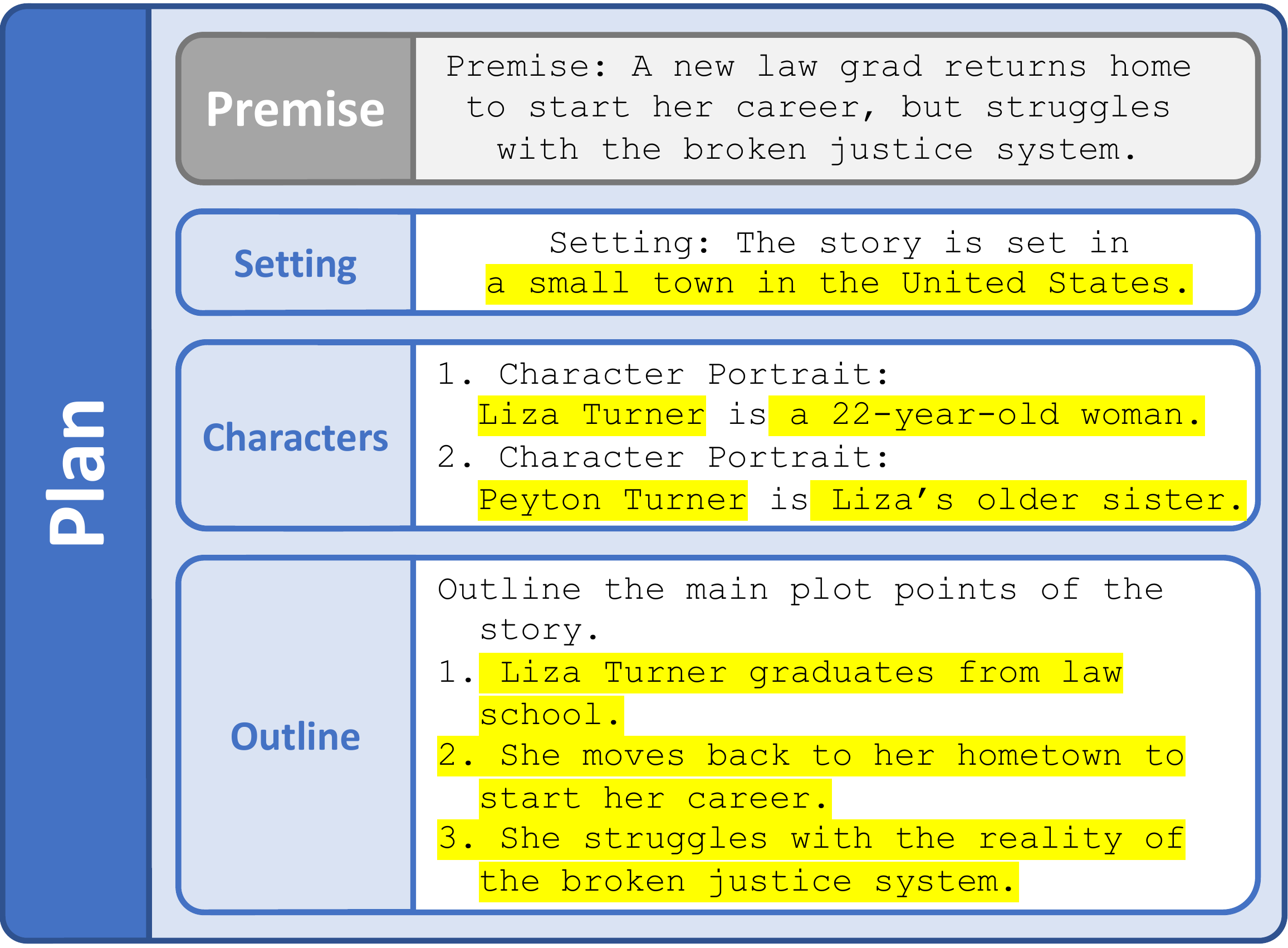}
\caption{Illustration of \oursabstract{}'s \plan{} module, which prompts a language model to generate a setting, characters, and outline based on the premise. Highlighting indicates generated text.}
\label{fig:plan}
\vspace{-0.5em}
\end{figure}

The \plan{} module augments a story premise with a setting, characters, and outline (Figure \ref{fig:plan}). 
% These details are subsequently used in a structured prompt at each step of the generation which is designed to
% keep the story relevant to the outline and premise while simultaneously maintaining overall coherence (Fig.\ref{fig:method}b).
% \subsubsection{Setup and Outline Generation}\label{sec:setup_outline}
% Concretely, we expand the initial premise into a more detailed plan by generating three main components in order: a setting, a list of characters, and an outline, as shown in Fig.\ref{fig:method}a.

The setting is a simple one-sentence extension of the premise, obtained by using \texttt{The story is set in} to prompt GPT3-Instruct-175B~\cite{ouyang2022training}, a version of GPT3 finetuned to better follow human instructions. 
Next, we use GPT3-Instruct-175B to generate up to three character names and then descriptions, conditioned on the premise and setting. For names, we do rejection sampling using simple heuristics to filter out malformed outputs (Appendix \ref{appendix:character_names}).
Finally, we prompt GPT3-Instruct-175B to write a numbered outline of the story and parse the output into a list of outline points, re-sampling until the list is well-formed. 

These plan components, themselves generated by prompting, will be repeatedly reused to compose prompts for generating story passages in the \draft{} module; hence \textit{recursive reprompting}.

% Throughout this procedure, we additionally employ rejection sampling to filter out overly repetitive outputs, such as those which simply repeat long sections of the initial premise verbatim.

\subsection{\draft{} Module}\label{sec:prompt}

\begin{figure}[htbp]
\centering
\includegraphics[width=1.0\linewidth]{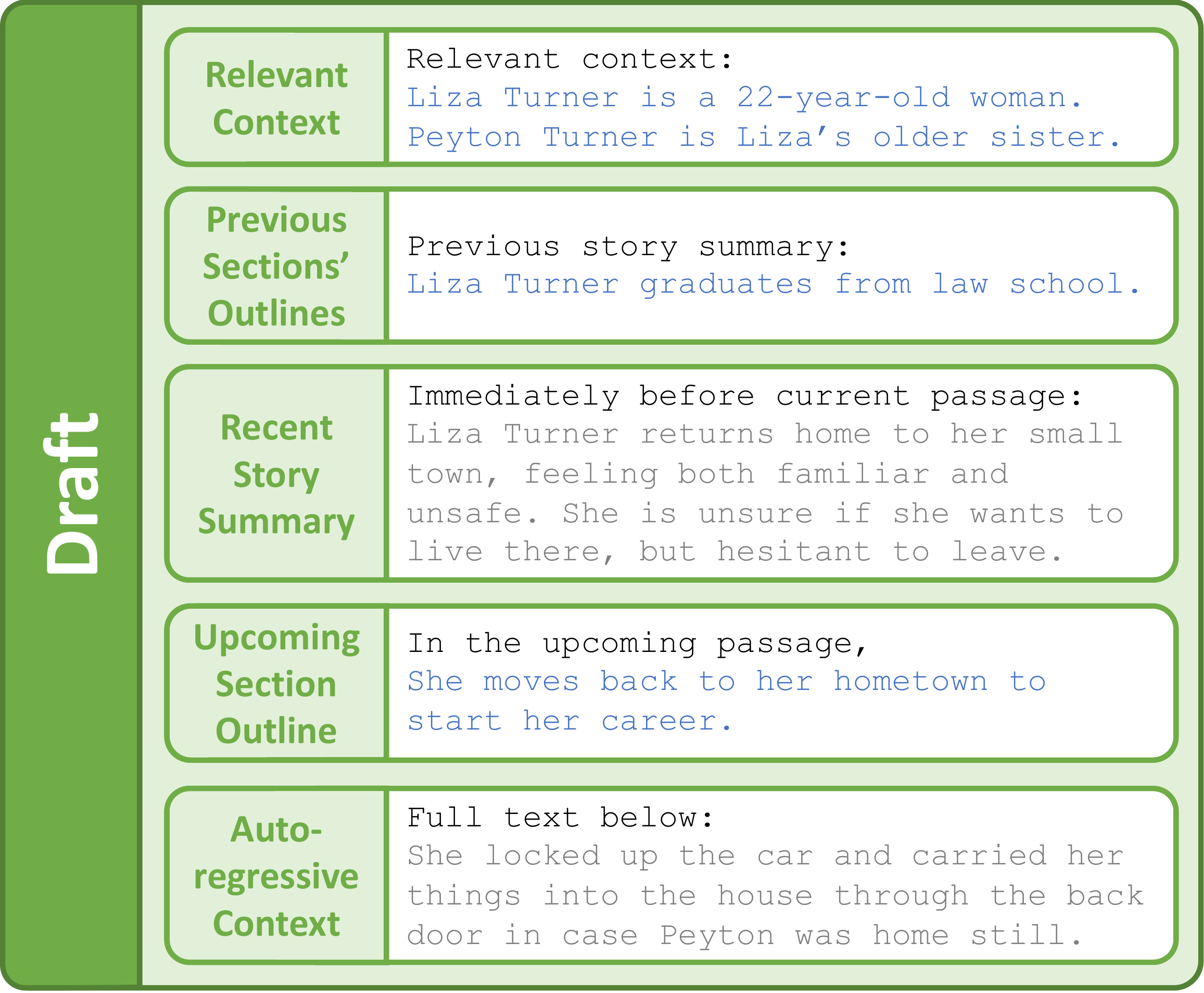}
\caption{Illustration of the prompt constructed in \oursabstract{}'s \draft{} module to generate each next story continuation. Our recursive reprompting approach combines pieces of the plan (blue) and previously generated story (grey) into a single prompt by concatenating the depicted components in order.}
\label{fig:draft}
\vspace{-0.5em}
\end{figure}

% TODO refer to example prompt in figure somewhere; add notation for different concepts
% TODO flip order, use bold text headers

% We now describe how our \draft{} module generates each next passage of the story given the high-level plan from Fig.\ref{fig:method}a and the previously generated story. 

For each point of the outline, we will generate several
story passages before moving on to the next outline point. Each passage is generated as a fixed-length continuation from a structured prompt, which is composed by our recursive reprompting procedure as shown in Figure \ref{fig:draft}. 

The prompt begins with a selection of ``Relevant Context'' shown at the top of Figure \ref{fig:draft}. As the story progresses, we dynamically update the list of character descriptions using a named-entity-recognition-based pipeline, which identifies new entities from each new story passage using Flair~\cite{akbik2018coling} and writes descriptions using GPT3-Instruct-175B. Thus ``Relevant Context'' initially contains all of the premise, setting, and characters shown in Figure \ref{fig:plan}, but subsequently selects only what is most relevant to the most recent story passage using a pretrained Dense Passage Retrieval (DPR) model~\cite{karpukhin2020dense}. 

The remainder of the prompt can be viewed as a coarse-to-fine description of the previous story, following the intuition that an author needs detailed information
about the most recent passage but perhaps only higher-level information about much earlier passages. As shown in Figure \ref{fig:draft}, we include ``Previous Sections' Outlines'' as a very high-level summary of previous larger story sections, followed by a ``Recent Story Summary'' written by GPT3-Instruct-13B\footnote{As economical usage of large language models is becoming increasingly important~\cite{strubell2019energy}, we use the 13B model where we observe it is not substantially worse.}
of a few penultimate passages. At the end we repeat verbatim the immediately preceding passage as ``Autoregressive Context'' from which point the story should continue. Finally, to enforce relevance to the current outline point, 
we include the ``Current Section Outline'' in the prompt just before ``Autoregressive Context.'' %\yuandong{Better visualize the ordering of each part in the prompt sent to GPT-3.}

% TODO mention third person since it's harder to track entities in first person stories. maybe note that most are third person anyway

%The prompt consists of (1) a list of relevant entities, (2) the outline up until the current point, (3) a summary of the previous text, (4) the current outline point, and finally (5) the immediately previous continuation.

% TODO check prompt components

% TODO discuss how we move on to the next outline section

% TODO mention that we use GPT3-175B not instruct
% TODO note selecting entities is a lot like RAG and other similar approaches, but we need to use GPT3-175B to ensure high quality generations. even curie much worse.
Finally, the full prompt is fed to GPT3-175B to generate the next story passage.\footnote{This step does \textit{not} use GPT3-Instruct-175B, as we observed in preliminary experiments that an earlier version of GPT3-Instruct-175B would
frequently repeat sections of the prompt. Generators other than GPT3-175B are also possible in principle:
for example, retrieval-augmented architectures like RAG~\cite{lewis2020retrieval} or architectures designed for long-range dependencies like S4~\cite{gu2021efficiently}. However, it is critical to use a sufficiently high-quality language model: even scaling down to GPT3-13B resulted in noticeably less coherent outputs in our preliminary experiments. %Thus we use GPT3-175B throughout.
}

\subsection{\rewrite{} Module}
\label{sec:reranker}

\begin{figure}[htbp]
\centering
\includegraphics[width=1.0\linewidth]{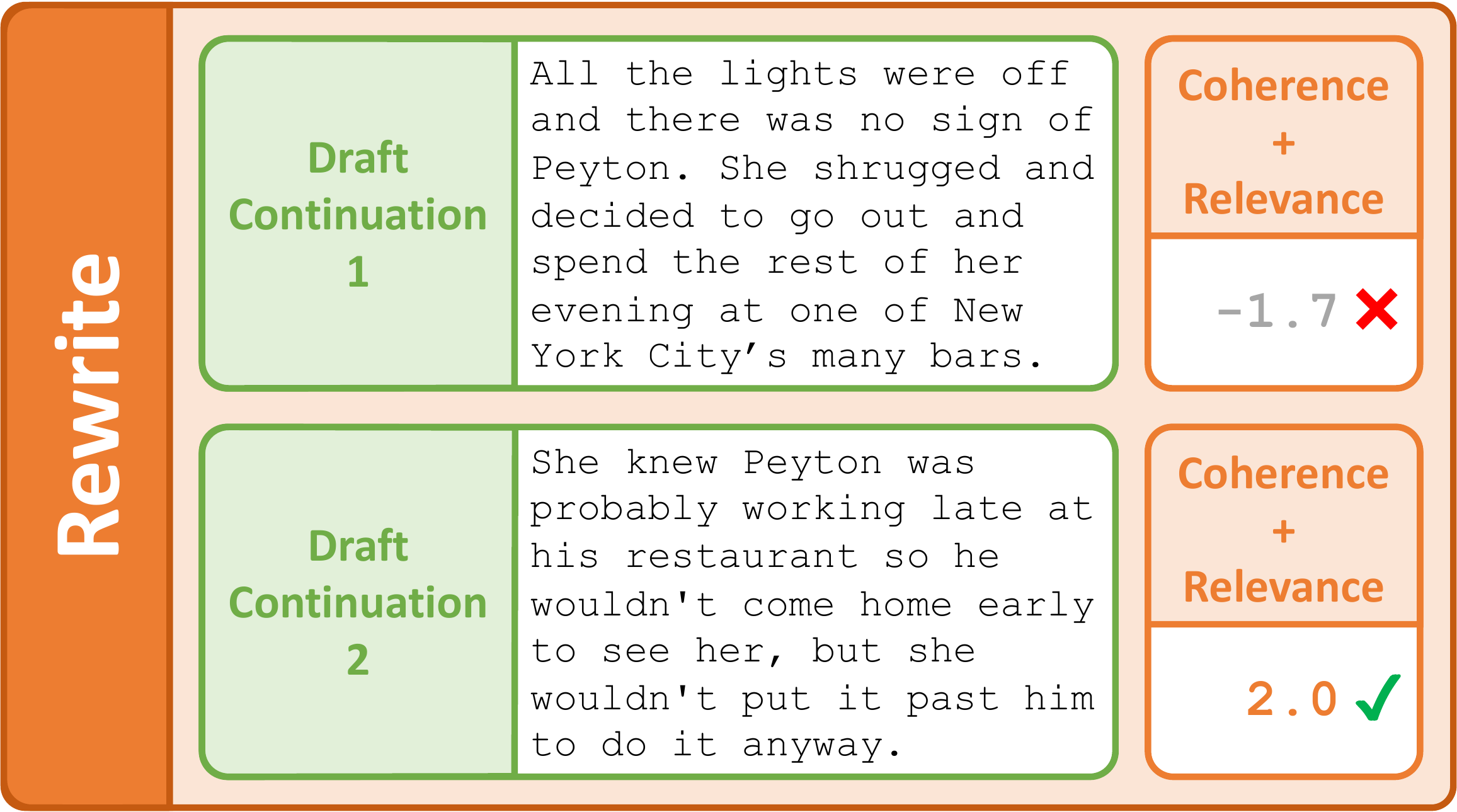}
\caption{\oursabstract{}'s \rewrite{} module reranks the \draft{} module's continuations for coherence and relevance.}
\label{fig:rewrite}
\vspace{-0.5em}
\end{figure}

The generator's first output continuation is often low-quality, even with the planning and recursive reprompting in the \plan{} and \draft{} modules. Humans may encounter a similar problem after a first draft, particularly upon receiving feedback from others, and be forced to rewrite a passage altogether. 
Our \rewrite{} module models this rewriting process by reranking \draft{} module outputs based on coherence with
the previous passage and relevance to the current outline point (Figure \ref{fig:rewrite}).
% Specifically,
% When humans need to revise text, they can either wholly rewrite when the text is utterly unsatisfactory, or they can simply edit 
% when the existing text is decent but not quite good enough. Frequently, they may do both, first rewriting and subsequently editing further.
% We implement this intuition as a proof of concept in our revision module, which first selects the best of several continuations via reranking, and second
% improves the chosen continuation further by editing. Concretely, 
% we rerank by overall coherence and premise relevance.
% , 
% while our editing system will focus on detecting and correcting long-range factual inconsistencies.
% \subsubsection{Reranking System}\label{sec:reranker}
% Specifically, we employ a coherence model and a relevance model,
% which together rescore outputs (Fig.\ref{fig:method}c) based on the joint probability of coherence with
% the previous passage and relevance to the current outline point.

We note that this \rewrite{} module is the only part of \oursabstract{} which uses prior story data. All of the modules which actually \textit{generate} text (\plan{}, \draft{}, and to some extent \edit{}) do not require prior data.%Thus \oursabstract{}'s generation is not limited to the domain of any particular story dataset. 

\medskip
\noindent\textbf{Coherence Reranker.}
We train a discriminative model to predict whether a continuation is coherent with the previous story. As data, we split stories from the WritingPrompts dataset~\cite{fan2018hierarchical} into passages up to 1000 tokens long, labeling the ending up to 200 tokens as the gold continuation. Inspired by the contrastive learning setup of \citet{wang2020narrative} and \citet{guan2020knowledge}, we obtain negative examples by replacing the gold continuation with a random other continuation from either the same story or a different one. We then finetune a pretrained Longformer-Base~\cite{beltagy2020longformer}
to classify whether a continuation is the true continuation for a given passage.
% Following the contrastive learning setup of \citet{wang2020narrative} and \citet{guan2020knowledge}, negative examples are constructed by replacing the true passage with either a different passage from the same story or a random passage from another story.

\medskip
\noindent\textbf{Relevance Reranker.}
We train a relevance model with the same architecture as our coherence model to predict whether a continuation is relevant to the current outline point. We construct a dataset of 2000 training examples, where each example consists of a 200-token story passage from WritingPrompts and a brief summary written by GPT3-Instruct-13B. Negative examples are constructed by selecting the summary of a different passage, whether in the same story or a different one.

% \textbf{Coherence Model.}
% We additionally train a coherence model with the same architecture as our relevance model. The data is sampled from WritingPrompts stories
% of X words range TODO. Training examples consist of contiguous 1000-token (TODO check length) story passages, and the task is to detect
% whether the last X tokens are the the true continuation of the preceding passage. As in the relevance reranker, negative examples for contrastive learning are constructed by
% replacing the last X tokens by a different passage from the same story (either earlier or later in the story) or a random passage from a different story. 

% \textbf{Relevance Model.}
% We train a discriminative model for detecting whether a continuation is relevant to the current outline point. 
% We split stories from the WritingPrompts dataset~\cite{fan2018hierarchical} into passages up to 1000 tokens long, and prompt
% GPT3-Instruct-13B for one-sentence summaries of each passage. We then finetune a pretrained Longformer-Base model~\cite{beltagy2020longformer}
% to classify passage-summary pairs based on whether or not the summary corresponds to the preceding passage. 
% Following the contrastive learning setup of \citet{wang2020narrative} and \citet{guan2020knowledge}, negative examples are constructed by replacing the true passage with either a different passage from the same story or a random passage from another story. %TODO double check repeat vs shuffle, was there prior context?

\medskip
\noindent\textbf{Additional Heuristics.} % TODO does this even merit its own header? frame this as preserving narration style
Finally, we filter out continuations with some writing problems which are easy to detect via rule-based heuristics.
For example, we check for repetition issues,
e.g., repeating chunks of the structured prompt. 
% The latter also often involve jarring changes in narration style. 
Similarly, to maintain consistent narration, we filter out first person continuations to enforce a consistent third person perspective. Full details in Appendix \ref{appendix:heuristics}.%(searching for the presence of ``I'' and ``we'' outside of quotes) to detect and filter out first person continuations
% as we focus on generating third person stories as discussed in Section TODO.

\subsection{\edit{} Module}\label{sec:editor}

\begin{figure}[htbp]
\centering
\includegraphics[width=1.0\linewidth]{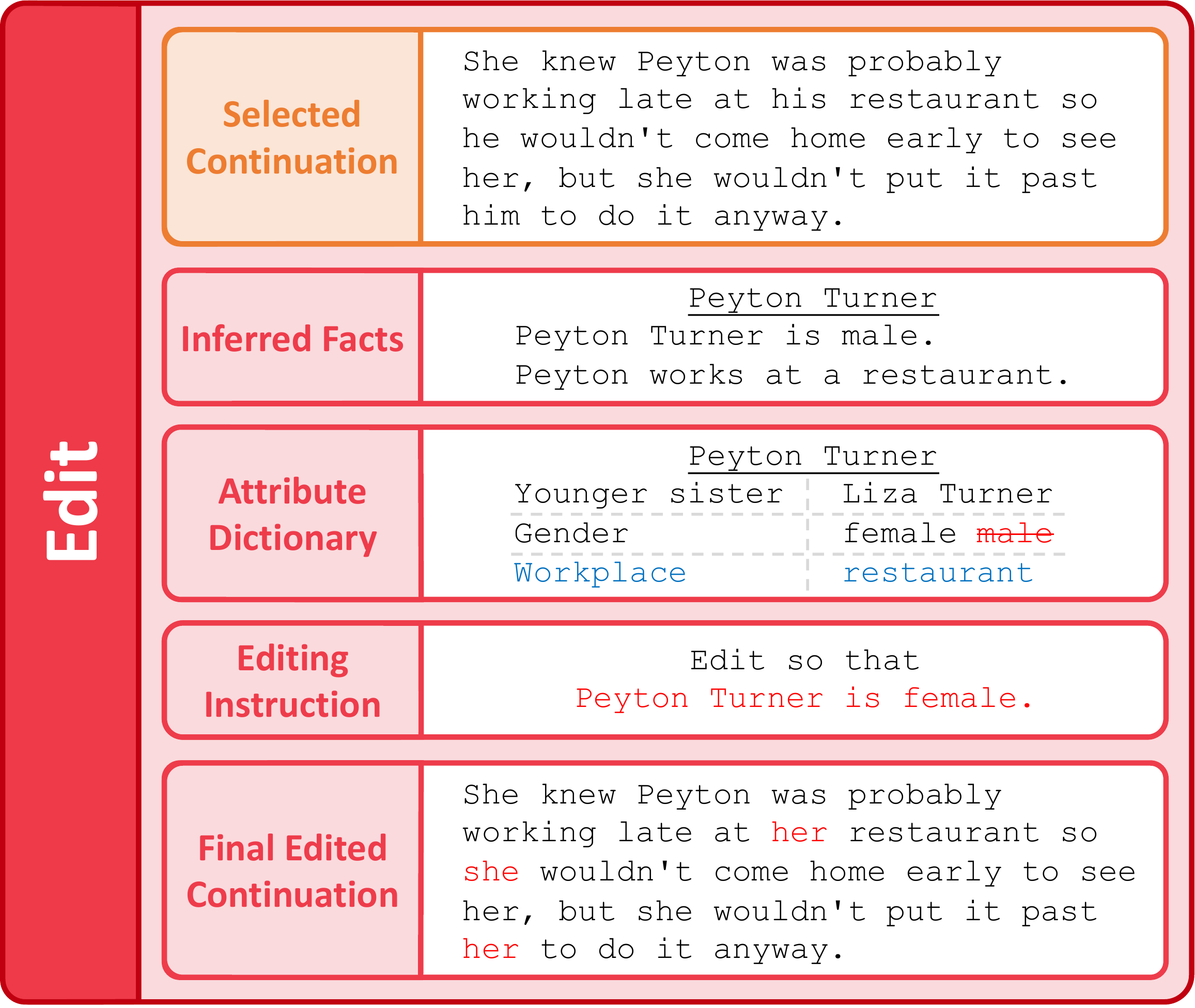}
\caption{Illustration of \oursabstract{}'s \edit{} module. Starting from the \rewrite{} module's best continuation, we infer natural language facts about each character, and convert them to attribute-value pairs. New values (blue) are added to the attribute dictionary, and contradictory values (red) are corrected.}
\label{fig:edit}
\vspace{-1em}
\end{figure}

In contrast to the \rewrite{} module which reranks complete alternate continuations, the \edit{} module makes local edits to further refine a passage produced by careful planning, drafting, and rewriting. %, our \edit{} module makes further local edits. 

Specifically,
% In principle, one could edit outputs with any goal in mind, such as further improving relevance and coherence. %However, we design our \edit{} module to tackle a problem which we argue is most appropriately solved by post-editing: 
% Here 
we aim to remove long-range factual
inconsistencies. %If Alice is introduced as Bob's sister early on in the story, she should not suddenly
% become Bob's mother a thousand words later, even if she does not appear at all in the intervening passage. 
When a human detects a small factual discontinuity upon proofreading, he or she might simply edit the offending detail, rather than making major changes to the high-level plan or doing substantial rewriting. 
Our Edit module mimics this process in two steps: \textit{detecting} factual inconsistencies, and \textit{correcting} them.

\medskip
\noindent\textbf{Detecting Factual Inconsistencies.}
An inconsistency involves two statements. As the number of statement pairs scales quadratically with story length, naively comparing all pairs can result in a sea of false positive ``contradictions'' (Section \ref{sec:editor_analysis}). Flagging inconsistencies while avoiding false positives requires overwhelming precision.

% We consider detection the harder problem compared to correction

% , because detecting factual inconsistencies in a story
% thousands of words long is akin to finding a needle in a haystack---while simultaneously avoiding labeling any random strand of hay as a needle, which could lead to incorrect ``corrections.'' Thus the detection system must be overwhelmingly precise.
% to avoid returning dozens of false positive ``inconsistencies'' for every real one detected. 
% Moreover, stories present a unique
% challenge in the form of change over time: Alice may go to the grocery in one scene and to the airport in another, which
% any entailment model will classify as a contradiction, but which in truth is nothing of the sort. % TODO discuss contrast with traditional fact verification/entailment setting where fewer neutrals; cite some literature. super hard even though it's not multihop
\textit{Task Framing.} To make the task more tractable,
%and to avoid many false positives, 
we focus on factual inconsistencies 
in character attributes (e.g., age, occupation, 
relationship to another character). %Such an approach avoids many (but not all) examples of ``contradictions'' which are merely changes in time, 
% and we leave a more principled handling of such cases to future work. 
At a high level, our detection system maintains a compact knowledge base in the form of Figure \ref{fig:edit}'s ``Attribute Dictionary'' for each character. With each new story passage, we check for contradictions against only these attribute-value dictionaries instead of all previous text.
The dictionaries are then updated for the new passage, and new dictionaries are created for new characters when detected as described in Section \ref{sec:prompt}.
% Upon finding an attribute-value pair for some character which an entailment model judges inconsistent with the existing dictionary, we flag a contradiction. 

% TODO Check models
Thus, the core of our detection system is a high-precision information extraction procedure for obtaining attribute-value pairs for a given character from a story passage. Rather than hard-coding a fixed set of attributes, our system is inspired by Open Information Extraction~\cite{etzioni2008open}, in order to capture the wide variety of possible attributes which may be salient in different stories. 

\textit{Implementation Details.} We begin by prompting GPT3-Instruct-175B for a numbered list of facts about the given character, shown as ``Inferred Facts'' in Figure \ref{fig:edit}. Each fact is fed with a few-shot prompt to GPT3-Instruct-13B to extract attribute keys. We then prompt GPT3-Instruct-13B with the fact and each attribute key to obtain complete attribute-value pairs. In steps prone to hallucination, we generate three outputs and keep only those which are repeated, or entailed by other outputs according to a BART-Large-based~\cite{DBLP:journals/corr/abs-1910-13461} entailment model trained on MNLI~\cite{N18-1101}. See Appendix \ref{appendix:editing_details} for complete details on information extraction, with example prompts.
%for complete implementation details and example prompts.

Finally, we add new pairs to our dictionary, and use the entailment model to flag contradictions between new and old values for the same key.% according to a BART-Large~\cite{DBLP:journals/corr/abs-1910-13461} entailment model trained on MNLI~\cite{N18-1101}. %Finally, for relations between characters which are only detected for one character but not the other, we ``complete'' the relation
% by adding it to the other character's dictionary as well. 
%TODO check details to see if there were any other details in the steps to report here

% TODO note it's kind of engineering heavy; this is just a proof of concept and we envision a more general system. 

\medskip
\noindent\textbf{Correcting Factual Inconsistencies.}
Once an inconsistency is detected, we frame the task of correcting it as controlled text editing. The original natural language fact (i.e., ``Inferred Facts'' in Figure \ref{fig:edit}) from which we extracted the contradicted attribute-value pair now becomes the basis for 
the ``Editing Instruction'' in Figure \ref{fig:edit}. This instruction is then fed along with the original continuation to the beta GPT3 Edit API.
% In this work we use the beta GPT3 Editing API off-the-shelf, prompting it to correct for the desired fact.

    % - [ ] need a planning system
    %     - [ ] our instantiation of planning system: describe the premise setting characters outline etc. 
    % - [ ] also need an editing system; can't keep track of all details in the context window, and it's hard to do long term memory anyway
    %     - this is a general revision system. when people revise, they can either wholly rewrite, or they can revise without completely rewriting. both have been tried in prior work, in this work we try a mix of both approaches, with the goal of improving the coherence and relevance to the outline. 
    %     - [ ] discuss the reranking system and different reranking components, there's two rerankers and also the heuristics for not going into bad narrative style (?)
    %     - [ ] our instantiation of editing system: structured approach that extracts key-value attributes of characters
    % - note limitations: you have to use a GPT3 scale system otherwise you're tackling qualitatively different challenges, e.g. preliminary BART experiments; even curie is noticeably less coherent.
\section{Evaluation}\label{sec:evaluation}

\begin{table*}[htbp]
\small
\centering
\begin{tabular}{lccccc}
\toprule
\textbf{Method}          & \textbf{Interesting }$\uparrow$ & \textbf{Coherent} $\uparrow$ & \textbf{Relevant} $\uparrow$ & \textbf{Humanlike} $\uparrow$ & \textbf{Misc.\ Problems} $\downarrow$ \\
\midrule
\rolling{}         & 45.0          & 45.7     & 44.0       & 74.0        & 1.20           \\
\ours{}           & 54.3        & \textbf{60.0}       & \textbf{64.0}       & \textbf{83.3}      & \textbf{1.07}          \\

\midrule

\rollingfinetune{} & 52.7        & 48.7     & 49.3     & 74.7      & 1.48          \\
\ours{}            & 53.7        & \textbf{60.0}       & \textbf{65.3}     & 80.0        & \textbf{1.35}          \\
\bottomrule
\end{tabular}
\caption{\small Comparison of \ours{} against two baselines, \rolling{} and \rollingfinetune{}, in two separate experiments. The first two rows show a pairwise comparison between \rolling{} and \ours{} and the last two rows show the equivalent comparison between \rollingfinetune{} and \ours{}. Bolding indicates significant differences with $p<0.05$ on a paired $t$-test. Workers judged stories from \ours{} as significantly more coherent and relevant to the initial premise, in addition to having fewer writing problems. }
\label{tab:results_baselines}
\vspace{-0.5em}
\end{table*}

% define new column type
\newcolumntype{Y}[1]{%
  >{\small\everypar{\hangindent=1em}\arraybackslash}p{#1}%
}

% \subsection{Evaluation Setup.} 

\noindent\textbf{Task Setup.} We frame the task as generating a story given a brief initial premise. As a ``story'' is difficult to define in a rule-based manner, we do not impose any rule-based constraints on acceptable outputs, but will instead evaluate via several human-annotated metrics as described later.

To generate the initial premises, we prompt GPT3-Instruct-175B with high temperature to acquire 100 diverse premises.\footnote{Combining this simple premise generation scheme
with \oursabstract{} yields a story generation system which operates fully from scratch, with no input premise required.} %and compare the corresponding stories generated by \oursabstract{} to those from baselines. 
All premises and stories are in English. 

\medskip
\noindent\textbf{Method Instantiation.} For fair comparison, it is desirable for the concrete implementation (henceforth \ours{}) of our \oursabstract{} framework to output stories of consistent length. While \oursabstract{} is capable of generating shorter or longer stories (see e.g., our 7500-word example in Appendix \ref{appendix:longer_story}), here we aim for roughly 3000 tokens (2000-2500 words).\footnote{See Appendix \ref{appendix:length_analysis} for analysis on how story length may impact quality.} Thus we re-sample the initial outlines (Section \ref{sec:setup_outline}) until they contain exactly three points, and generate exactly four 256-token continuations for each outline point before moving on to the next. As a story-ending mechanism, we use the GPT3-175B Insert API to complete the story to the suffix ``The End.'' Of course, more adaptive schemes for moving on to the next outline point and/or ending the story are possible, and we explore one possible ``outline alignment'' method in Appendix \ref{appendix:longer_story}. 

% TODO describe main details / numbers choices, e.g. num characters. discuss the 1024 context limit and what composes it. 

% In particular, as a concession to evaluation practicality, we re-sample the outlines described in Section TODO until they contain exactly three points. 
% Additionally, while expanding each point of the outline, we generate exactly four length-256 continuations before moving on to the next point, 
% rather than using an adaptive stopping criterion. The result is that each generated story is of almost exactly the same length (the only 
% variation being the length of the ending from Section TODO), facilitating fairer comparison. TODO check the actual length we used, was it 4/4?

% TODO describe any additional changes/details worth noting. 

\medskip
\noindent\textbf{Baselines.} As prior methods focus on dramatically shorter stories compared to \oursabstract{},
they are difficult to compare to directly.\footnote{Even the \textit{premises} used as starting points in our task can be as long or longer than the final stories generated in several previous works. We believe that adapting any of the prior systems from our related work to function on our long-form story generation task could be an interesting contribution in its own right. In fact, \oursabstract{} itself can be viewed as our attempt to extend and combine high-level planning and revision ideas from prior work, while simultaneously redesigning them to be able to leverage large out-of-the-box pretrained generators (GPT3), to scale up to long-form generation.} Instead, we use the following two GPT3-175B-based baselines.\footnote{Smaller (non-GPT3-175B) generators yielded qualitatively worse outputs in preliminary experiments.}
%we observe that generation quality suffers
% noticeably when decreasing the size of the generator, and qualitative issues which are rarer in GPT3-175B may become frequent (e.g., highly repetitive generations).
% Hence we did not formally evaluate non-GPT3-175B baselines. 
% }
% TODO are any better justifications needed for not comparing to prior methods?

\begin{enumerate}[topsep=0pt,itemsep=-1ex,partopsep=1ex,parsep=1ex]
    \item \rolling{}, a baseline which generates 256 tokens at a time via GPT3-175B using the premise and all previously
    generated story text as the prompt, left-truncating the prompt if it exceeds 768 tokens. Hence, a ``rolling window''
    with maximum context length 1024 (the same maximum context length used in \ours{}). 
    After 3072 tokens are generated, we use the same story-ending mechanism as \ours{}.%, completing the story to the suffix ``The End.'' %Thus the final
    % stories are almost exactly the same length as those generated by \ours{}. 
    \item \rollingfinetune{}, which is identical to \rolling{} except that GPT3-175B is first finetuned on several hundred passages from WritingPrompts stories of at least 3000 tokens.\footnote{We initially considered a third rolling window baseline using GPT3-Instruct-175B rather than GPT3-175B, but observed that this baseline frequently devolved into highly repetitive text or gibberish. Thus we do not report a formal comparison. In any case, \rolling{} is in some sense the best comparison, as \ours{} uses the same un-finetuned GPT3-175B generator.}
\end{enumerate}

\noindent\textbf{Metrics.} As our main metrics, we track the percentage of stories which are:

\begin{enumerate}[topsep=0pt,itemsep=-1ex,partopsep=1ex,parsep=1ex]
    \item \textbf{Interesting.} Interesting to the reader.
    \item \textbf{Coherent.} Plot-coherent.
    \item \textbf{Relevant.} Faithful to the initial premise.
    \item \textbf{Humanlike.} Judged to be human-written.
\end{enumerate}

We additionally track how often generated stories suffer from any of the following writing issues: 

\begin{enumerate}[topsep=0pt,itemsep=-1ex,partopsep=1ex,parsep=1ex]
    \item \textit{Narration.} Jarring change(s) in narration and/or style. 
    \item \textit{Inconsistent.} Factually inconsistent or containing very odd details. 
    \item \textit{Confusing.} Confusing or difficult to follow. 
    \item \textit{Repetitive.} Highly repetitive. 
    \item \textit{Disfluent.} Frequent grammatical errors. 
\end{enumerate}

Binary indicators for these issues are summed and reported together as \textbf{Misc.\ Problems} in the main text, with individual numbers in Appendix \ref{appendix:writing_problems_metrics}.  

All metrics are judged by Amazon Mechanical Turk workers, who are shown a premise
and two corresponding stories in random order: one generated by \ours{} and one by a baseline. 
%The workers are asked to quickly skim both stories before annotating. 
For interest, coherence, and relevance, workers indicate which story is better,
or that both or neither are good. For the remaining metrics (humanlike and writing issues), workers mark each story individually. See Appendix \ref{appendix:mturk_survey} for an example Mechanical Turk task.

Each pair of stories is judged by three workers. We report results
individually for each pairwise comparison between \ours{} and each baseline, never mixing numbers from different comparisons. 
Workers are segregated by comparison to prevent bias. 

% Full details on the Mechanical Turk task, including an example, are in Appendix \ref{appendix:mturk_survey}. 

% TODO note that baselines based on smaller models just face qualitatively different issues?

% TODO main results table against the two baselines. bold p<0.05, italics <0.1?

\begin{table}[!t]
\small
\begin{tabular}{@{}Y{0.95\linewidth}@{}}
\toprule
\textbf{\texttt{PREMISE:}} \texttt{After years of being content with only a small inheritance, Sarah's life is turned upside down when she discovers she is actually a millionaire. She must learn to navigate her new life of luxury while also dealing with the guilt she feels over her new-found fortune.}\\
\midrule
\textbf{\texttt{GENERATED OUTLINE:}}\\
\texttt{1. Sarah learns that she is a millionaire.}
\\
\texttt{2. Sarah must learn to navigate her new life of luxury.}
\\
\texttt{3. Sarah deals with the guilt she feels over her new-found fortune.}
\\
\midrule
\textbf{\texttt{GENERATED STORY:}} \\
\texttt{Sarah Jensen never knew she had inherited a fortune until her mother died. In fact, she had never met her father \textcolor{gray}{...[15 words]...}} \\
\texttt{The only thing Sarah did know was that he had died long ago and that her mother had been left a sizeable inheritance. \textcolor{gray}{...[23 words]...}} \\
\texttt{Sarah's mother never spent any of the money ... all that mattered to her was how much it would mean to her daughter when she inherited it \textcolor{gray}{...[81 words]...}} \\
\texttt{Sarah learned that she was a millionaire. \textcolor{gray}{...[417 words]...}} \\
\texttt{She would never have had a chance like this - an opportunity to be financially independent, to live on her own \textcolor{gray}{...[544 words]...}} \\
\texttt{Now that she was rich, things were going to be different for her. She could afford a hotel suite and a lavish meal if she wanted it. \textcolor{gray}{...[450 words]...}} \\
\texttt{The words "Forgive me, father, for I have sinned" came unbidden into her mind as she reached out for her drink and took a large sip straight from the glass. \textcolor{gray}{...[383 words]...}} \\
\texttt{But the guilt was weighing on her now. \textcolor{gray}{...[346 words]...}}\\
\bottomrule
\caption{A heavily abridged \ours{} example generation (see Appendix \ref{appendix:examples}, Tables \ref{tab:premise5_ourssetup} and \ref{tab:premise5_ours} for complete setup and story). While certainly not perfect, the story demonstrates strong long-range coherence and faithfulness to the premise.}
\vspace{-2em}
\label{tab:abridged_example}
\end{tabular}
\end{table}

\begin{table}[!t]
\small
\begin{tabular}{@{}Y{0.95\linewidth}@{}}
\toprule
\textbf{\texttt{PREMISE:}} \texttt{After the loss of her father, Shannon is determined to finish college and make something of herself. However, when she's sucked into a time portal and lands in the 19th century, Shannon has to find a way back to her own time before it's too late.}\\
\midrule
\textbf{\texttt{GENERATED OUTLINE:}}\\
\texttt{1. Shannon is transported to the 19th century where she meets Jaxon Montgomery.}
\\
\texttt{2. Shannon and Jaxon fall in love, but she is still determined to find a way back to her own time.}
\\
\texttt{3. Shannon discovers that her father is still alive in the 19th century and she must choose between staying with Jaxon or returning to her own time.}\\
\midrule
\textbf{\texttt{GENERATED STORY:}} \\
\texttt{Shannon was running late and hurried along the sidewalk. She was supposed to meet her best friend Hannah \textcolor{gray}{...[423 words]...}} \\
\texttt{When Jaxon proposed, we'd already been dating for over a year, so I [Hannah] went with it. \textcolor{gray}{...[470 words]...}} \\
\texttt{Thinking about him [Jaxon] made Shannon smile slightly as she recalled the first time they met ... right here in this backyard after noticing one another edging closer together \textcolor{gray}{...[197 words]...}} \\
\texttt{Shannon smiled when she looked into his eyes \textcolor{gray}{...[176 words]...}} \\
\texttt{[Jaxon asks,] "What do you mean that you're from the future?" \textcolor{gray}{...[319 words]...}} \\
\texttt{She looked him [Jaxon] straight in the eyes and said firmly, "I must go back to my time now.\textcolor{gray}{...[199 words]...}} \\
\texttt{She felt tears stinging in her eyes \textcolor{gray}{...[73 words]...}} \\
\texttt{There was no way she could ever go back to her own time after all this. \textcolor{gray}{...[287 words]...}} \\
\texttt{Shannon looked down at her feet again, then back up at Jaxon and said, "My father is alive?" \textcolor{gray}{...[47 words]...}} \\
\texttt{Jaxon gently rubbed Shannon's back in support and quietly said, "Yes, my love. He is alive and well. \textcolor{gray}{...[52 words]...}}\\
\texttt{Jaxon shook his head and said, "No, Shannon. I want you to be happy. And if that means going back to your own time, then so be it." \textcolor{gray}{...[72 words]...}}\\
\texttt{Shannon Randall vanished from the 19th century, never to be seen again.}\\
\bottomrule
\caption{Another heavily abridged \ours{} example generation (see Appendix \ref{appendix:examples}, Tables \ref{tab:premise2_ourssetup} and \ref{tab:premise2_ours} for complete setup and story). \ours{} initially fails to follow the premise and outline, and in the beginning Jaxon is incorrectly introduced as Hannah's love interest. However, both issues are corrected in the subsequent story.}
\vspace{-2em}
\label{tab:abridged_example2}
\end{tabular}
\end{table}

\begin{table*}[]
\small
\centering
\begin{tabular}{lccccc}
\toprule
\textbf{Method}          & \textbf{Interesting }$\uparrow$ & \textbf{Coherent} $\uparrow$ & \textbf{Relevant} $\uparrow$ & \textbf{Humanlike} $\uparrow$ & \textbf{Misc. Problems} $\downarrow$ \\
\midrule

\noplanner{}       & 50.3                 & 46.7              & 50.7              & 70.0               & 1.33                   \\
\ours{}            & 59.7                 & \textbf{63.3}     & \textbf{63.7}     & \textbf{80.0}      & 1.25                   \\
% \midrule

% \norerankeditor{}  & 43.3                 & 48.7              & 49.0              & 69.0               & 1.25                   \\
% \ours{}            & \textbf{58.3}        & 58.7              & \textbf{64.3}     & 73.3               & \textbf{1.04}          \\
\midrule

\norerank{}        & 46.3                 & 42.3              & 42.7              & 59.7               & 1.48                   \\
\ours{}            & \textbf{56.7}        & \textbf{56.0}     & \textbf{63.3}     & 67.3               & \textbf{1.17}          \\
\midrule
\noeditor{}        & 55.0                 & 60.3              & 59.3              & 87.7               & 1.10                   \\
\ours{}            & 57.0                 & 57.3              & 59.3              & 87.0                 & 1.12                   \\

\bottomrule
\end{tabular}
\caption{\small Ablations on individual components of \ours{}, 
removing the \plan{}, \rewrite{}, and \edit{} modules respectively. Each two rows show a pairwise comparison experiment between \ours{} and the corresponding ablation.
%removing the planning module, removing the revision module (both rerankers and editor), removing only rerankers, and removing only the editor respectively. 
Bolding indicates significant differences with $p<0.05$. Both the \plan{} and \rewrite{} module are critical to performance, but the \edit{} module makes little difference.}
\label{tab:ablations}
\vspace{-0.5em}
\end{table*}

\medskip
\noindent\textbf{Results.} As shown in Table \ref{tab:results_baselines}, \ours{} is highly effective at writing a longer story following a desired premise 
while maintaining a coherent overarching plot, validating our design choices inspired by the human writing process as well as our recursive reprompting approach to generation. \ours{} significantly and substantially improves over \rolling{} and \rollingfinetune{} in both coherence and relevance.
 Annotators also marked \ours{}'s stories as having significantly fewer miscellaneous writing problems. Finally, \ours{} demonstrates strong performance in an absolute sense: annotators thought 83.3\% and 80.0\% respectively of \ours{}'s stories were written by humans in the two comparisons. Table \ref{tab:abridged_example} shows a heavily abridged example story by \ours{}, exhibiting strong coherence and premise relevance.
 
 Nonetheless, we observe qualitatively that \ours{} still has much room for improvement. Two common issues are illustrated in Table \ref{tab:abridged_example2}. First, while \ours{}'s stories almost always follow the premise to some degree---unlike our baselines' stories---they may fail to capture all parts of the premise, and may fail to follow parts of the outline generated by the \plan{} module (e.g., the first part of the premise and outline in Table \ref{tab:abridged_example2}). Second, due to failures in the \rewrite{} and especially \edit{} modules, there remain some confusing passages or contradictory statements: for example, in Table \ref{tab:abridged_example2}, the character Jaxon has a contradictory identity in some places. 
 
However, unlike rolling window methods, \ours{}'s planning infrastructure is able to ``self-correct'' back to the original high-level plot despite early errors in generation. The latter part of the story in Table \ref{tab:abridged_example2} illustrates this interesting capability.
 
 % Some passages remain confusing or contradictory.
 % %, perhaps partially due to sometimes very exotic premises. 
 % While nearly all of \ours{}'s stories follow the initial premise to a decent degree---which cannot be said of stories written by baselines---\ours{}'s stories often deviate from parts of our more detailed outline (Figure \ref{fig:plan} bottom), especially for more unusual or outlandish premises. 
 
 See Appendix \ref{appendix:examples} for additional complete, i.i.d.\ examples of stories from both \ours{} and baselines.
\section{Analysis}\label{sec:analysis}

\subsection{Ablation Study}

\noindent\textbf{Ablated Modules.} We investigate the relative contribution of the individual modules of \oursabstract{}: \plan{}, \draft{}, \rewrite{}, and \edit{}.
We ablate each module in turn as follows, except the \draft{} module as it is unclear how our system would operate without it.

\begin{enumerate}[topsep=0pt,itemsep=-1ex,partopsep=1ex,parsep=1ex]
    \item \noplanner{}, a version of \ours{} without the \plan{} module. Accordingly, we remove the recursive reprompting in \draft{}. Thus \noplanner{} generates text identically to the \rolling{} baseline, but is revised by our \rewrite{} and \edit{} modules. 
%     \item \norerankeditor{}, a version of \ours{} without the revision module. It creates
%     structured prompts in the same way as \ours{} but simply accepts the first generated output
%     without reranking or further editing. 
% \end{enumerate}

% In addition to ablating the revision module as a whole, we ablate the two main submodules which comprise it as well:

% \begin{enumerate}
    \item \norerank{}, a version of \ours{} without the \rewrite{} module reranking. %in which the revision module no longer reranks,
    % but which still edits as described in Sec. \ref{sec:editor}.
    \item \noeditor{}, a version of \ours{} which no longer edits using the \edit{} module.
    %still reranks as in Sec. \ref{sec:reranker}, but no longer edits afterward. 
\end{enumerate}

\medskip
\noindent\textbf{Results.} Table \ref{tab:ablations} shows that both the \plan{} and \rewrite{} modules, mimicking the human planning and rewriting processes, are critical for overall plot coherence
and premise relevance. %TODO additional observations, interest/humanlike, other writing issues; emphasize more how motivated+useful these are
% However, upon closer inspection of the subsystems within the revision module, we observe that the main driver of performance gain
% is the reranking system, while 
However, the \edit{} module contributes little to these metrics. We also observe qualitatively that there remain many continuity issues in \ours{}'s final stories which are not resolved by our \edit{} module, but which could be fixed by an attentive human editor. Such continuity issues range from non-character-centric inconsistencies, to facts which change over time, to outline plot points which were omitted in the story.
% TODO speculate that the bad results might also be that inconsistencies are hard to find by skimming

\subsection{Further Analysis of \edit{} Module}\label{sec:editor_analysis}

We use a controlled setting to investigate if the \edit{} module can at least detect the character-based factual inconsistencies for which it is designed. We will refer to our detection subsystem as \structured{} to avoid conflation with the \edit{} module as a whole.

\medskip
\noindent\textbf{Task Setup.} We construct an evaluation dataset as follows. First we generate setups following our \plan{} module, up to but not including the outline. For each setup $s$ we randomly resample a character's description until we manually observe a contradiction with the original, yielding a contradictory setup $s'$. For each of $s$ and $s'$, we generate a story ($t$ and $t'$), resampling until the contradicted attribute appears in the story. If the resampling fails after 5 attempts we restart the whole procedure. We generate 50 $(s, s', t, t')$ tuples in total; see Appendix \ref{appendix:consistency_example} for an example. 

The task is then framed as classification: the method should judge $(s, t)$ and $(s', t')$ as consistent and $(s, t')$ and $(s', t')$ as contradictory. Thus the 50 $(s, s', t, t')$ tuples yield 200 input pairs. %The metric is ROC-AUC score~\cite{hanley1982meaning} based on predicted contradiction probabilities.

\medskip
\noindent\textbf{Baselines.} We construct two simple baselines using the same BART-Large-MNLI entailment model used in \structured{}. Given a $(s, t)$ pair, the first baseline, \entailment{}, simply checks each sentence of $s$ pairwise against each sentence of $t$, and returns the maximum probability of contradiction across all pairs. The second baseline, \entailmentdpr{}, checks each sentence of $t$ against only one sentence of $s$ based on relevance judged by DPR~\cite{karpukhin2020dense}.

% behavior in excerpts which do contain
% factual inconsistencies related to character attributes, which the editing systems is designed to fix. 
% Specifically, we will measure our system's ability to \textit{detect} such inconsistencies. 

% TODO setup, discuss biases in construction of dataset (TODO can refer to example where something got detected to show example)

% TODO discuss baselines: simple entailment model based system. also DPR based entailment system to compare. tune threshold on dev set?

\begin{table}[]
\small
\centering
\begin{tabular}{lc}
\toprule
\textbf{Method}         & \textbf{ROC-AUC} $\uparrow$ \\
\midrule
\entailment{}     & 0.528 \\
\entailmentdpr{} & 0.610 \\
\structured{}           & \textbf{0.684}\\
\bottomrule
\end{tabular}
\caption{ROC-AUC score of predicted contradiction probabilities for different methods on our evaluation set. \structured{} outperforms our two entailment-based baselines.}
\label{tab:consistency_detection}
\vspace{-0.8em}
\end{table}

\medskip
\noindent\textbf{Results.} As shown in Table \ref{tab:consistency_detection}, when detecting character-based inconsistencies, \structured{} outperforms the two baselines according to the standard ROC-AUC metric for classification~\cite{hanley1982meaning}. Indeed, the most naive \entailment{} system's ROC-AUC score is barely better than chance performance (0.5), highlighting the core challenge wherein the detection system must be overwhelmingly precise. Moreover, \structured{} is designed to scale to longer passages; we hypothesize that the performance gap compared to baselines would widen in an evaluation with longer inputs such as the stories from our main experiments. 

Even so, the absolute performance of all systems remains low, even in this simplified setting. Additionally, many of our generated full stories contain non-character-based inconsistencies, such as in the setting or current scene. Some stories also contain false positives (flagged non-contradictions), such as character attributes which change over time.
% TODO note that entailment dpr might be worse in a real setting where the language isn't just stating facts like in the setup string. and although we didn't formally evaluate, we expect that the DPR thing would scale poorly to longer passages while our more structured representation should scale better - explain. and even so we're already better here. ours is designed to work on longer text than these passages. 
% TODO show an example where something got detected
% TODO numbers, also discuss the entailment baseline's threshold that you used and how many false positives it gets at any reasonable threshold
% However, the overall accuracy still leaves much to be desired. 

% Upon closer inspection, we observe that not all stories contain
% factual inconsistencies which can plausibly be fixed by our \edit{} module.
% %, causing the metrics for 
% %\noeditor{} to be very similar to \ours{}. 
% Most stories either do not contain obvious inconsistencies, or 
% contain harder-to-detect inconsistencies which our narrow-focused \edit{} module is not designed to handle. Some stories also contain false positives (flagged non-contradictions), such as character attributes which change over time in the story.

Additionally, while we did not formally analyze the 
GPT3 Edit API's ability to \textit{correct} inconsistencies after they are detected (as this system is largely not our contribution), % is this excuse ok? or TODO just say we focus on the detection analysis since that part's harder
we generally observed that it can fix isolated details but may struggle with larger changes. It also 
sometimes makes undesired edits or additions. 
Taken together, the compounding errors from the detection and correction subsystems make it difficult for our current \edit{} module to effectively improve factual consistency over a multiple-thousand-word horizon, without simultaneously introducing unnecessary changes. 
% TODO Talk more about how you really need this sort of thing to in principle be able to detect and fix inconsistencies; you can't include all the info elsewhere

    % - [ ] ablations with/without planner (just rolling window + editing system on history) to show that it helps (human eval). prob compare with no beam search
    %     - [ ] GENERATE: our stories, no planning system with prompt (rolling window after initial infer attributes string)
    %     - [ ] GENERATE: our stories, no editing system whatsover (no revision, no editing)
    %     - [ ] GENERATE: our stories, no consistency editing but with reranking
    %     - [ ] GENERATE: our stories, no reranking but with editing
    %     - [ ] EXPERIMENT: get human ratings for overall, overarching plot coherence (beginning/middle/end), story continuity?, consistency of details
    % - [ ] additional evaluation of the structured editor system
    %     - [ ] discuss simple baseline approaches for editing
    %     - [ ] discuss construction of our dataset for evaluating this automatically (dev + test), kind of tailored toward our system just as a check that it's working as claimed, show examples
    %         - [ ] CREATE: test dataset for this
    %     - [ ] show our proposed system vs simple baselines, emphasize a lot of room for improvement here and our dataset hopefully can facilitate further improvement
    %         - [ ] EXPERIMENT: our consistency system on test consistency dataset
    %         - [ ] EXPERIMENT: simple entailment baseline system on test consistency dataset

    %         example stories / passages (including examples where the editor fixed something)
\section{Discussion}

We have considered the problem of automatically generating longer stories, proposing the \oursabstract{} framework 
as an initial attempt at addressing the challenges of maintaining long-range coherence and premise relevance. 
Our \ours{} implementation exhibits strong performance on these metrics while generating stories over 2000 words long. 

At its core, \oursabstract{} is a system for emulating the human writing process for long-form generation while leveraging only general-purpose language models in the generation procedure. Thus concepts from \oursabstract{} can potentially be adapted to non-story domains as well, especially the idea of dynamically re-injecting contextual information into a prompt. Moreover, should human interaction be desired, \oursabstract{} is in principle highly controllable: most modules operate almost entirely in natural language.
% TODO summarize why we're great, discuss applications to other domains, controllability

Nonetheless, our main goal remains to further improve automatic long-form story generation. While \ours{}'s stories are an order of magnitude longer than those from prior work,
most humans would still consider them to be ``short stories''---and on the shorter side at that. Our long term goal is to generate interesting, long-range-coherent stories 
of greater length---perhaps what humans might call ``novellas''---and eventually full-length novels. One step in this direction
could be to extend \oursabstract{} using multiple levels of hierarchical outline generation to obtain a much more detailed initial plan, as we do in Appendix \ref{appendix:longer_story} to generate a 7500-word story. 

In our view, the greatest barrier to further increasing story length is evaluation, which frustrates efforts to benchmark systems during both test time and development.
In this work, we have compared \ours{} to baselines solely through human evaluation, 
%Due to the length of the stories which we ask our non-expert annotators to read, some of the annotated metrics
%are quite noisy. 
which can be both noisy as well as costly even with non-expert annotators. While prior works have proposed some possible measures~\cite{barzilay2008modeling,castricato2021towards}, 
we hope that analyzing our generated stories (both \ours{} and baselines) can inspire further research on metrics for which we currently rely solely on human annotation. For example, while there exist reasonable metrics for text similarity on a sentence or paragraph level, long-form generation could benefit from metrics detecting when a longer passage begins on-topic but slowly veers off-topic, or when a passage uses on-topic vocabulary but is otherwise nonsensical in context. Similarly, improved metrics for \textit{long-range} factual contradictions could greatly aid efforts to improve generations' factual consistency, such as our \edit{} module. Even if new metrics do not completely replace human annotations, they could help us both to evaluate longer stories as well as conduct more detailed ablation studies with larger sample sizes.

Additionally, while \ours{}'s stories are relatively plot-coherent and faithful to the premise, substantial gaps remain along other axes compared to even beginner human writers. One such axis is long-range factual continuity: while we believe our structured detection-correction method is a human-like approach, our current \edit{} module is certainly not human-level. Moreover, human stories exhibit long-range continuity along many axes other than just factual attributes of characters, such as overall theme; scenes and world setting; pace and tempo of storylines; and 
%delicate design of hints and story hooks 
foreshadowing
before major events. It remains highly nontrivial to incorporate such considerations into automatic story generation.

\section*{Limitations}

The difficulty of evaluating long-form generation greatly constrains our experiments. Specifically, we are limited in the sample sizes of all our experiments as well as our ability to run more detailed ablations. Improved evaluation would also enable us to evaluate stories much longer than the current 2000-2500 words: while \oursabstract{} is capable of generating such stories (Appendix \ref{appendix:longer_story}), we do not formally evaluate them in this work. Note that compared to evaluation costs, the API costs associated with the actual story generation are significantly lesser. 

The difficulty of careful evaluation also affected system development. Many system design choices (e.g., prompt design, reranking heuristics) and hyperparameters (e.g., length of each story continuation, thresholds for checking contradiction in the \edit{} module) are simply selected manually, rather than chosen based on careful validation. Thus it is likely that substantial room for improvement remains in the detailed design of our individual modules. 

Many of our modules are custom-designed for story generation, especially the structured attribute-value dictionary for story characters used in the \edit{} module. Adaptation to a generation domain other than stories, at least in our current setup, may also require manually re-designing prompts and experimenting with parameters.

% Similarly, evaluation constraints limit the sample sizes in all of our experiments as well as our ability to run more detailed ablations. Improved evaluation would also enable us to evaluate stories much longer than the current 2000-2500 words: while \oursabstract{} is capable of generating such stories (Appendix \ref{appendix:longer_story}), we do not formally evaluate them in this work.

Additionally, there remains substantial room for improvement in our \edit{} module. While we believe that a structured detection and correction system such as our \edit{} module is a principled way to address the important problem of long-range factual continuity, empirically our current implementation does not improve our main metrics (Table \ref{tab:ablations}). Even in the controlled setting where it outperforms our baselines (Table \ref{tab:consistency_detection}), the absolute ROC-AUC score remains low. Moreover, it is designed to handle specifically contradictions related to character attributes, which we observe are a common but certainly not all-encompassing class of errors. 

Finally, we expect that \oursabstract{}'s performance may decrease in languages which lack strong general-purpose language models such as GPT3. 

% TODO: API costs, evaluation, editing system doesn't work that well yet, english only due to large LMs
\section*{Acknowledgements}

We thank the Berkeley NLP group and our anonymous reviewers for their helpful feedback which helped us to greatly improve the paper.  This work was
supported by Berkeley AI Research, Meta AI, Open Philanthropy, DARPA 
under the SemaFor program (HR00112020054), the Machine
Common Sense (MCS) program under Cooperative
Agreement N66001-19-2-4032, and the NSF through
a fellowship to the first author. The content does
not necessarily reflect the position or the policy
of the government, and no official endorsement
should be inferred.
\section*{Ethics Statement}

Strong natural language generation systems present opportunities for abuse, for example in fake news generation. We have attempted to mitigate this issue by focusing on the comparatively innocuous task of story generation. Additionally, in our \edit{} module we have explored methods for maintaining long-range factual consistency as a way to safeguard against model hallucination, and we envision that our \edit{} module could be adapted to incorporate a real-world knowledge base as needed to aid truthful generation. 

Our system relies heavily on pretrained general-purpose language models, specifically GPT3 in our implementation, and thus may inherit the problematic biases associated with such models~\cite{radford2019language,brown2020language,lucy2021gender}. These biases may be amplified in stories, which could negatively affect human readers. However, our overall framework \oursabstract{} is not necessarily tied to GPT3, and can in principle function with any other general-purpose language model. Thus, improvements in debiasing language models can translate into our \oursabstract{} framework as well. Additionally, one could apply controlled generation approaches~\cite{dathathri2019plug,krause2020gedi,yang2021fudge} for debiasing text to our generation procedure.

Finally, as mentioned in Limitations, \oursabstract{}'s performance is tied to the quality of the base language model used as a generator, and thus may suffer on non-English languages. 

\bibliography{anthology,custom}
\bibliographystyle{acl_natbib}

\newpage

\appendix

% \onecolumn

\section{Character Name Generation}\label{appendix:character_names}

We elaborate on our name generation scheme used in the \plan{} module (Section \ref{sec:setup_outline}).

Names are generated by GPT3-Instruct-175B with a prompt consisting of the premise, setting, and any previous character descriptions, shown in Table \ref{tab:name_prompt}. Thus if e.g., a name already appears in the premise, it can be easily copied. After each name, we generate the corresponding description, and both the name and description are appended to the prompt before generating the next name.

\begin{table}[!htbp]
\small
\begin{tabularx}{\linewidth}{X}
\toprule
\texttt{Premise: Cathy is a high school student who is trying to figure out her future. She has been diagnosed with a rare disease that will cause her to slowly go blind. As she tries to make the most of her remaining sight, she also must come to terms with the fact that she may never be able to see again.}\\\\
\texttt{Setting: The story is set in a small town in the United States.}
\\
\\
\texttt{List the names and details of all major characters.}
\\
\\
\texttt{1.}
\\
\\
\texttt{Full Name:}
\\
\bottomrule
\caption{An example prompt used when generating the first name in the \plan{} module.}
\label{tab:name_prompt}
\end{tabularx}
\end{table}

To ensure that we sample reasonable names, we use several heuristics as follows. Each time we generate a name, we sample 10 names in total, and filter out those containing any of a fixed set of strings which we observed were problematic (e.g., story roles like ``protagonist,'' or character attributes like ``age'' and ``gender'' which are not names). We additionally filter out strings with punctuation and strings not in the premise but which appear multiple times in the 10 generated strings (to add more diversity to the names). Finally, we prefer names with two words in them in an effort to get characters' full family names. 

While these simple heuristics are sufficient for this work, there remains ample room for improvement both in generated names' quality (avoiding the occasional edge cases which escape our heuristic filters) as well as fairness (by using a generation system which is perhaps less biased than GPT3). 
\newpage

\section{Details on Additional Reranking Heuristics}\label{appendix:heuristics}

We elaborate on the details of the additional filtering heuristics used in our \rewrite{} system (Section \ref{sec:reranker}). There are a few broad categories of problems which we aim to largely filter out with simple heuristics.

First, we filter out any empty outputs. 

Second, we aim to reduce repetition in the generation both within itself and with the prompt. We simply check for repeated sequences of 5 words or more, and also check if the edit distance between any two sentences is a sufficiently small fraction of their length. 

Third, we aim to avoid jarring changes in narration. For example, this can result from the GPT3 generator reverting to the style of the prompt, with e.g., headings for story commentary or author notes. Thus we filter out any generations containing any of a fixed list of strings, such as ``\textbackslash nComment'' and ``copyright''. For some strings which may reasonably appear in a normal story passage as well, we filter out passages if two or more appear. We also filter out generations where any paragraph contains a colon within the first few words (a likely indicator of an analysis header). 

Fourth, we aim to maintain consistent third person narration, so we detect whether a continuation is written in first or second person by searching for the presence of ``I,'' ``we,'' and ``you'' outside of quotations and filter out such continuations. 

\newpage

\section{Details on Editing System Information Extraction}\label{appendix:editing_details}

As discussed in Section \ref{sec:editor}, the core subroutine of our \edit{} module's detection system is an information extraction system for gathering structured information about a given character from a newly generated passage. We will illustrate this process using a running example taken mid-generation for a story starting from the plan shown in Table \ref{tab:editor_example_setup}.

\begin{table}[!htbp]
\small
\begin{tabularx}{\linewidth}{X}
\toprule
\texttt{Premise: In a future world where the sun has gone out, a group of people huddle around a fire in a small cabin. They are waiting for a message from the outside that will tell them what to do next.}
\\
\\
\texttt{Setting: The story is set in a dark cabin lit only by a fire.}
\\
\\
\texttt{Characters:}
\\
\texttt{1. Karen Zellerion is a strong and determined woman. She is the leader of the group and is always looking for ways to help her people.}
\\
\texttt{2. Luke Zellerion is Karen's husband and the second-in-command of the group. He is a skilled hunter and often uses his knowledge to help the others.}
\\
\texttt{3. Maria Zellerion is Karen and Luke's daughter. She is a bright and curious girl who is always asking questions about the world they live in.}
\\
\\
\texttt{Outline:}
\\
\texttt{1. The group receives a message from the outside that tells them to go to a certain location.}
\\
\texttt{2. The group sets out on their journey, encountering various challenges along the way.}
\\
\texttt{3. The group arrives at the location and discovers what they are supposed to do next.}
\\
\bottomrule
\caption{The plan generated by the \plan{} module for our running example illustrating the \edit{} module's attribute-value detection procedure.}
\label{tab:editor_example_setup}
\end{tabularx}
\end{table}

We begin by feeding GPT3-Instruct-175B a prompt containing the passage, the name of the character, and a request to list facts about the character, as shown in Table \ref{tab:editor_prompt1}. We generate 3 outputs, parse the lists into individual facts, and retain those facts which are agreed upon by at least 2 of the outputs (according to an entailment model) to reduce hallucination. 

\begin{table}[h]
\small
\begin{tabularx}{\linewidth}{X}
\toprule
\texttt{"I don't know," answered Maria, "but I'll go with you if you're going to meet someone." so they left together.}
\\
\\
\texttt{They walked down to the main part of the community where everyone else was gathering and found an older woman there named Lucy that Karen knew.}
\\
\\
\texttt{"You know what's going on don't you?" Karen asked Lucy.}
\\
\\
\texttt{"Yes, I do but I don't think I can tell you much because I've been sworn to secrecy," she replied. "I just have a feeling that if we're going to get any answers from this person we're looking for we're going to have to see them face-to-face," assured Lucy. They all agreed that this was the best way to do things and went looking for their neighbor Lizzy who had a car that worked. They met her again in the cafeteria and told her about their plans and Lizzy said she would like nothing better than a little adventure so she agreed to take them as long as no one got hurt along the way.}
\\
\\
\texttt{Question: List very brief facts about Lucy's appearance, personality, and relationship to other characters.}
\\
\\
\texttt{1. Lucy \hl{is an older woman.}}
\\
\\
\texttt{\hl{2. Lucy is sworn to secrecy.}}
\\
\\
\texttt{\hl{3. Lucy is a good friend of Karen's.}}
\\
\bottomrule
\caption{An example prompt for listing initial facts about a given character based on a newly written passage, used in the \edit{} module's detection procedure. We show one of the three generated continuations in highlighting. (Note that Lucy was not one of the original three characters generated by the \plan{} module, but rather was detected and added to our knowledge base over the course of generation as discussed in Section \ref{sec:prompt}.)}
\label{tab:editor_prompt1}
\end{tabularx}
\end{table}
%[" is an older woman.\n2. Lucy is sworn to secrecy.\n3. Lucy is a neighbor of Lizzy's.", " is an older woman.\n2. Lucy is sworn to secrecy.\n3. Lucy is a good friend of Karen's.", ' is an older woman.\n2. Lucy is knowledgeable about the situation with the person they are looking for.\n3. Lucy is sworn to secrecy.\n4. Lucy has a good relationship with Karen.']

Next, we extract the attribute keys from each fact. This is done via a few-shot prompt to GPT3-Instruct-13B, selecting examples based on DPR relevance from a small collection of about 80 handwritten examples, as shown in Table \ref{tab:editor_prompt2}. Note we do not keep the attribute \textit{values} generated in this step as we observed frequent hallucination. Additionally, we filter out any attribute keys which return either no answer or a sufficiently low-confidence result from a T5-large-based UnifiedQA question answering model~\cite{khashabi2020unifiedqa} when given either the fact or original passage as context. 

\begin{table}[h]
\small
\begin{tabularx}{\linewidth}{X}
\toprule
\texttt{Extract attributes from the given context using the format Attribute: Value.}
\\
\\
\texttt{----}
\\
\texttt{Context (Nora Johnson): Selma Vincenti is Nora's friend who recently got engaged to Bill.}
\\
\texttt{Nora Johnson's friend's name is Selma Vincenti}
\\
\texttt{Nora Johnson is Selma's friend}
\\
\texttt{----}
\\
\texttt{Context (Shannon): Kathleen O'Brien is Shannon's mother.}
\\
\texttt{Shannon's mother's name is Kathleen O'Brien}
\\
\texttt{Shannon is Kathleen's daughter}
\\
\texttt{----}
\\
\texttt{Context (Rachel Kim): Rachel Kim's father loves her children dearly.}
\\
\texttt{Rachel Kim's gender is female}
\\
\texttt{----}
\\
\texttt{Context (Johnny): Johnny is a friendly and outgoing person, and he loves spending time with his sister Mira.}
\\
\texttt{Johnny's gender is male}
\\
\texttt{Johnny's sister's name is Mira}
\\
\texttt{Johnny is Mira's brother}
\\
\texttt{----}
\\
\texttt{Context (Tina Palmer): Tina Palmer befriends Amy Sinkhorn.}
\\
\texttt{Tina Palmer is Amy's friend}
\\
\texttt{Tina Palmer's friend's name is Amy Sinkhorn}
\\
\texttt{----}
\\
\texttt{Context (Lucy): Lucy is a good friend of Karen's.}
\\
\texttt{Lucy \hl{is Karen's friend}}
\\
\texttt{\hl{Lucy is a good friend of Karen}}
\\
\texttt{\hl{Karen is Lucy's friend}}
\\
\bottomrule
\caption{An example prompt for extracting attributes from a natural language fact (``Lucy is a good friend of Karen's.'') in the \edit{} module. Attribute key-value pairs are extracted from each generated line in a rule-based manner, and we discard outputs for which our rule-based parser fails (both the second and third output lines in this case). After extraction, we keep only the key, while the value is discarded due to a high rate of hallucination in this step; we regenerate it later.}
\label{tab:editor_prompt2}
\end{tabularx}
\end{table}
%[" is Karen's friend\nLucy is Karen's friend's friend\n\nValue: female.", " is Karen's friend\nLucy is a good friend of Karen\nKaren is Lucy's friend", " is Karen's friend\nLucy is a good friend of Karen's"]

\FloatBarrier

To recompute the attribute values, we prompt GPT3-Instruct-13B with the original fact, character name, and attribute key as shown in Table \ref{tab:editor_prompt3}, and take the most agreed upon of 3 outputs as the attribute value. We filter out any key-value pairs which are not entailed with sufficiently high probability by the original fact from which they were extracted.

\FloatBarrier

\begin{table}[h]
\small
\begin{tabularx}{\linewidth}{X}
\toprule
\texttt{Lucy is a good friend of Karen's.}
\\\\
\texttt{Lucy is Karen's \hl{friend.}}\\
\bottomrule
\caption{An example prompt for extracting values after identifying attribute keys in the \edit{} module. In this case, the character for which we are inferring is Lucy, and the attribute key is ``Karen's.''}
\label{tab:editor_prompt3}
\end{tabularx}
\end{table}

\FloatBarrier

After acquiring key-value pairs, we need to update the structured attribute dictionary for the given character. When we detect a conflict (i.e., an attribute key is already present in the dictionary), we compare the new and old attribute values using an entailment model by converting the attribute-value pairs into simple sentences in a rule-based manner (e.g., ``gender: female'' in Karen's dictionary will convert to ``Karen's gender is female.''). If one attribute value entails the other, then we keep the former as the attribute value. If there is a neutral relation, we make no change. If there is a contradiction, we flag it for editing. 

Lastly, we can ``complete'' attributes involving other characters in the dictionary. For example, if Ben's teacher is Anna, GPT3-Instruct-175B can infer that Anna's student is Ben, and add this relation to our dictionary for Anna. Additionally, we can infer that Anna's relationship to Ben is ``teacher'' and that Ben's relationship to Anna is ``student.'' An example of this procedure is shown in Table \ref{tab:editor_prompt4}.

\FloatBarrier

\begin{table}[h]
\small
\begin{tabularx}{\linewidth}{X}
\toprule
\texttt{Lucy is Karen Zellerion's friend.} 
\\\\
\texttt{Karen Zellerion is Lucy's \hl{friend}.}\\
\bottomrule
\caption{Example prompt for ``completing'' attributes involving other characters in the \edit{} module. Note that we automatically matched ``Karen'' to our existing character ``Karen Zellerion.'' From the initial fact that Lucy is Karen's friend, we infer that Karen is Lucy's friend, that Lucy's friend is Karen, and Karen's friend is Lucy. (This example also hints at one limitation of our current system, namely, that it implicitly assumes one value per attribute: e.g., if Lucy had a second friend it would flag a contradiction.)}
\label{tab:editor_prompt4}
\end{tabularx}
\end{table}

\FloatBarrier

For the controlled setting evaluation in Section \ref{sec:editor_analysis}, we modify the system to output continuous probabilities of contradiction (to compute a ROC-AUC score) rather than discrete decisions on whether a previously detected attribute is contradicted. Thus for each passage, we simply return the entailment model's maximum probability of contradiction observed across all attribute key conflicts. 

\begin{table*}[htb]
\small
\centering
\begin{tabular}{llcc}
\toprule
\textbf{Model} & \textbf{API Endpoint}    & \textbf{Average Calls} & \textbf{Average Tokens} \\
\midrule
GPT3-175B & \texttt{davinci}  & \phantom{0}12.0 & 34510.0 \\
GPT3-Instruct-175B & \texttt{text-davinci-002} & \phantom{0}70.2 & 25558.0 \\
GPT3 Edit API & \texttt{text-davinci-edit-001} & \phantom{00}7.0 & 19425.2 \\
GPT3-Instruct-13B & \texttt{text-curie-001} & 362.6 & 48401.8 \\
\bottomrule
\end{tabular}
\caption{\small For each API endpoint that we use, we report the average number of API calls and tokens processed per story generated by \ours{}. Note that for the Edit API, we simply add the total number of tokens in both prompt and output when calculating the number of tokens, although it is not obvious if this is the appropriate count. Calls to the Insert API are included under \texttt{text-davinci-002}.}
\label{tab:api_usage}
\end{table*}

\section{Data on API Usage}\label{appendix:api_usage}

In Table \ref{tab:api_usage}, we report the average number of API calls and number of tokens processed (including both prompts and generations) for each GPT3 API endpoint across 5 runs of \ours{}, using the same settings as in our main experiments.

The large number of tokens generated from GPT3-175B and GPT3-Instruct-175B can be attributed to our filtering and reranking in the \plan{} and \rewrite{} modules; typically we generate 10 outputs per call. The \edit{} module is responsible for most of the GPT3-Instruct-13B usage as well as some of the GPT3-Instruct-175B usage. Finally, the \edit{} module is naturally the sole user of the Edit API, which also involves rejection sampling when the API either makes no change or returns an overly lengthy response. 

The total cost for generating a single \ours{} story with these settings adds up to a few dollars. The baselines and ablations require fewer calls than reported here.

\begin{table*}[htbp]
\small
\centering
\begin{tabular}{lccccc}
\toprule
\textbf{Method}          & \textbf{Interesting }$\uparrow$ & \textbf{Coherent} $\uparrow$ & \textbf{Relevant} $\uparrow$ & \textbf{Humanlike} $\uparrow$ & \textbf{Misc.\ Problems} $\downarrow$ \\
\midrule
\oursshortlength{}         & 44.7&	47.3	&59.3&	89.3&	\textbf{1.29}           \\
\ours{}         &  52.0	&56.0&	62.0&	87.3&	1.45           \\

\midrule

\ourslonglength{} & \textbf{64.0}	&60.0&	58.0&	85.3&	1.77          \\
\ours{}            & 42.0	&51.3&	58.0	&82.0	&1.68         \\
\bottomrule
\end{tabular}
\caption{\small Comparison of \ours{} against versions generating shorter and longer stories (\oursshortlength{} and \ourslonglength{} respectively). The first two rows show a pairwise comparison between \oursshortlength{} and \ours{} and the last two rows show the equivalent comparison between \ourslonglength{} and \ours{}. Bolding indicates significant differences with $p<0.05$ on a paired $t$-test. In most metrics the differences are insignificant.}
\label{tab:length_analysis}
\vspace{-0.5em}
\end{table*}

\section{Dataset Usage}

The only preexisting story dataset used in this work is the WritingPrompts dataset~\cite{fan2018hierarchical}, which is used to train our relevance and coherence rerankers (and the generator for the \rollingfinetune{} baseline). GPT3 is additionally used to derive summaries of WritingPrompts passages for training the relevance reranker. Finally, we generated some examples of contradictory story setups and story beginnings when analyzing our \edit{} module in Section \ref{sec:editor_analysis}, which relied solely on prompting GPT3, and not any preexisting dataset. 

All data used or generated for this paper, together with documentation, can be found through our codebase located at \url{https://github.com/yangkevin2/emnlp22-re3-story-generation}.

\newpage

\section{Length vs. Story Quality Analysis}\label{appendix:length_analysis}

In our main experiments, we ran \ours{} with three outline sections and generated four 256-token passages per outline section. Here, experiment with generating from \ours{} using the same outlines, but with two or six 256-token passages per outline section instead. We refer to these modified version of \ours{} as \oursshortlength{} and \ourslonglength{} respectively. The results are shown in Table \ref{tab:length_analysis}.

For the most part, the sample size of 50 stories for this comparison proved insufficient to draw clear quantitative conclusions on the impact of length on \ours{} story quality. However, interestingly, annotators judged the longer stories to be more interesting. Additionally, it seems intuitive that longer stories are more likely to suffer the presence of writing problems at some point in the story simply due to having more total text. 

Qualitatively, we also observe that the generator may become repetitive or lose the plot thread over longer time horizons, but ending generation too early can also yield stories which seem “truncated” before they reach the main plot points. Trying to balance these factors by determining the length of story passages more dynamically could be an interesting avenue for future research.

\newpage

\section{Full Metrics for Miscellaneous Writing Problems}\label{appendix:writing_problems_metrics}

\begin{table*}[htb]
\small
\centering
\begin{tabular}{lcccccc}
\toprule
\textbf{Method}                              & Narration $\downarrow$ & Inconsistent $\downarrow$ & Confusing $\downarrow$ & Repetitive $\downarrow$ & Disfluent $\downarrow$ &\textbf{Misc. Problems} $\downarrow$\\
\midrule
\ours{}            & 0.15      & 0.27         & 0.24      & 0.3        & 0.11 & \textbf{1.07}    \\
\rolling{}         & 0.2       & 0.28         & 0.3       & 0.29       & 0.13  & 1.2   \\
\midrule
\ours{}           & 0.21      & 0.35         & \textbf{0.29}      & 0.3        & 0.2 &  \textbf{1.35}      \\
\rollingfinetune{} & 0.24      & 0.32         & 0.37      & 0.31       & 0.23  & 1.48    \\

\bottomrule
\end{tabular}
\caption{\small Fraction of stories marked with individual writing problems from pairwise comparison of \ours{} against two baselines, \rolling{} and \rollingfinetune{}. Bolding indicates significant differences with $p<0.05$. Differences in individual problems are largely not significant, but they become significant in aggregate (Misc. Problems)}
\label{tab:writing_problems_baselines}
\end{table*}

\begin{table*}[htb]
\small
\centering
\begin{tabular}{lcccccc}
\toprule
\textbf{Method}                              & Narration $\downarrow$ & Inconsistent $\downarrow$ & Confusing $\downarrow$ & Repetitive $\downarrow$ & Disfluent $\downarrow$ &\textbf{Misc. Problems} $\downarrow$\\
\midrule
\ours{}      & 0.23      & 0.31         & 0.31      & 0.25       & 0.15      & 1.25          \\
\noplanner{} & 0.29      & 0.32         & 0.34      & 0.21       & 0.18      & 1.33          \\
\midrule
\ours{}      & \textbf{0.26}      & 0.35         & \textbf{0.25}      & 0.17       & 0.15      & \textbf{1.17}          \\
\norerank{}  & 0.43      & 0.34         & 0.38      & 0.14       & 0.18      & 1.48          \\
\midrule
\ours{}      & 0.23      & 0.29         & \textbf{0.22 }     & 0.25       & 0.14      & 1.12          \\
\noeditor{}  & 0.19      & 0.26         & 0.28      & 0.25       & 0.12      & 1.1           \\
\bottomrule
\end{tabular}
\caption{\small Fraction of stories marked with individual writing problems from pairwise comparison of \ours{} against ablations which remove the \plan{}, \rewrite{}, and \edit{} modules respectively. Bolding indicates significant differences with $p<0.05$. Differences in individual problems are largely not significant.}
\label{tab:writing_problems_ablations}
\end{table*}

We show the metrics for individual writing problems as described in Section \ref{sec:evaluation}. Tables \ref{tab:writing_problems_baselines} and \ref{tab:writing_problems_ablations} show the results for the main baselines and ablations respectively. The differences in individual metrics are largely not significant (although \ours{} is never significantly worse), but in many cases become significant when taken in aggregate.

\FloatBarrier

\section{Mechanical Turk Evaluation Details}\label{appendix:mturk_survey}

\begin{figure*}[htp]

\subfloat{
  \includegraphics[clip,width=\textwidth]{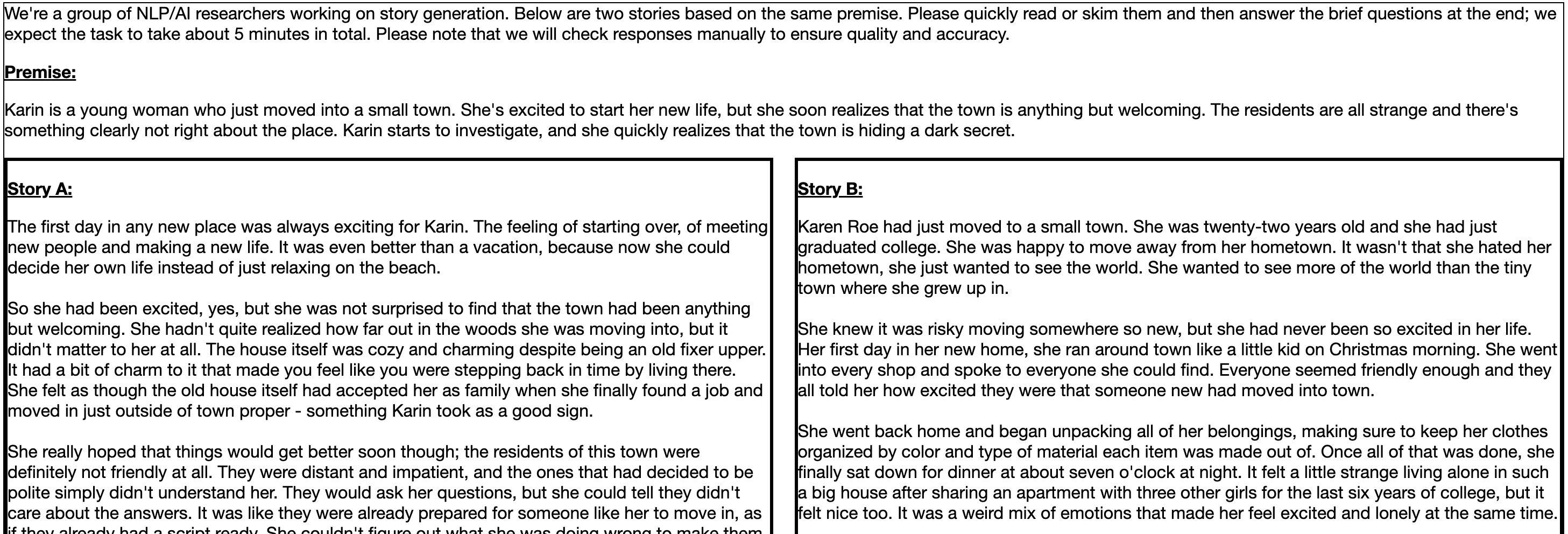}
}

\subfloat{
  \includegraphics[clip,width=\textwidth]{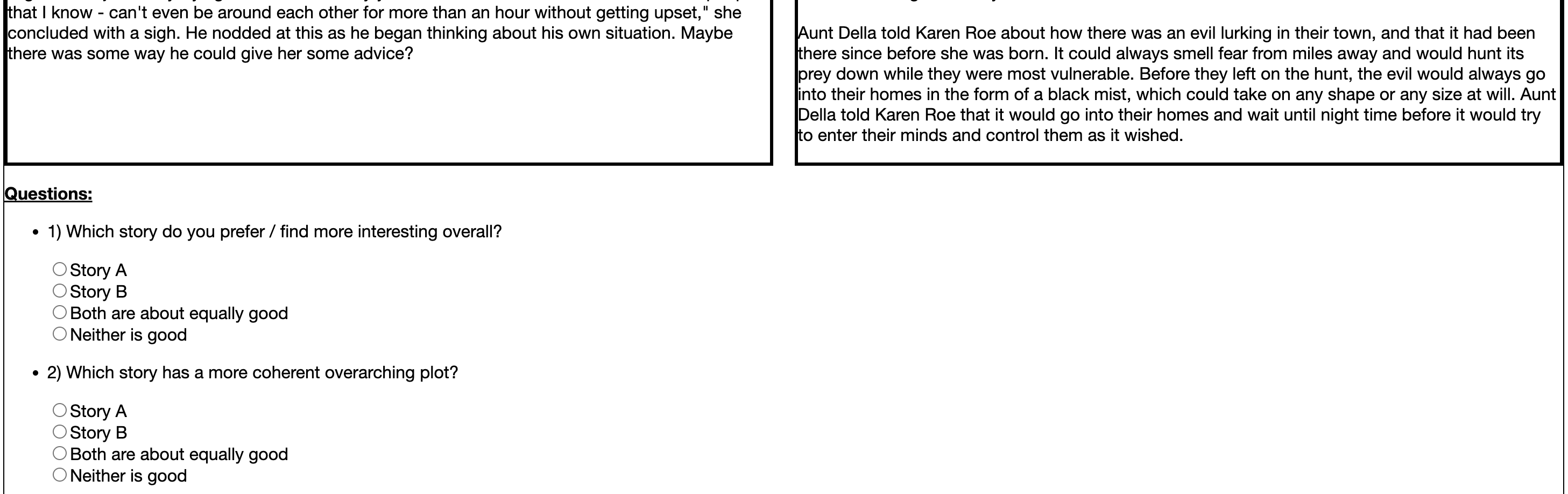}
}

\subfloat{
  \includegraphics[clip,width=\textwidth]{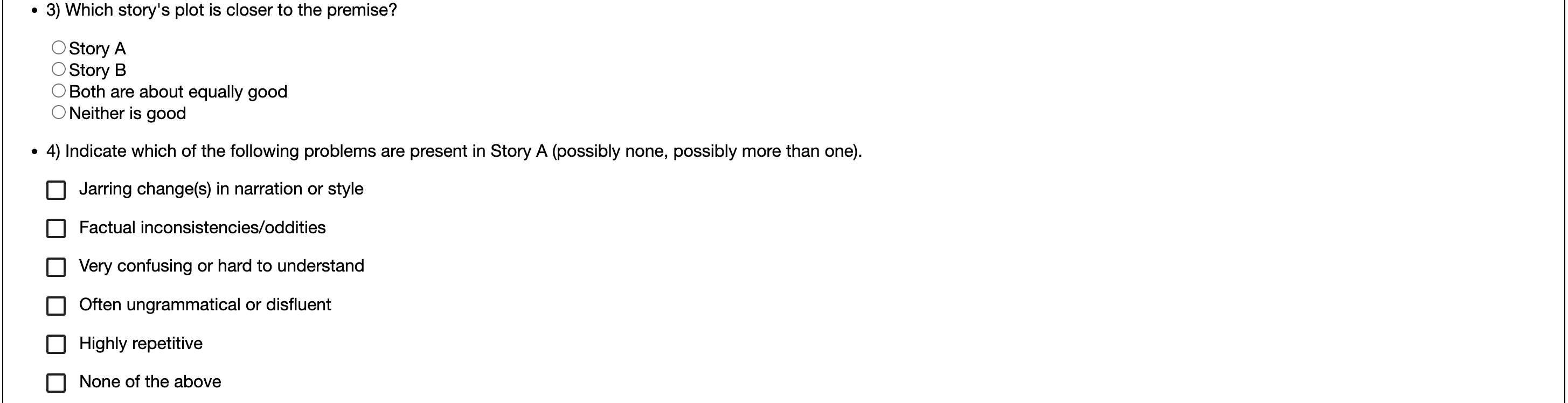}
}

\subfloat{
  \includegraphics[clip,width=\textwidth]{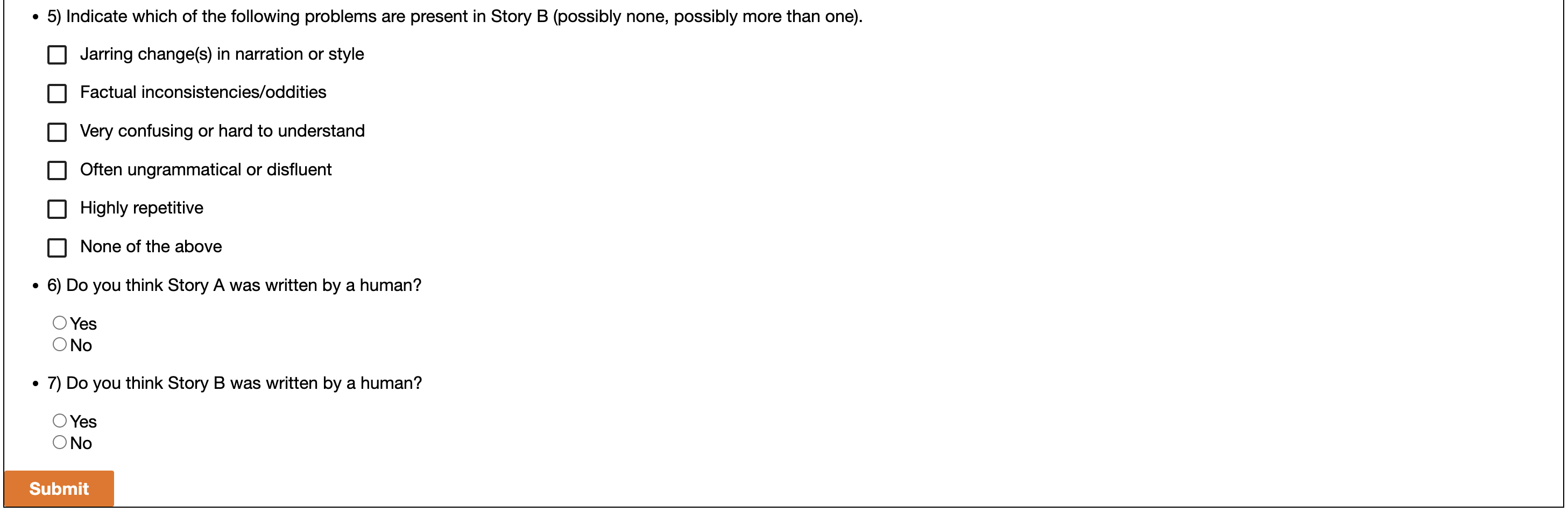}
}

\caption{Example of a Mechanical Turk survey from our evaluation. The actual stories are mostly omitted as we are simply showing the format of the survey.}
\label{fig:mturk}
\end{figure*}

In Figure \ref{fig:mturk} we show an example Mechanical Turk survey from our evaluation in which the annotator is asked to answer questions comparing two stories. Workers were paid \$1.50 per hit. 

% \FloatBarrier
% \newpage

\begin{table*}[htb]
\small
\centering
\begin{tabular}{lcccccc}
\toprule
\textbf{Method} & \textbf{Interesting} & \textbf{Coherent} & \textbf{Relevant} & \textbf{Humanlike} & \textbf{Misc Problems} \\
\midrule
\ours{} vs. \rolling{}         & 0.20                  & 0.07              & 0.05              & \phantom{-}0.06               & \phantom{-}0.04                   \\
\ours{} vs. \rollingfinetune{} & 0.04                 & 0.04              & 0.09              & -0.05              & -0.03     \\
\bottomrule
\end{tabular}
\caption{\small Fleiss' kappa for agreement on individual metric annotations in pairwise comparisons between \ours{} and baselines. Overall the agreement is relatively poor.}
\label{tab:annotator_agreement_baselines}
\end{table*}

\begin{table*}[htb]
\small
\centering
\begin{tabular}{lcccccc}
\toprule
\textbf{Method} & \textbf{Interesting} & \textbf{Coherent} & \textbf{Relevant} & \textbf{Humanlike} & \textbf{Misc Problems} \\
\midrule
\ours{} vs. \noplanner{} & -0.01       & \phantom{-}0.08     & 0.12     & 0.07      & -0.04         \\
\ours{} vs. \norerank{}  & \phantom{-}0.05        & \phantom{-}0.03     & 0.06     & 0.06      & \phantom{-}0.00             \\
\ours{} vs. \noeditor{}  & \phantom{-}0.05        & -0.03    & 0.07     & 0.07      & \phantom{-}0.02  \\
\bottomrule
\end{tabular}
\caption{\small Fleiss' kappa for agreement on individual metric annotations in pairwise comparisons between \ours{} and ablations. Overall the agreement is relatively poor.}
\label{tab:annotator_agreement_ablations}
\end{table*}

\newpage\phantom{none}
\FloatBarrier

\newpage

\section{Annotator Agreement}\label{appendix:annotator_agreement}

The evaluation task (Appendix \ref{appendix:mturk_survey}) asks annotators to ``quickly read or skim'' two fairly lengthy passages in order to be able to evaluate more stories. Thus many details may be missed. Moreover, many of our metrics are by nature rather subjective. Thus it is expected that individual labels may be highly noisy, resulting in poor annotator agreement. While we expect that agreement would be better with expert annotators, this would significantly increase the cost burden. 

Indeed the agreement as measured by Fleiss' kappa, while usually positive, is low on most of our comparisons (Tables \ref{tab:annotator_agreement_baselines} and \ref{tab:annotator_agreement_ablations}). 

\newpage

\section{Example Stories}\label{appendix:examples}

Here we show the stories generated by \ours{} and the \rolling{} and \rollingfinetune{} baselines on the first five premises in our main evaluation, i.e., the examples are i.i.d.\ and non-cherry-picked. We note that even with the same premise, there are of course many possible stories to be written, and re-generating could easily result in a completely different story with different strong points and problems. Nevertheless, we show these examples to provide a sense of the overall level of quality as well as to illustrate some types of problems which may arise. 

We individually summarize what we view as the strengths and weaknesses of each generated story in the caption at the bottom of each example to facilitate faster reading, while also describing general qualitative trends here. Unsurprisingly, both \rolling{} and \rollingfinetune{} exhibit a common failure mode in which they generate a rambling series of coherent short passages which are largely irrelevant to the premise and even to each other. In contrast, although there are some confusing passages mixed in, \ours{}'s stories of the same length generally maintain a coherent overarching plot for most of the generation. Moreover, our stories' overarching plots are clearly relevant to the initial premise, although they may deviate from the detailed outline which we generate in our \plan{} module.

Some of the problems in \ours{} may be partially due to the concessions we make due to evaluation costs and limitations (Section \ref{sec:evaluation}, Method Instantiation). We direct the reader to Appendix \ref{appendix:longer_story} for an example of a much longer (7500-word) story generated without these limitations, which we believe better represents the ceiling of what our \oursabstract{} framework is capable of.

\newpage

\onecolumn
\subsection{Examples for Premise 1}

\begin{table}[!htbp]
\small
\begin{tabularx}{\linewidth}{X}
\toprule
\textbf{Premise 1}\\
\midrule
\texttt{A woman discovers that her husband has been cheating on her and decides to get revenge.}\\
\bottomrule
\caption{The premise for the first group of example generated stories. A very generic prompt overall. }
\label{tab:premise1}
\end{tabularx}
\end{table}

\begin{table}[!htbp]
\small
\begin{tabularx}{\linewidth}{X}
\toprule
\textbf{Initial Setup and Outline for \ours{}, Premise 1}\\
\midrule
\texttt{Premise: A woman discovers that her husband has been cheating on her and decides to get revenge.}\\
\\
\texttt{Setting: The story is set in the present day, in a small town.}\\
\\
\texttt{Characters:} \\
\texttt{1. Diane Chambers is a beautiful woman in her early thirties. She has long dark hair and blue eyes. She is married to Mark Chambers and is a stay-at-home mom.}\\
\texttt{2. Mark Chambers is a successful businessman in his early forties. He is tall and handsome, with brown hair and green eyes. He is married to Diane Chambers and is the father of two young children.}\\
\texttt{3. Karen Johnson is a beautiful blonde women in her early thirties. She is Mark Chambers' mistress and is also married with two young children.}\\
\\
\texttt{Outline: }\\
\texttt{1. Diane Chambers discovers that her husband has been cheating on her with another woman.}\\
\texttt{2. Diane Chambers decides to take revenge on her husband by having an affair of her own.}\\
\texttt{3. Diane Chambers' affair is discovered by her husband, leading to a confrontation between the two.}\\
\bottomrule
\label{tab:premise1_ourssetup}
\caption{\ours{}'s initial setup and outline for Premise 1. The characters and outline make perfect sense for the given premise.}
\end{tabularx}
\end{table}

% \afterpage{
% \small
% \begin{xtabular*}{\linewidth}{p{\linewidth}}
\begin{small}
\begin{longtable}{p{0.9\textwidth}}
\toprule
\textbf{\ours{} Story for Premise 1}\\
\midrule
\texttt{"When it rains, it pours," Mark Chambers said to his wife, Diane.}
\\
\\
\texttt{"What's that supposed to mean?" she asked him.}
\\
\\
\texttt{"It means that I've been living a double life for the last five years, and now everyone knows about it," he replied. "I've been cheating on you with Karen Johnson."}
\\
\\
\texttt{Diane stared at her husband long and hard before replying to him. "Karen Johnson? As in the other woman in your office?"}
\\
\\
\texttt{Mark nodded his head in confirmation. He looked at his wife with eyes as appealing as possible, hoping to win her back on her side. "Diane, I'm sorry for what I did to you," he told her. "But this has been going on for five years now."}
\\
\\
\texttt{Diane didn't seem impressed by her husband's apology at all. She sat firmly on the moonlit sofa that stood in the middle of their room and folded her hands neatly across her lap. Her expression was one of disappointment and disgust; but mostly she just seemed sad. "Why are you telling me this now, Mark?" she queried him. "You told me that you were going to Chicago for a business meeting," she said.}
\\
\\
\texttt{Mark could feel his wife's pain as clearly as if it had been his own. He tried to think desperately of how he could tell her everything without using the words 'cheating', 'affair' or 'deception'. But no matter how hard he thought, there was just no other way to say it; after all, that was exactly what he had done. He had been lying to her for five long years and now she knew about it. And it was about time that she did know about it. Any more lies would do more harm than good. So Mark grasped his wife's hand firmly and began to speak to her with a sincere tone in his voice. "Diane," he started, "I lied to you when I said that I was going on a business trip this weekend. The truth is that Karen Johnson is my lover." Diane's eyes widened as she understood what he had just said. "What?" she asked in a soft, low voice.}
\\
\\
\texttt{Mark took his wife's head in his hands and continued to explain, "I've been seeing her for five years now. She is the one I'm supposed to be traveling to Chicago with right now." Diane sank back into the sofa, looking quite shocked and lost. "I'm so sorry," Mark said sincerely.}
\\
\\
\texttt{Diane was still silent for a long moment before she spoke again, this time in a very low voice, almost barely audible. "I knew there was something wrong," she said quietly. "You haven't touched me in months." She looked him directly in the eye then and asked him urgently, "Why didn't you tell me sooner? Why didn't you stop?" But Mark could only smile sadly at her and shake his head from side to side. They both knew that there was no answer for those questions; he had no excuses for what he'd done to her; no explanation that would make it alright to cheat on your wife of ten years with another woman; not even an apology that could ever make up for his behavior. Diane looked away from him then and stared out the window for a moment. She was still in shock. It was painful for her to have to admit that she had known all along that there was something horribly wrong with her marriage, but she had just ignored the signs and never realized it. Mark reached over and took her hand and she turned to look at him again. "I'm sorry," he said again.}
\\
\\
\texttt{Diane squeezed his hand back, though mournfully, for at the moment those two words seemed so inadequate; they seemed like such a terribly little consolation for all that he had put her through. But Mark pressed on: "I am going to do everything I can now to make this right," he told her, "to make sure you never regret marrying me." She tore her eyes away from him then because she couldn't stand to look at him any longer; his love for his mistress was too obvious in his eyes, as he looked into hers.}
\\
\\
\texttt{Diane looked around the living room of their apartment and noticed how different everything looked; how strange it all seemed now that she knew what had been going on since the day they'd moved here so many years ago: ten years ago in fact. She thought about the day they had taken their vows and she realized they were only part of a charade. It was as if everything she'd experienced since then had all been an illusion; it didn't even feel as though they'd ever been real to her.}
\\
\\
\texttt{For a long time, she kept her eyes on the floor and she began to work up a fury inside of herself; for the first time in ten years, anger finally replaced hurt and disappointment in her heart. She felt betrayed, abandoned and sad all at once: but most of all, she was ashamed that she had gotten into this mess by marrying someone who hadn't really loved her.}
\\
\\
\texttt{She felt foolish now for having been so blind for so long. She wondered how on earth he could have deluded her for so long; deluded himself as well as there was no way he could have ever truly believed he loved his mistress either. Nothing about their relationship had felt real: not the way he kissed her nor the way he made love to her nor even the way he looked at her – nothing was real except the fact that he didn't love Diane any longer or if he ever had at all. She felt lost and confused, but she also felt a sense of glee deep within her. She realized that she had been an idiot for so long, but now it was time to do something right. It was time to set things right and now that she'd found out what kind of a man he really was, it was time to get her own back too. It was time to take revenge on him and make him pay for what he had done; she finally had the chance to show him he wasn't the only one who could play games when it came to marriage.}
\\
\\
\texttt{She wanted to hurt him as much as he had hurt her; more than anything else in the world, Diane Chambers wanted to make Mark Chambers suffer for what he had done. She didn't want him anymore: not really. Maybe not ever again, but there was no way she would ever let him be with another woman again either: there were some things never meant for sharing - and that included other women too. It was Saturday afternoon, one week after she'd found out about his affair and she was eager to leave him: she wanted to be as far away from the house as possible and to start the next phase of her life without her husband. After all, it was clear that he no longer wanted her, not really; he didn't love her anymore so it would serve him right if she left him. After all, he had been the one who'd started this whole thing in the first place and now it was his turn to feel like a fool for having hurt Diane again: he had made a laughing stock of them both and deserved every single second of shame and embarrassment. Diane Chambers was done playing games; done pretending that everything was alright when it wasn't ... not at all.}
\\
\\
\texttt{As soon as Diane's children were asleep and Mark realized how serious she was about leaving, he came into the guest room where she lay sobbing in bed alone; he looked at her with despair in his eyes but never said a word. Instead of asking or begging her not to go, Mark simply tried to strangle her - but Diane fought back. She fought hard and neither man nor woman was going to kill or even hurt her ever again. When the fight was over, Diane ordered her husband to leave immediately; she told him that if he didn't go she would call the police. Of course, not a single word of what had actually transpired would be uttered aloud; it wasn't a crime to kill one's wife after all - at least, not in Mark's eyes - but he knew that he wouldn't be able to kill his wife and get away with it. He had to get out of the house before Diane called the cops and told them that he'd tried to kill her... so Mark left.}
\\
\\
\texttt{Diane Chambers had never been so happy in her life as when she heard the front door slam shut behind him. She no longer cared about anything else in this world other than getting her revenge on her husband for betraying their marriage vows, for hurting her once again after everything they'd been through together. She knew that she'd pay for it later with heartache, but she believed that this was well worth any amount of pain or sorrow... anything at all... because ultimately it was only fair that he should suffer just as much as she was suffering now. It was a couple hours later when Mark found himself back on their front doorstep, pounding on the door and begging Diane to let him in. This wasn't how he'd imagined things going, but now that he'd been rejected an opportunity to explain for his actions, he decided to beg for one last chance to make things right between the two of them. He truly loved his wife and wanted to be with her always - but she was having none of it; she grabbed the phone and called her brother.}
\\
\\
\texttt{Mark waited while Diane's brother picked up the phone and heard him say, "Hello?" Then Diane said, "I need you now. Right now." And they both watched while Mark raced down the street toward his car with a look of confusion upon his face.}
\\
\\
\texttt{**********************************************}
\\
\\
\texttt{Meanwhile...}
\\
\\
\texttt{Mark Chambers pulled into the driveway of Doug Johnson's house, unaware that he was being followed by a car containing Doug's wife, Karen Johnson. She'd been waiting for hours for Mark Chambers to come home; she knew that something must have gone wrong because Mark had promised to bring home some unexpected money today. She had waited in the car, parked across from Doug's home, certain that he would soon appear. Her patience was starting to wear thin when finally she saw him drive up to the front door.}
\\
\\
\texttt{She watched as he walked inside and she waited to make sure that Diane was not with him. She left her car and dashed across the street and carefully hung back in a tree, waiting for just the right moment to strike. She couldn't believe how calm she was as she watched him walk inside and reach up to knock on Diane's bedroom door; then she waited until they were both standing outside at the top of the stairs looking down at her before making her approach...}
\\
\\
\texttt{**********************************************}
\\
\\
\texttt{Mark Chambers reached out his hand and tried to pull his wife into his arms, but she pushed him away from her. Tears were streaming down her face as she told Mark that their marriage was over; Mark tried again to reach for Diane but she pushed harder this time, causing him to fall back onto his rear end on the top stair of their staircase. Diane started yelling at Mark, "Do you have any idea what you've done? Do you have any idea what it's like for me now? How can you possibly justify this betrayal of our marriage? What about our children? What about me?"}
\\
\\
\texttt{Diane stopped for a moment to catch her breath and Mark tried to explain but she yelled at him to shut up. Mark apologized, but Diane wouldn't hear it. She asked him if he was ashamed of her and he assured her that there was nothing he could possibly be ashamed of; she explained to him that he had just made an adulterer out of her.}
\\
\\
\texttt{Mark had tears in his eyes as his wife looked down at him, but no words came out of his mouth. He knew that it would not help things any if he said the wrong thing now; even though the way Diane looked right now was better than how she was looking when Mark walked in the door earlier, after seeing Karen Johnson...}
\\
\\
\texttt{After diane chambers caught them in bed together, karen johson and mark chambers were both at his home when diane arrived home from work early. She came in the house, screaming and crying and when she saw karen johson and mark chambers, she ran over to them.}
\\
\\
\texttt{karen johson tried to hold diane back, but diane pushed her away. She was crazed with rage and anger as she continued to yell at mark chambers. She demanded that he get out of her house because their marriage is over. Mark Chambers tried desperately to explain but diane would not listen..}
\\
\\
\texttt{For a moment, diane stopped shouting and looked at karen johson on the floor - she looked back at Mark Chambers - then she turned around, grabbed her keys off the table by the door and ran out of the house. Mark Chambers was left sitting on the floor in shock. He just didn't know what to do. He watched Karen Johnson as she walked out of his house then he started walking towards his bedroom...}
\\
\\
\texttt{Until they could figure things out, they would take turns staying at one another's houses; neither one wanting to be alone in their own homes unless they were working.. .}
\\
\\
\bottomrule
\caption{\ours{}'s generated story for Premise 1. The story follows the initial premise reasonably well, with Diane learning that Mark is cheating on her. Subsequently, ``She wanted to hurt him as much as he had hurt her; more than anything else in the world, Diane Chambers wanted
to make Mark Chambers suffer for what he had done.'' However, the execution of the ``revenge'' is somewhat lacking. In particular, \ours{} does not follow the outline it generated initially, which stipulates that Diane has an affair of her own: Diane appears to lock Mark out of the house, but does not engage in an affair. The story is for the most part highly coherent and understandable, up until a somewhat confusing ending beginning with the sections denoted by a string of ``\texttt{*}''s. Additionally, there is at least one notable factual inconsistency near the end, when Mark goes to Karen's house but then appears to find Diane's bedroom in that house. Finally, there is an odd passage at the very end where names become lowercased.}
\label{tab:premise1_ours}
\end{longtable}
\end{small}
% \end{xtabular*}

% }\label{tab:premise1ours}

\begin{small}
\begin{longtable}{p{0.9\textwidth}}
\toprule
\textbf{\rolling{} Story for Premise 1}\\
\midrule
\texttt{“Linda, would you come in here please?” he called.}
\\
\\
\texttt{I was just thinking of heading out to the mall to see if I could get a birthday gift for my little brother when he called. He had been acting strange all week, but I didn’t know what was going on. I raced back into the dining room, where my husband was waiting for me. As I came around the corner, I saw him holding an open box of condoms in his hands and a wicked smile on his face.}
\\
\\
\texttt{What the hell? He ran his hand through my hair and said, “Linda, baby…I think you are getting screwed real good by one of your customers at work. And they aren’t getting off with their fingers this time!”}
\\
\\
\texttt{Angry tears sprang to my eyes as I grabbed the box from him and threw it across the room. “How dare you! Where did you get these? Who told you this crap?” I demanded to know immediately. Then it dawned on me that someone had put him up to this. It just wasn't like him to do something like this without provocation! “I’ve had my suspicions for awhile now, and after I found this in the garbage, I decided it was time to confront you.”}
\\
\\
\texttt{“I do not cheat on you, Brent! Where did you get this?!” I screamed.}
\\
\\
\texttt{He smiled smugly at me and said, “Well, my friends said that they put a rush on this package for when I got home from work. Now I just want to know who the lucky guy is…so I can come over there and kick his ass!” He was smiling in satisfaction as he walked around the dining room table toward me to give me a hug.}
\\
\\
\texttt{I recoiled from him as if he were covered in slime and shouted, “Get the hell away from me! How dare you accuse me of such a thing! Why would you think that just because some guy is nice to me that he wants to screw me too? You are such an asshole! And where did you get these from anyway? Who has been spying on us?! Brent, if you cheat on me and then try to make it my fault, I swear…” I stopped talking and looked up at him with shock. “What? You think I got these from you? Why would I get a box of condoms from you?” he asked me, with a look of equal shock.}
\\
\\
\texttt{“Someone has been following us and telling everyone in our cities that they know what we are up to! Brent, I don’t know who it is, but someone has been trying to break us up for awhile now. I am so sorry that you found out like this. I swear to you that nothing is going on between me and anyone else but you. My heart belongs to you and only you, Brent! Please forgive me for any tears that have come from my eyes lately, it was nothing more than stress from work and trying to get my store open on time.”}
\\
\\
\texttt{Brent looked at me in amazement as he took my hands in his and said, “Linda…I love you more than life itself. You are the best thing that has ever happened to me and I could never go through this again…I don’t want anyone else but YOU! Think about how our kids would feel if they found out their mother had been cheating on their father? I know that neither of us have cheated before, but you have worked with this guy for awhile now, I mean…are you sure that nothing is going on?”}
\\
\\
\texttt{“I swear to you that I will never again work for this guy. If I do have a problem or need help, I will approach you first before I go to him! Brent, please don't leave me! Tell me what to do and I will do anything to make this up to you!”}
\\
\\
\texttt{“I don't want a divorce. We are not going anywhere. We'll just play it by ear and see how it goes. If he wants more than just friendship and gets pushy with you, then we'll worry about it then. For now, let's keep going forward and forget the past. We'll forget all about Jonathon and his friends too…okay?” He smiled at me as he put his arms around my waist and pulled me tightly against him.}
\\
\\
\texttt{“Thank you for taking me back…my love! You have no idea what a relief it is that you believe me over somebody who was jealous of us! Now we can go on with our lives like normal again!” I kissed him and he started to take my jacket off. “What are you doing? I just bought this dress and now you want to mess it up again?”}
\\
\\
\texttt{“You think that I don't know my own wife by now? How long do you think it's going to take before Jonathon pops up again? You won't be able to stop him from coming after you, Linda…he has a lot of money to make this happen. Are you going to get rid of your phone and have me change mine too so he can't track us that way either? One way or another, he is going to come after you one way or another…so why not be ready for him for when he does show up next week, huh?”}
\\
\\
\texttt{I sighed in defeat as I agreed with him. “You're right…this is supposed to be a honeymoon, but between the two of us being so stressed out with work lately…this is more like hell! Let's go upstairs and we can start enjoying our time together again! I'm sorry for the distrust that I put in you and hopefully from now on, we will have nothing but trust between us both! Now let's make love! ”}
\\
\\
\texttt{He pulled my hair back and said, “I want to make love to you…but I also want to talk for awhile first.”}
\\
\\
\texttt{“About what?”}
\\
\\
\texttt{“About why Jonathon didn't think that I was worthy of you or vice versa. I mean, why he thought that he was better than me or vice versa, as far as being with you goes. I mean, we are both good looking guys and have pretty much the same background in life. We have about the same income, too! There must be something else that makes him so much better than me! Do you think that it's…?” Brent fell silent as he looked at me to finish his question.}
\\
\\
\texttt{“I don't know what it is…but if this is going to be a problem between us Linda and Brent, then I'll make sure he never comes after either of us again! I'll call up my boss at the bank and ask him if there is anything that we can do about this guy!” He reached out for his cell phone and started dialing a number in the computer screen instead of the normal buttons on his phone pad. “Brent, it's okay…I'm not going to worry about Jonathon anymore. I've wasted so much time and energy trying to figure out this whole situation that I've really neglected you and you are still with me. Maybe he came onto me because he thought that of all the women at work, I was the easiest for him to get. Maybe some of the others turned him down and it made him mad. Who knows?” Brent smiled as he put his arms around me again and kissed me tenderly as his hands moved up and down my back.}
\\
\\
\texttt{He whispered, “I love you, Linda! Let's go upstairs so we can spend some quality time together!” He hugged me tightly before we went upstairs to our bedroom where I knew he wanted to make love to me for hours. This time we would be making love together instead of against each other like last night earlier today!}
\\
\\
\texttt{The next morning Brent got up early while it was still dark outside and went down into the garage where his garage door opened automatically as a large truck backed up into it! I wondered what was going on since this was the first time that he had ever taken the car down there. Brent came back into the house and handed me a computer pad and said, “I need you to go online to one of these sites and then I'm going to have you read off the numbers that you see on the screen, okay?”}
\\
\\
\texttt{“What are we looking for?”}
\\
\\
\texttt{“I just want to get some information downstairs so I can put his truck on a list. It's nothing against you, Linda…it's just that now that we know his name, I want him to know that we will do everything possible in order to protect ourselves! This is a guy who tried to blackmail both of us with sex pictures!...you are too special for me, Linda. There is no way I am going to let him take over our lives again!” We held hands as we went into the den and sat down in front of the computer screen together as he typed in his information. My hands were shaking a little bit as I asked myself again if maybe this was all my fault for flirting with Jonathon at work? Maybe he was the type of guy who would do anything to get a woman and that included blackmail!}
\\
\\
\texttt{I held up the pad and read off the information about Brent's truck, “License number: 70855, make: Ford, model: pickup…” Brent continued, “Year: 2012.” He continued to type in his information as I kept on reading, “Color: gray with gray trim.” Brent finished typing in everything, then pressed a button on the computer pad and said, “Well done! Now all we need to do is wait and see if he tries anything else with us! If he does…I will have something on him to show the police! We aren't going to let him push us around anymore!” I felt better knowing that at least we were doing something about this instead of just letting it happen again.}
\\
\\
\texttt{I took a deep breath as I turned around and asked him if he would be willing for me to go to an art class so that I could take a class on how to paint? He smiled at me as he said that we should both enroll together. I told him no because it was my hobby and not his. Maybe we could enroll a couple sometime after Christmas when the kids were back in school and we didn't have all of the family or holiday activities to go to. Sometimes it was nice just to be alone with each other, but of course I didn't want to tell him that!}
\\
\\
\texttt{Our morning sex session was much better this time and Brent seemed so happy that he could do things with me that he didn't want to do with anyone else. I told him that I did too and I knew this was something special that only we could share together. He kissed me deeply as he whispered, “I love you, baby!” I felt so safe right then as he held me tightly and made love to me like he had never done before!}
\\
\\
\texttt{From the next day on, Brent spent several hours each day at his computer looking for ways in which we could protect our home from any burglars, such as using an alarm system and also having security cameras installed outside along with motion detectors who would fire off a high-powered laser beam if anyone tried to enter the house. They would also have a speaker system that would sound an alarm if anyone even tried to come within ten feet of our house! He said it was also important to make sure that our front door was set up so that it could not be opened with a key or a credit card. These doors had a metal piece of steel that went across it and you had to know the secret way in order to get inside.}
\\
\\
\texttt{He helped me watch some Netflix movies about home security which helped me learn how to better protect our home. I told him how glad I was that he was so concerned about this because I wanted us to feel safe here at home while we were spending time together. We had such a wonderful evening together and I knew in my heart he was going the extra mile for me. Sometimes I wondered if all men were like him or if perhaps he was unique in his ways and we just happened to both be compatible! If only I knew for sure!}
\\
\\
\texttt{The next day Brent got up early again, but instead of heading down into his garage where he normally worked, he came into the bedroom where I was laying as Jonathon crept out of my room! I watched him get dressed while I laid there in bed with the covers pulled up over my breasts and watched him dress. Why was he sneaking out of my room? What had they been doing all this time down in the garage? And what had he just been doing in my room? And who was I kidding...I had already put it together on my own! Two and two equals four!}
\\
\\
\texttt{“Where are you going, honey?” I asked as I pretended I hadn't seen Jonathon. Brent slowly looked back at me and said that he was going to work for a couple of hours so that he could get some things done. His uncle needed some help with something so he would be working from home today since everyone else wasn't available. He told me to meet him out by the garage when I'm ready and we can go to his shop and pick up some more items.}
\\
\\
\texttt{I held back my tears as I told him that it was okay if he needed to work alone for a couple of hours because I plan to enjoy the rest of my morning sleeping in. He gave me a quick kiss on the cheek before he walked out of the room and quietly shut the door behind him.}
\\
\\
\texttt{I couldn't keep it up any longer as I got out of bed and ran over to the window where I watched Brent and Jonathon drive off in the truck together! What the hell were they doing together? Where were they going? What else were they hiding from me? Was there anything else that I didn't know about these two men, who I thought I knew so well?}
\\
\bottomrule
\caption{\rolling{}'s generated story for Premise 1. The story deviates completely from the premise. In fact, it seems that despite the initial accusations, neither Linda nor Brent actually cheated on the other, and they quickly reconcile. The story also ends oddly, and there is some confusion about what exactly is the relation between the couple and Jonathon.}
\label{tab:premise1_rolling}
\end{longtable}
\end{small}

\begin{small}
\begin{longtable}{p{0.9\textwidth}}
\toprule
\textbf{\rollingfinetune{} Story for Premise 1}\\
\midrule
\texttt{Bitch. Dyke. Slut. Those words flashed across my mind like a neon sign, burning into my head so deep that I could see them against the inside of my eyelids.}
\\
\\
\texttt{It was no secret that I'd been dating Erin for the past year, but everyone knew she and I were just friends. Or that's what they thought anyway.}
\\
\\
\texttt{My husband was one of those guys who constantly craved attention from other women, so it wasn't surprising when he'd started to wander away from our marriage not long after we got engaged. It had been a horrible time in our relationship, but after getting engaged we'd decided to stick it out and get married anyway even though he wasn't what most people would consider faithful anymore. We both knew he was a pig, but he kept telling me how much of a disaster it would be if we broke up and tried to get married two years later after the divorce papers were finalized.}
\\
\\
\texttt{Even though we no longer shared a bed together, we decided it would probably work out in the end because I didn't want to be tied down to someone who wasn't right for me for the rest of my life. I just needed to wait until the right guy came along.}
\\
\\
\texttt{That's when I met Erin.}
\\
\\
\texttt{I'd been working at a local bakery when some of the men who worked at my husband's company came in one day to get ice-cream cakes for their boss' birthday. My boss suggested I go in and ask if they'd like a cake for their boss after she went back into the kitchen, so she could take care of some things and leave me alone with the customers for a few minutes. It was just after lunch so there weren't many people in the store, but I hesitated for a second because I'd been told over and over again that interaction with customers was part of my job description.}
\\
\\
\texttt{"Come on, you can do it." My boss said from behind me, pushing me out from behind the counter and out into the store where they were waiting. She walked past me and straight up to them without saying anything else, leaving me standing there feeling like an idiot.}
\\
\\
\texttt{"Hi, how can I help you?" I asked nervously as they looked at me expectantly. Did they really want another cake? Did they want two cakes? "Oh, do you work here?" One of the men asked. He had dark hair and blue eyes, and I looked at him with a nod.}
\\
\\
\texttt{"Yes I do." I said as confidently as possible, but I still felt like they could see right through me. There was something in the way they were looking at me that made me feel uncomfortable.}
\\
\\
\texttt{"I know this is a bit awkward, but we've been told to come in here and ask if any of you have been making special deliveries for Charles or Jessica." The man who'd asked me about my job continued as though he hadn't said anything out of the ordinary.}
\\
\\
\texttt{"I'm sorry," I said without much inflection, feeling uncertain about where this conversation was going to go. "I can't help you."}
\\
\\
\texttt{"Oh come on," The second man teased her as he took a bite of his donut from one of the boxes he'd been carrying with him. He had dark hair and brown eyes, not to mention a hint of an Irish accent that sent chills down my spine even though it was kind-of cute. "You can tell us."}
\\
\\
\texttt{"Can you two mind yourselves?" The first man said, glaring at the taller man as he looked back at him with a smirk. "We're here on official business and I'd hate for you to get in trouble for all of this." He turned back to me and gave me a look that was both stern and disheartening. "I'm sure you've seen the news lately about my boss's wife, right?"}
\\
\\
\texttt{"I haven't seen any of those magazines." I said quickly when he let out a little snort of laughter at my response. "But yes, I have heard from one of my coworkers that there was some sort of scandal going on with their marriage."}
\\
\\
\texttt{I looked down at the floor as I felt myself begin to blush profusely. In spite of everything that had happened in our own relationship, it pained me to hear about other couples' problems. Particularly when it came to happy marriages like Charles and Jessica's. "So," The tall man said as he sat up straight and leaned forward towards me with a look of interest in his eyes. "Have they been hiring someone else?" "What for?" I asked with a frown. I wasn't sure what to do with this information. If I told them anything then they'd know that I'd been listening in on their conversation and that would get me fired. "I haven't heard anything about any of that."}
\\
\\
\texttt{"You will soon." The first man said, giving me a look of frustration before signaling for the other two men to follow him out of the bakery.}
\\
\\
\texttt{My boss walked out from the kitchen to see who had left without buying anything and looked back at me questioningly as she surveyed the half-empty boxes in front of me, an unspoken question hanging between us. I gave her a shrug and a forced smile before going into the kitchen to start smashing pies into paper bags so we'd have something to give all of our customers who came in with questions about the strange meeting we'd just had.}
\\
\\
\texttt{It wasn't until late that night that I realized there was probably a reason why those men had been asking about deliveries for Charles and Jessica. They were usually sent in by someone who was trying to hide something, like how they were cheating on their spouse or they were trying to get pictures of them doing something bad so they could blackmail them later down the road. I didn't know what the magazine wanted, but I was pretty sure I did not want to be involved in any of it.}
\\
\\
\texttt{- Chase -}
\\
\\
\texttt{"He's not here." The blonde who answered the door said with a smile and a shrug when she looked out of her doorway to see my best friend standing there. It took everything in me not to roll my eyes at her, even though she had a pretty great smile. "He's out on a date. Sorry!" She giggled as she closed the door in his face, leaving him on the other side looking completely dumbfounded.}
\\
\\
\texttt{"Well, that was rude." He said with a smirk that made me want to punch him in his stupid face. "And you know how much I love being told that he's out on a date."}
\\
\\
\texttt{"I don't." I said dryly, unable to tear my eyes away as he ran his fingers through his short black hair and sighed loudly before turning around and walking down the steps again. "So why are you here?" I asked curiously as he approached me and gave me one of those infuriating smiles that always made me melt inside. "Guilty conscience, I guess." He said with another shrug and a shake of his head. "I mean, it's not like I sent him out on a date so that I could try to get him to come over and see you or anything, but… I know he feels horrible about what happened at the restaurant."}
\\
\\
\texttt{"Well, then he shouldn't have lied about wanting to see me in the first place!" I groaned and dropped my face into my hands as he looked back at me with amusement. "I don't know why they always think they can just say whatever the fuck they want without thinking about how we feel."}
\\
\\
\texttt{"Because they're assholes?" He suggested helpfully before scrunching up his nose apologetically. "Sorry!"}
\\
\\
\texttt{"Don't be sorry!" I said with a laugh before leaning forward towards him and placing a hand on either side of his face as I looked into his eyes. "You're so fucking sweet! Just take me home now because it's getting late."}
\\
\\
\texttt{"What are you going to do?" Chase asked curiously as we walked down the street together towards my apartment building. "Do you have any idea where he is?" "No clue." I said with a shake of my head. "But I think that's what makes it so fun." He chuckled as I pushed him up against the front door and kissed him deeply. "Will you fuck me tonight?"}
\\
\\
\texttt{"Of course I will, baby." He said with a grin as he pulled back from me and wrapped his arms around my waist. "How would you ever think that–" The sound of talking coming from inside the apartment building made me look over my shoulder at the staircase behind us to see Charles standing at the top, looking down at us with a confused expression on his face. "Fuck!" We both said together as we broke away from each other and pretended to be doing nothing wrong.}
\\
\\
\texttt{Chase had been so close to getting me into bed… but then again, this situation probably wasn't exactly ideal for him either.}
\\
\\
\texttt{"Hi!" I called up to him with a fake smile before waving enthusiastically like he didn't know why we were standing outside the apartment building of the guy who was supposed to be on a date with another girl. "How are you? Fancy seeing you here, huh?" "Don't have any idea what you're talking about, Chase." Charles said as he walked towards us with his hands in his pockets and a smirk on his face. "Have a good night, though! And please tell me you didn't send Rory home in a cab?"}
\\
\\
\texttt{"Of course he did." I said dryly, pulling out my keys and unlocking the front door to let all of us walk inside. "He's my boyfriend now and he has to do whatever I say."}
\\
\\
\texttt{"Ah yes." Charles said with a grin as we all walked into my apartment building. "I wonder where I've heard that before…" "You're such an asshole!" I laughed as Chase pulled me away and into the elevator with him. "I don't know why I let you live here!" "Well, I pay rent just like everyone else." He pointed out as we rode up the elevator together. "Plus, it's not like there's anything wrong with my apartment…"}
\\
\\
\texttt{"You could barely fit one person in there." I pointed out playfully as the doors opened up again. "It's barely bigger than a closet!" "It's more than a closet." He retorted with a grin as we walked down the hallway towards my apartment. "And I don't know why you're still here… it's the weekend, baby!" He said in an exaggeratedly excited voice as we reached my door and I unlocked it and opened it up to let him in.}
\\
\\
\texttt{"Get your head out of the gutter!" I laughed as he walked past me into the apartment, shaking his head at me like I was just so adorable for being such a prude. "I just have work tomorrow."}
\\
\\
\texttt{"Nah, you don't have work tomorrow! Rory is working until six o'clock in the morning since he was stupid enough to miss his date with your sexy ass…"}
\\
\\
\texttt{"You're loud," I said with a laugh as I closed the door behind him and locked it before walking over towards him and leaning against the kitchen counter next to him. "You're supposed to be quiet when you're trying to be seductive." He chuckled before pulling me closer to him by my waist and kissing me deeply again. It really wasn't fair how good he was at this… yet somehow Chase seemed like he wasn't even trying at all. "So that's how it is?" I laughed as we pulled away from each other and he gave me a sly grin. "You're going to seduce me now?"}
\\
\\
\texttt{"I don't want to seduce you, Rory." He said with a chuckle. "I want you to want me…" "You know I do." I said with a smile as I reached out and touched his face before kissing him lightly.}
\\
\\
\texttt{"Then why are you being so difficult?" He asked with a grin before he kissed me back and trailed his hands down my sides. "And what do you mean 'until six?' That's midnight, baby… it doesn't take that long to watch somebody sleep."}
\\
\\
\texttt{"Yeah, but he did the whole 'sleep-with-you-and-then-leave' thing last time we were together…"}
\\
\\
\texttt{"He also had to leave because your stepbrother came home in the middle of the night… something tells me that wasn't an accident." He laughed as he pulled away from me, leaning back against the kitchen counter again. "And speaking of which…" Chase added, reaching into his pocket. "I got you something…"}
\\
\\
\texttt{"Oh? What's that?" I asked a bit curiously as I raised an eyebrow and looked at him. "Is that the reason why you were acting so weird on the phone, by the way?"}
\\
\\
\texttt{"No, that was all because of Georgia." He laughed before he handed me the box. "I got this for you in case you didn't have any plans with Rory tonight, but I really hope you guys can spend some time together tomorrow night…" "For my birthday?" I asked with a little smile as Chase nodded and pulled out a card from his back pocket.}
\\
\\
\texttt{"I know it's not for a week after your birthday… but we have some stuff to do and I didn't want to bring it up if Rory was busy with work…"}
\\
\\
\texttt{"That's fine," I said, smiling at him again as I opened up the card and read it silently. It said 'happy birthday' across the top of it in pretty cursive writing followed by 'you're officially an adult now!' under that and then there was just one long paragraph detailing all of Chase's plans for us over the next week and how he wanted to make our first anniversary special. After looking through all of it, I put it back together again and handed it to him before opening up the box again.  "You got me a necklace?" I asked with a little bit of surprise as I pulled out the silver chain with a small heart shaped locket attached to it.}
\\
\\
\texttt{"I know it's not much, but I thought you'd like it." He said with a shrug as he watched me put it on before I turned and looked at him again. "Do you like it?"}
\\
\\
\texttt{"Chase…" I started, shaking my head a little and reaching up to touch the locket. "It's perfect… thank you so much."}
\\
\\
\texttt{"I love you, Rory." He said softly as he moved closer to me and wrapped his arms around me. "Happy birthday."}
\\
\bottomrule
\caption{\rollingfinetune{}'s generated story for Premise 1. The beginning of the story is confusing and it is unclear if the protagonist is actually cheating on her husband with Erin. After the first couple of paragraphs, neither the husband nor Erin are ever mentioned again, and the story becomes about the main character's interactions with their customers at the bakery. Finally, the protagonist has a conversation with Chase, who professes his love for Rory. Weirdly, based on this conversation, Rory is at times the main character and at times a different character altogether.}

\label{tab:premise1_rollingft}
\end{longtable}
\end{small}

\FloatBarrier

\newpage
\subsection{Examples for Premise 2}

\begin{table}[!htbp]
\small
\begin{tabularx}{\linewidth}{X}
\toprule
\textbf{Premise 2}\\
\midrule
\texttt{After the loss of her father, Shannon is determined to finish college and make something of herself. However, when she's sucked into a time portal and lands in the 19th century, Shannon has to find a way back to her own time before it's too late.}\\
\bottomrule
\label{tab:premise2}
\caption{Premise for the second group of example generated stories. A somewhat exotic premise which may be more challenging to follow.}
\end{tabularx}
\end{table}

\begin{table}[!htbp]
\small
\begin{tabularx}{\linewidth}{X}
\toprule
\textbf{Initial Setup and Outline for \ours{}, Premise 2}\\
\midrule
\texttt{Premise: After the loss of her father, Shannon is determined to finish college and make something of herself. However, when she's sucked into a time portal and lands in the 19th century, Shannon has to find a way back to her own time before it's too late.}
\\
\\
\texttt{Setting: The story is set in both the present day and the 19th century.}
\\
\\
\texttt{Characters:}
\\
\texttt{1. Shannon Randall is a young woman in her early twenties. She has long brown hair and brown eyes.}
\\
\texttt{2. Jaxon Montgomery is a young man in his early twenties. He has dark hair and blue eyes.}
\\
\texttt{3. William Gray is a middle-aged man in his forties. He has graying hair and brown eyes.}
\\
\\
\texttt{Outline:}
\\
\texttt{1. Shannon is transported to the 19th century where she meets Jaxon Montgomery.}
\\
\texttt{2. Shannon and Jaxon fall in love, but she is still determined to find a way back to her own time.}
\\
\texttt{3. Shannon discovers that her father is still alive in the 19th century and she must choose between staying with Jaxon or returning to her own time.}
\\
\bottomrule
\caption{\ours{}'s initial setup and outline for Premise 2. The characters and outline make perfect sense for the given premise. }
\label{tab:premise2_ourssetup}
\end{tabularx}
\end{table}

\begin{small}
\begin{longtable}{p{0.9\textwidth}}
\toprule
\textbf{\ours{} Story for Premise 2}\\
\midrule
\texttt{Shannon was running late and hurried along the sidewalk. She was supposed to meet her best friend Hannah Montgomery in a few minutes at the restaurant where they both worked. Shannon worked as a server at the little diner in town while Hannah waited tables at the local inn.}
\\
\\
\texttt{Shannon had been out later than she intended with her boyfriend, Marcus, and now she was running late. She turned the corner onto Main Street and spotted Hannah already sitting at a table outside the restaurant on one of the benches out front.}
\\
\\
\texttt{Hannah's dark brown hair shone in the afternoon sunlight like a halo surrounding her head. Her brown eyes sparkled with mischief and excitement, as usual, when she looked up for Shannon to join her for lunch. "Sorry I'm late," Shannon apologized breathlessly as she sat down across from Hannah.}
\\
\\
\texttt{"I figured you'd show up eventually," Hannah said as she arched an eyebrow suspiciously in Shannon's direction. "Or did your boyfriend finally call it quits?"}
\\
\\
\texttt{"No way," Shannon replied quickly as she shook her head back and forth slowly from side to side. "Marcus definitely isn't calling it quits on us now that we're moving in together." "You're moving in with your boyfriend?" Hannah's voice grew louder with each word. "What happened to your townhouse?"}
\\
\\
\texttt{"I sold it," Shannon revealed as she shrugged her shoulders. "I figured that Marcus and I should combine our finances since we're about to be married." Shannon blushed as a smile spread across her face. "Plus, I've always wanted the experience of living on a farm."}
\\
\\
\texttt{Hannah rolled her eyes back into her head as she laughed at Shannon's response. "You are so lucky I'm your friend."}
\\
\\
\texttt{Shannon smiled as she looked around at the other diners nearby before speaking again. "Speaking of friends, you haven't told me anything lately about you and your brother Jaxon."}
\\
\\
\texttt{Hannah's eyes got big, and the look on her face was priceless. She opened and closed her mouth several times like a fish out of water before she finally whispered, "I can't believe you brought that up right now after everything I've told you lately!" Shannon picked up her glass of water and took a sip as she glanced at her friend skeptically. "It's just … I mean you two were supposed to get married two years ago, but you broke it off at the last minute. Why didn't you ever tell me what happened?"}
\\
\\
\texttt{Hannah sighed as she began twisting her napkin around in her hands while she looked down. "I don't know if I'm ready to discuss that yet ... but I will tell you this much. When Jaxon proposed, we'd already been dating for over a year, so I went with it. A few months before our wedding date, Jaxon told me he wasn't ready to settle down …"}
\\
\\
\texttt{"We've known each other since we were kids," Hannah said softly as she glanced towards Jaxon and Ben who were out on the front porch talking. "I know that Jaxon is the least likely of all of us to come out of his shell." She quickly glanced back at Shannon before she continued speaking. "But if there's anyone who can help him open up … it's my brother." She smiled as she reached across the table to pat Shannon's hand gently. Shannon glanced over her shoulder towards the copy of Romeo and Juliet that Hannah had been reading. She chuckled as she leaned back in her chair, shaking her head slightly. I think this is just another one of your crazy schemes … trying to set me up with Jaxon. She laughed lightly before taking a sip of her water again.}
\\
\\
\texttt{Hannah's face flushed as she stood abruptly and began clearing their plates from the table. "My brother is not a project!" she exclaimed defensively. Shannon could see tears welling up in Hannah's eyes as she turned to look at her, but she remained quiet.}
\\
\\
\texttt{I'm sorry … Shannon said softly as she slipped out of her chair and walked over to Hannah who was standing behind the small table with their empty plates in one hand and silverware in the other. She took the dirty dishes from Hannah's hands and carried them inside before walking back outside with a fresh glass of water for each of them. After silently handing Hannah hers, Shannon sat back down in the patio chair across from her, sighing softly as she took another sip from hers. You're right … I'm sorry. She sighed softly as she set her water glass back down on the small table before them. She shook her head slightly, not sure what to do. She had not expected Hannah to react that way and she felt like an ass for making her cry.}
\\
\\
\texttt{Hannah smiled as she picked up her glass and took a long drink from it before setting it back down on the table. "It's not you … it's just that this whole thing has been so difficult for him." Her eyes shone brightly with unshed tears as she looked at Shannon, then she quickly glanced away, picking up her water again to take another sip from it. "I know I can be hard on him sometimes … but he's my younger brother."}
\\
\\
\texttt{Shannon reached across the small table and covered Hannah's hand with one of hers. "I know he is." Her heart ached at the thought of Hannah being anything less than sweet and nurturing with anyone, especially Jaxon who had obviously always been close to his younger sister. Thinking about him made Shannon smile slightly as she recalled the first time they met … right here in this backyard after noticing one another edging closer together while they stood behind their respective parent's as each spoke with someone across the yard from them. She had been more than a little nervous then, having no idea how to introduce herself to him, while he seemed the very same way at first. But soon they were willing to step out from behind their parents and shyly introduce themselves.}
\\
\\
\texttt{"You know … maybe I should just go talk with him for a bit …" Hannah said as she looked at her hands where Shannon's still covered her own. When Shannon didn't respond, Hannah peered up at her from beneath her lashes and was surprised to see her eyes focused intently on something over the top of Hannah's head. She followed Shannon's gaze and saw Jaxon approaching them from across the yard, making his way through the crowd gathered together in their small backyard for the cookout that had been going on all afternoon.}
\\
\\
\texttt{"Are you two okay?" He asked as he came up beside them and sat down in one of the two chairs facing them across the table with his sister next to him on one side and Shannon opposite him on the other side.}
\\
\\
\texttt{Shannon smiled when she looked into his eyes, blue like he shared with his younger sister beside him, and nodded slowly before she said quietly, "Yeah … we're fine." When he didn't say anything else, she looked at Hannah and asked, as quietly as she could, "Did you tell him?"}
\\
\\
\texttt{Hannah nodded and replied, just as quietly, "Yeah … I told him."}
\\
\\
\texttt{Shannon sighed and then looked back at Jaxon. His eyes were focused intently on her face now, almost like he was trying to figure things out with his eyes. She asked him anxiously before he could any questions of his own about what she was doing there, "So are you going to ask me how I got here?"}
\\
\\
\texttt{He nodded slowly without taking his eyes off of her face and said softly so that anyone but Shannon would have had a hard time hearing him in the crowd around them again, "I'm assuming that you're not local to this area…" He trailed off when he got no response from her and finally asked a little louder than he had been speaking previously, "What do you mean that you're from the future?"}
\\
\\
\texttt{She shook her head slowly as if his words were nonsense to her. She had already expected this question. She knew that he would probably think she was nuts just as everyone else who had ever found out about her time traveling did. She could see in his eyes that he was feeling the same sort of disbelief and shock as everyone else had, yet she also noticed something else in his eyes as well. There was a look of intrigue on his face that made her think he may actually believe what she had told him so far. He didn't seem to be looking at her like she was insane or like she was lying to him or making things up or trying to get attention, like so many other people did when they found out about it. She found herself hoping desperately that he would believe her and not try to lock her up somewhere for psychiatric evaluation or otherwise take measures to keep her from leaving "his" time.}
\\
\\
\texttt{She decided to take a gamble with him anyway, since there really wasn't anything left for Shannon to lose at this point anyway. She said quietly, "Are you serious?"}
\\
\\
\texttt{He nodded, seeming even more intrigued now than before and replied, "Of course I am." She was silent for a moment and just looked down at her hands. She glanced back up at him and she said quietly, "I am so sorry. I never meant to involve you in this."}
\\
\\
\texttt{He nodded and said carefully, "I know you didn't."}
\\
\\
\texttt{She nodded, looking down at her hands again. She could feel tears stinging her eyes as fear over the fact that if he wasn't going to help her now, then no one was coming to help her. She could feel herself on the verge of tears again, but she quickly pushed them back down and picked up her head again. She looked him straight in the eyes and said firmly, "I must go back to my time now."}
\\
\\
\texttt{He nodded slowly and replied in a low voice, "So it seems."}
\\
\\
\texttt{She stared silently for a moment before continuing to speak as if she was trying to convince herself that it really didn't matter anyway since no one was ever going to believe what had happened anyway even if he did believe her about everything else. She said quietly, "No one will ever believe it anyway so it really doesn't matter what you do with me from here on out. I will have no further trouble from you or anyone else. It is time for me to leave."}
\\
\\
\texttt{Jaxon nodded and said quietly, "Where is your time machine?"}
\\
\\
\texttt{She looked at him in shock and when he realized what he had said, he quickly glanced at Hannah who was staring at Shannon with wide eyes. The two of them stared at each other for a moment and then Jaxon turned back to face Shannon who was watching him closely. She had just begun to speak when her father chose that moment to ask, "What are you talking about? What time machine? Where are you going?"}
\\
\\
\texttt{Shannon looked up and stared into William's eyes which were filled with concern. She felt tears stinging in her eyes again and she quickly stumbled to her feet and rushed out the door.}
\\
\\
\texttt{She did not stop running until she reached the creek. It seemed like nothing changed here. The water was still as clear as it was yesterday, the trees were still tall, but this time there were no animals anywhere around her or her family's property even though she had been told that there should be plenty of them everywhere around here. She laid down on the ground and wrapped her arms tightly around her knees. There was no way she could ever go back to her own time after all this. She did not want to leave and she could not return. Perhaps it would have been better if she had stayed dead with Jaxon in that place, but she couldn't help thinking about how much he loved her and how he had been waiting for so long for her to come back to him. He had told her all about it yesterday and despite everything, she still felt a sense of love towards him that was overwhelming. She wanted desperately to see him again, but if he was going to keep trying to protect her from harms way, then perhaps it might be better that she left him alone for awhile so that he wouldn't cause the same accident in an attempt to get them both back together in their own time.}
\\
\\
\texttt{Shannon wiped the tears from her eyes, then stood up and walked slowly back up the hill towards Hannah's house where Jaxon would be waiting for her with a working time machine. The first thing Shannon did after sitting down opposite Jaxon was apologize. After what felt like an eternity of silence, Jaxon finally spoke up saying, "I forgive you." Shannon looked down at her feet, then back up at her lover. "Thank you. I am so sorry."}
\\
\\
\texttt{Jaxon smiled and shook his head, running his fingers through his blond hair. "How many times could I have gotten myself killed trying to find you back in your own time? Your father must be half out of his mind with worry." Jaxon did not mention how worried he had been while Shannon was gone, but he knew that it was the only way to get her to go back in time without a fight or argument. Shannon looked down at her feet again, then back up at Jaxon and said, "My father is alive?" Tears began streaming down Shannon's face and she leaned forward towards him while also moving closer to him on the bench they were sitting on as well. She put her face into Jaxon's shoulder and cried softly until she was able to gain some control over herself.}
\\
\\
\texttt{Jaxon gently rubbed Shannon's back in support and quietly said, "Yes, my love. He is alive and well." Shannon took a few deep breaths to calm herself down and then said, "I have to go see him." Jaxon nodded and said, "Yes, you do." Shannon looked at Jaxon with a hint of sadness in her eyes and said, "I'm sorry. I know you were hoping I would stay with you."}
\\
\\
\texttt{Jaxon shook his head and said, "No, Shannon. I want you to be happy. And if that means going back to your own time, then so be it." Shannon nodded and leaned forward to give Jaxon a kiss on the cheek. They both stood up and walked towards the time machine. Shannon hesitated for a moment before getting in, but she knew that she had made her decision. She turned back to look at Jaxon one last time before getting into the time machine. He smiled and nodded at her, telling her that everything would be alright.}
\\
\\
\texttt{And with that, Shannon Randall vanished from the 19th century, never to be seen again.}
\\
\bottomrule
\caption{\ours{}'s generated story for Premise 2. The story never explicitly mentions that Shannon gets whisked back to the 19th century. Additionally, the introduction of Jaxon (Shannon's supposed love interest, according to the outline) is very confusing, and it is unclear what his relation to Shannon and Hannah is. But despite these problems, the story clearly gets back on track later on. In the generally well-written and coherent second half of the story, Shannon and Jaxon---clearly in love---agonize over Shannon's need to return to her own time (which she eventually does).}

\label{tab:premise2_ours}
\end{longtable}
\end{small}

\begin{small}
\begin{longtable}{p{0.9\textwidth}}
\toprule
\textbf{\rolling{} Story for Premise 2}\\
\midrule
\texttt{"Excelsior!" The crowd cheered, and the boy knew he had done it. He had finally flown. What a glorious day, he thought as he landed daintily out of his family's sight. He couldn't have been happier to have took flight. His family would be proud of him now just like they always said they'd be. His father remained doubtful though.}
\\
\\
\texttt{As soon as the boy landed, he knew that something was not right. He didn't hear the bullfrogs croaking, or the bees buzzing, or even the crickets chirping. They all seemed to go silent at the same time just moments ago when he took flight for the first time without fail and without accident. It was almost like something…no someone else had scared them off from their normal daily routines as if they were all afraid of something in particular – a something that caused nothing but fear in every creature around him – but fear of what?}
\\
\\
\texttt{The boy looked around and noticed something strange going on with his surroundings then too. His family's farm was normally a very busy place with people coming and going all the time. He could hear their voices and movements in the fields as they worked, but now it was silent as a tomb. "How odd," he thought, "The farm is dead silent."}
\\
\\
\texttt{He decided to leave it at that for the moment because something caught his attention. It seemed to be coming from the fields, which was very odd considering that there was no one on the farm at this moment. The boy turned toward where it seemed to be coming from and then weaved around some apple trees until he saw what looked like a woman standing near one of the haystacks. At first glance, he didn't think anything of it – after all, women were always coming by his family's farm and handling business for his father. But then he took a closer look at her attire and noticed that she was wearing pretty strange clothes for an ordinary woman back in this little farming community in southern Kentucky. The woman was wearing black dress pants and a green and yellow plaid shirt as well as a straw cowboy hat. She had very flushed red cheeks and lively brown eyes, but he noticed something else too – clumps of dirt on her boots. The boy looked down at the ground and saw that her boots were covered in mud, so he questioned whether she had ever been to this farm or not. But then why was she here? The boy wasn't sure if he wanted to find out the answer to that one just yet. He knew it would be best to get some help before dealing with such strangers like this one.}
\\
\\
\texttt{So, being a brave boy, the boy tried to gather his courage up enough to go over there and ask her what she was doing on his family's farm – but before he could take his first step toward her direction, the woman spoke up instead. She said in a soft yet somewhat commanding voice while putting on dark gloves: "Where is it? What did you do with it?"}
\\
\\
\texttt{The boy stopped short in his tracks when he heard this woman's words; they seemed strange for someone like her to say them at all. And then what did she mean by "it"? The boy wasn't sure. He looked over at her again and noticed that she was staring at the haystack where he had found his flying-boots earlier that day. It was then that something inside him told him that this woman knew something about his boots – or rather, the golden egg that he had placed inside. This egg was the reason why he took flight and became a dragon for one afternoon, so what would this woman know about it?}
\\
\\
\texttt{He considered confronting her right then and there, but before he could make any decision, the woman closed her eyes very slowly as if she were asleep – or like she was looking around for something else in particular. After a couple of seconds of silence, the woman began to search once more around the haystack with both hands. The boy watched her out of curiosity; after all, there was no one else around to help him find out what this strange woman wanted with his boots – or with his egg for that matter – but just as he hoped to speak up to ask her what she was doing she did so first by saying: "I know you're here somewhere. I can feel it." The boy stared, but what she said next surprised him. She continued her search by saying: "I sense another egg – a golden one, just like the one I'm looking for."}
\\
\\
\texttt{The boy's eyes widened when he heard this. An egg? This woman was looking for a golden egg more than just his boots? He wasn't sure what to think of that one. And then the boy recalled something that his father had told him about many moons ago. He told the boy that some kind of an ancient legend said that one could take flight and become a dragon if they wore these shoes – or rather, these boots – and then placed a gold egg inside them before reciting some particular words about turning into a dragon or something like that, but he couldn't remember the exact words. "I will have to remember to ask my father about it again," the boy thought. The woman continued to search around the haystack for any sign of another egg, which made the boy wonder what she was really doing here in the first place. He was about to ask her that when he heard a bunch of clapping – like his father clapped his hands together – and as if this sound were a cue, the boy watched in awe as the woman then turned into a giant red dragon. The boy had seen dragons before; they were several times larger than him, but he had never seen one transform into a dragon right in front of his very eyes! It was so…wonderful! It was almost like magic!}
\\
\\
\texttt{The dragon flapped her wings once before swaying back and forth while gliding down over to where she had been searching around. She hovered over the haystack, which caused a lot of bits of hay to fly up into her face. This made the dragon sneeze for a moment before she snapped her head towards where the sneeze came from and looked straight at where the boy now stood with both his hands above his head as if he were surrendering himself to this creature that could breathe fire from its mouth.}
\\
\\
\texttt{"Well hello there!" the dragon woman said with a grin before taking a deep breath: "Are you my dinner or something?"}
\\
\\
\texttt{The boy was sure that he had seen this dragon before. He couldn't quite put his finger on it, but he knew that he had seen her somewhere before. She seemed familiar somehow. "Uh…no, um…not yet," the boy said. The fire coming out of the dragon's mouth didn't scare him one way or another, but after seeing how this woman transformed into this enormous beast, he wasn't about to get too close to her for fear for what she might do to him if he did so.}
\\
\\
\texttt{"Not yet? What do you mean by that?" the dragon asked as she landed in front of the boy while puffing out clouds of smoke from her nostrils, which made the boy cough a little bit in response. The beast then continued with: "Well don't just stand there! Come here and tell me why you're here in my home as I am trying to eat." The child felt like he was being bossed around by this dragon; that annoyed him somewhat. "I'm just here to get my boots back, ma'am," the boy said while holding up his right boot.}
\\
\\
\texttt{"What? That's it?" the dragon asked with a somewhat surprised look on her face. She then shook her head while saying: "Why didn't you simply fly up and get them yourself?"}
\\
\\
\texttt{This made the boy blush, but he continued to hold his ground as he responded by saying: "Well…I….uh…I can't fly."}
\\
\\
\texttt{This made the dragon laugh out loud at that. She laughed for a good minute before she finally calms down and looks him in the face. "Hey listen, kid," she said. "Why don't you go on home now? It's not safe here in this forest – especially not for dragons like me who are larger than life."}
\\
\\
\texttt{"But…" the boy wanted to protest, but didn't know what else to say other than: "It's just that I can't fly yet…"}
\\
\\
\texttt{"You mean your father never taught you how to fly?" the dragon asked as she sighed sadly at this realization. The boy nodded as he looked down at his feet while blushing at this revelation. "I see," the dragon said. "Well then I guess I don't have a choice in this matter." The dragon then flapped her wings and took a few steps back to make some room for the boy to take off into the air, which he did with very little effort. It felt like he was riding an invisible swing that took off without him even having to push off from the ground or anything. He simply got on and then it happened naturally on its own, which made him wonder if his father had done something to help him with that too.}
\\
\\
\texttt{"I think so," the boy said as he looked down at his boots that were now back on his feet for the moment. "It would make sense since I always got everything I wanted without any effort," the boy thought with a hint of bitterness in his voice. The dragon then looked up at the boy and smiled. He was quite the handsome looking boy, she thought while watching him flap his arms back and forth like some sort of bird. He had short brown hair with a bit of green in it if you could get a close look at it – and she was very close to him right now! The boy also had blue eyes like hers, so that was another thing that made her smile even more as she watched him fly around without even having to use his hands to propel himself through the air.}
\\
\\
\texttt{"Are you trying to do loops down there?" the dragon asked from her position on the ground. She wanted to laugh again – but this time she held it in for fear that she might scare him off completely if he heard her chuckling like an old hag or something. It was just too funny to see a young boy flying around like a bird!}
\\
\\
\texttt{"Can't I be doing whatever I want?" the boy said as he circled above this woman below who used to be a dragon a moment before. "I mean…it's my dream, right?" he said while looking down at her with a sly smile on his face.}
\\
\\
\texttt{"Yes, but this is my dream too," the dragon said with a brief look of worry on her face before she continued by saying: "I want to be all alone in my forest so that I can eat without any interruptions like you are making right now." She was being harsh with her words, but she couldn't help it. Not after having this dream every day for almost twenty years straight!}
\\
\\
\texttt{"Sorry," the boy said in a soft voice as he found that he had reached his limit of flying through the air and landed back on the ground again – just like he did when he was trying to get his boots back. The dragon then came over and stood directly in front of him, which caused him to take a few steps backward as if he was nervous about something. "What's wrong?" the dragon asked while tilting her head to one side. A few seconds later, she realized what it was that was bothering him as she asked: "Are you afraid of me?"}
\\
\\
\texttt{The boy nodded while taking another step backward toward where his boots lay on the ground behind him. He knew how big this dragon was because he had seen her from up above! She was huge, yet she was kind enough to allow him to fly through the air by using her own body to propel him into the air like some sort of a launching pad that no one could resist taking advantage of because it was just so much fun! That part he liked, but what he didn't like was that there were spikes all over her body – especially on her back!}
\\
\\
\texttt{"Oh, so you're afraid of me," the dragon said as she looked down at where his boots had been lying a moment before. "Well, there's something I can fix easily enough!" The dragon then quickly bent down and picked up one boot in each of her front paws. She brought them closer to her chest so that she could then use her tongue and lick them clean. She licked very fast because she knew it wouldn't be long before the boy would try to grab his boots back from her if what she did didn't work out for him either.}
\\
\\
\texttt{"These taste really good," the dragon said from within an imaginary mindscape in which only two people existed: herself and the boy who wore those special boots.}
\\
\bottomrule
\caption{\rolling{}'s generated story for Premise 2. The story, about a boy who is able to fly and meets a dragon, is completely off-topic for the premise; there is not even any mention of the supposed main character Shannon (who is introduced as female in the premise). However, the story seems reasonably coherent overall.}

\label{tab:premise2_rolling}
\end{longtable}
\end{small}

\begin{small}
\begin{longtable}{p{0.9\textwidth}}
\toprule
\textbf{\rollingfinetune{} Story for Premise 2}\\
\midrule
\texttt{Shannon stepped around the corner and smiled at the elderly man who was whistling as he made his way up the sidewalk.}
\\
\\
\texttt{"Excuse me, sir." She waited to be certain that he was capable of hearing her before approaching. The man was older than she had expected, and judging by his gray hair, he was definitely not a freshman. He stopped whistling as she came closer and glanced at her curiously.}
\\
\\
\texttt{"Can I help you?"}
\\
\\
\texttt{"Yes, please." Shannon extended her hand to stop him from walking away. "I need to ask you something."}
\\
\\
\texttt{The elderly man stared at her hand with a wrinkled brow and then shook it gingerly. "I'm sorry, but I don't have much time to spare this morning," he said through pursed lips. He looked up at the sky as if trying to find more time for himself in a cloudless blue sky.}
\\
\\
\texttt{"Sorry," Shannon said quickly, letting go of his hand in order to put both hands beneath her chin in apology. She caught the distinct whiff of alcohol on his breath when their hands parted and tried not to grimace at the old man's breath before continuing on with her question. "I'm trying to find the engineering building. Do you know where it is?"}
\\
\\
\texttt{"Engineering building? Yes, I think I know how to get there." He smiled at her and then started to walk again. "It's that way." He pointed vaguely down the street. "But I think it's a bit of a walk."}
\\
\\
\texttt{Shannon thanked him and turned in the direction that he had indicated before jogging toward it. Even if he was a little drunk, she was grateful to have directions when she was trying to find one of the most popular buildings on campus. She pulled her cell phone out of her pocket while she walked, dialing up Jake's number with the intention of asking him for help when his voicemail picked up again. Her best friend hadn't been able to go with her this morning because he had been busy studying for another class, but there had been no reason for him not to call her back as promised last night. Shannon tried his number again just in case he'd already decided that he didn't need to check on her anymore because she had been fine last night, but then hung up when it went directly to voicemail once more. "I was just trying to help you out," she said to herself, turning off her phone with a frown. "But I can't even count on you for that."}
\\
\\
\texttt{She opted not to call Amanda this time, having already had her friend throw a fit the night before when she didn't answer her call. The girl could be a bit overbearing at times, and Shannon had decided that it might be better if she took care of herself for once instead of relying on others. She would have liked to at least tell her parents where she was going so they wouldn't get all worried again, but both of them were at work, and they wouldn't allow her to go anyway.}
\\
\\
\texttt{Shannon ducked into the engineering building after finding the right building entrance right beside the stairs. There were several other students waiting outside before entering as well, and she smiled politely when they glanced at her curiously before remembering that it was polite to greet other people if you walked by them. The door was propped open for those waiting to enter, and an older man ushered them inside as he checked their names off a list.}
\\
\\
\texttt{"You're here for your final?" he asked gruffly when he saw Shannon's name on the list next up. "Come on in."}
\\
\\
\texttt{Shannon nodded and slid inside, grateful not to have to wait outside anymore. She quickly found the room that she needed after following a winding set of hallways and taking several wrong turns, which resulted in her thanking more than one person for helping her get back to the main hall after she had turned a corner in the wrong direction.}
\\
\\
\texttt{Shannon was thankful for having studied that section of the manual well before class started, otherwise she would have been lost after getting lost. A short, older woman with gray hair introduced herself as Professor Birks and began to give an introduction about how this was an engineering class and not a magic class. Shannon smiled politely at the students gathered around the lab tables and benches at one end of the room while listening carefully to what the professor had to say. There was a warning about not wearing any loose jewelry while working with tools or chemicals, then another about using safety goggles when using any tools that might cause injury when they weren't supposed to.}
\\
\\
\texttt{Shannon tried to ignore the fact that there were only two other women in the classroom besides herself, sitting together near her at a table on her left side. It was better to not have any distractions, she decided. These people were merely a distraction. They all wore either uniforms or had on jeans, shirts and sweaters that were more suited for the weather outside than what she had on. Shannon had spent almost an hour combing her hair into the tightest bun that she could manage without looking like a topknot. She had also chosen a pair of jeans with a hole in one knee herself, along with her favorite sweater that showed off her curves in all the right places and helped to hide the plumpness of her thighs when paired with leggings. It wasn't proper gear for sitting on a stool and using power tools to cut through pieces of metal like the others around her, but it was what she had decided to wear for this particular occasion as she waited for class to start.}
\\
\\
\texttt{"So we'll be doing a class presentation," Birks explained as she walked between students, handing out pieces of paper that held their names and the topic they would be presenting on written on them. "You'll have an hour to prepare your presentation afterwards." Shannon's eyes widened when she opened hers, seeing that her topic was "Disposal of Used Automobile Parts." She glanced at the others, but they all seemed to be in their own little worlds as they read. Sure enough, the professor handed out a piece of paper with a tree and a picture of gas pouring out of its trunk for the next person to read.}
\\
\\
\texttt{"Since many car parts are made from oil-based plastics," Birks said from the front of the room once everyone had taken turns reading their topics aloud, "it's important that then be disposed of properly after they're no longer viable for use. We'll talk about how to properly dispose of your used parts before you present on this topic next week."}
\\
\\
\texttt{Shannon nodded along with the others as the woman continued to speak about what information would be due at that time and when it would need to be turned in by. She clamped her hands down on her knees as she tried not to look nervous when she could feel some eyes on her, but this was a group project and she didn't have much experience working with other people like this. "You have your topic?" a man's voice said from beside her. Shannon glanced over and looked up at the young man who had asked the question, seeing that he was looking back down at her with a hopeful smile. She smiled back and nodded, then turned back to the professor without saying anything. There was nothing she could think to say, after all.}
\\
\\
\texttt{***}
\\
\\
\texttt{* * *}
\\
\\
\texttt{"One more!" Shannon called out once she heard a clinking of forks against plates. She raised her right hand and moved it around in the air like a conductor, circling it once to see if anyone else would look up before they continued to chow down on their meal. Most of them did, though not very quickly; only two or three others had their heads up and looking over at her as she started waving by the time they started going again. One more reason why this wasn't as much fun in reality as it was in her head, Shannon thought with a sigh as she looked over at where she had been sitting before on a stool next to five others working together on this project. She had been the one to ask them if they wanted to go out to a restaurant together and they'd all agreed. It had been Shannon's idea, after all, but she was almost more nervous than she had been on her presentation.}
\\
\\
\texttt{"I'm really sorry that I was so quiet when we were working on our presentation," Shannon said as their waiter came over and took their plates away. She'd tried to get a few of them talking, but it always seemed like people were happier to talk about themselves than anything else. "I'm just not used to having people around me who are... like me." She raised her eyebrows and looked back at them with a mix of excitement and anxiety as she spoke.}
\\
\\
\texttt{The other girl giggled, shaking her head. "Oh come on," she said with a smile. "It's not like we're going around kicking puppies or anything."}
\\
\\
\texttt{Shannon shrugged and looked down at the table as the rest of the group joined in laughing behind her. The waiter came over and dropped off a glass of water for each of them before walking away. He was a young guy that didn't strike Shannon as the type who would look down on people for being unique, but she had never been good at reading facial expressions. She read lips and expressions on people's faces.}
\\
\\
\texttt{"How'd you get like that?" the girl asked once the waiter walked away again. The other guys in the group were looking to Shannon expectantly as well, though she wasn't sure if it was because they really wanted to know or if they were just trying to make a good impression by seeming polite enough to be interested in what she had to say.}
\\
\\
\texttt{"My dad was really into cars," Shannon said, smiling and looking down at her hands again on top of the table. Her fingers were shaking a little and she tucked them under so that no one could see them shaking. "He worked in an auto repair shop when I was growing up." She chuckled and looked up at them again with her eyes shining a little bit brighter than they were just a minute ago. "Whenever he fixed something I'd be at his side watching, sometimes helping with tiny things." She took in a deep breath and kept her eyes locked with the girl who had asked her the question, the one that she'd chosen to sit next to and started talking to the most. "I always loved it. I thought it was amazing that someone could make a car run again just by taking it apart."}
\\
\\
\texttt{"You've always been like that?" a guy on her other side said, raising his eyebrow one millimeter at a time as he looked over at her. His team had stayed fairly quiet during their time together, but now he seemed interested enough to speak up about what she was saying. Shannon wasn't sure if she liked or hated this guy, though she liked his friendliness toward her well enough.}
\\
\\
\texttt{"I haven't always been like this," Shannon said with another smile. She felt empowered by their interest in what she was saying and decided to share just how unique she was with them despite having only known them for an hour or two. "I... I used to be pretty normal growing up." She chuckled again with a bit of embarrassment and brushed back some hair that had fallen into her face as they all looked at her intently while they waited for more of her story. "When I got older I didn't want that anymore." She shrugged and looked down at her hands again, feeling a little embarrassed about the next part of her story. "I started getting into..." She looked back up at them and swallowed, swallowing hard to get the rest of the words out. "Dominant stuff."}
\\
\\
\texttt{She was looking at each of them as she said it, trying to gauge their reactions. Their eyes widened a little bit and they each took a deep breath after she said it, but they didn't seem disgusted by her admission. They were actually quite interested in what she had to say now and seemed open enough to ask some more questions without making her feel uncomfortable.}
\\
\\
\texttt{"I see," the girl said with a smile that seemed genuine enough that Shannon relaxed slightly. The girl leaned toward them slightly, resting one elbow on the table while keeping one hand on her knee, and flashed Shannon a big grin that made her cheeks flush just a little bit with embarrassment. "So what exactly are your kinks?" She laughed lightly as she spoke and reached up to tuck some hair behind one ear as she did.}
\\
\\
\texttt{Shannon frowned for just an instant before glancing around at each of them again and shrugging with a small smile on her face. "I usually like to take on more than one guy at the same time," she said hesitantly, her eyes darting between each of them a few times before she spoke up again. "And sometimes I like to be tied up. Just..." She looked down at her hands again and chewed on her bottom lip as she thought about how to properly explain it. "Just kind of bound while they do whatever they want with me."}
\\
\\
\texttt{"Have you ever used your safe word?" the boy on her right asked, leaning in slightly so that he didn't have to yell across the table. His team members were starting to look uncomfortable and Shannon almost couldn't blame them for feeling that way.}
\\
\\
\texttt{She grimaced and nodded slightly. "Yeah," she said quietly, smile fading a little bit as she spoke up again. "But I never used my safe word when I was with Remi or Adam." She blushed slightly and waited until her eyes met each one of theirs before continuing on when one of them tried to get in with a question before she finished talking about Remi and Adam.  "I... I trusted them."}
\\
\bottomrule
\caption{\rollingfinetune{}'s generated story for Premise 2. It mentions the character Shannon, but otherwise fails to follow the given premise. There is no clear overarching plot; rather, the story consists of disjointed scenes that seem related only by the fact that they all involve the character Shannon. }

\label{tab:premise2_rollingft}
\end{longtable}
\end{small}

\FloatBarrier

\newpage
\subsection{Examples for Premise 3}

\begin{table}[!htbp]
\small
\begin{tabularx}{\linewidth}{X}
\toprule
\textbf{Premise 3}\\
\midrule
\texttt{Approximately a year ago, 29 year old Payton Parskin was in a fatal car accident. She died instantly and went to Heaven. But, Payton made a deal with God that if he let her back on Earth for just one day, she would spend that day doing good deeds. The day comes and Payton wakes up alive and well. She isn't quite sure what to do with her day but she is determined to make the most of it.}\\
\bottomrule
\caption{Premise for the third group of example generated stories. Another fairly unique premise.}
\label{tab:premise3}
\end{tabularx}
\end{table}

\begin{table}[!htbp]
\small
\begin{tabularx}{\linewidth}{X}
\toprule
\textbf{Initial Setup and Outline for \ours{}, Premise 3}\\
\midrule
\texttt{Premise: Approximately a year ago, 29 year old Payton Parskin was in a fatal car accident. She died instantly and went to Heaven. But, Payton made a deal with God that if he let her back on Earth for just one day, she would spend that day doing good deeds. The day comes and Payton wakes up alive and well. She isn't quite sure what to do with her day but she is determined to make the most of it.}
\\
\\
\texttt{Setting: The story is set in the present day in a small town.}
\\
\\
\texttt{Characters:}
\\
\texttt{1. Payton Parskin is a beautiful young woman with long blonde hair and blue eyes. She is 29 years old and was killed in a car accident approximately a year ago.}
\\
\texttt{2. Landon Shaw is a handsome young man who was Payton's boyfriend at the time of her death. He is 27 years old and works as a mechanic.}
\\
\texttt{3. Maggie Shaw is Landon's sister and Payton's best friend. She is 25 years old and works as a teacher.}
\\
\\
\texttt{Outline:}
\\
\texttt{1. Payton Parskin makes a deal with God that if he lets her back on Earth for just one day, she will spend that day doing good deeds.}
\\
\texttt{2. The day comes and Payton wakes up alive and well. She isn't quite sure what to do with her day but she is determined to make the most of it.}
\\
\texttt{3. Payton spends the day doing good deeds for the people she loves and the people in her community.}
\\
\bottomrule
\label{tab:premise3_ourssetup}
\caption{\ours{}'s initial setup and outline for Premise 3. The characters and outline make perfect sense for the given premise.}
\end{tabularx}
\end{table}

\begin{small}
\begin{longtable}{p{0.9\textwidth}}
\toprule
\textbf{\ours{} Story for Premise 3}\\
\midrule
\texttt{Payton is getting ready to go out on a Saturday night when she hears the news that changes her life. She hears the news in a radio broadcast about a tragic car accident.}
\\
\\
\texttt{There is only one survivor, Payton Parskin.}
\\
\\
\texttt{Payton knows immediately that it's her who has survived. She feels guilty for surviving and wishes that she hadn't. She also wishes that everything would go back to normal so she could be with Landon again.}
\\
\\
\texttt{A year later and Payton still feels guilt for surviving while her two best friends were killed. One of them was Landon's sister Maggie Shaw, who was like a sister to Payton and the only friend Landon had growing up, so they were all very close after high school.}
\\
\\
\texttt{Landon and Payton had been dating for four months before her death, so he was very devastated by her death and has not really moved on or found anyone else since then. He hasn't dated anyone since his heartbreak a year ago, which was around the same time as his sister's death just days before Payton died in the accident. Payton isn't sure how to deal with her guilt. She is just not sure how to get over it, or if she even can get over it. She wishes Landon had moved on from Maggie's death, but he hasn't. Payton hopes that she can at least help him with healing through the guilt of her own death, by spending the day on Earth and doing good deeds for people who need help.}
\\
\\
\texttt{She doesn't want to reveal herself to him necessarily, since she knows he wouldn't believe her anyway, and it's last minute so she has no time to plan anything if he does see her for some reason, but she really wants to make things right for Landon and Maggie by helping others out in the world instead of being selfish and doing nothing but feeling bad about herself.}
\\
\\
\texttt{She needs to ask God for permission first though before returning back to Earth. She asks him if one day is enough time for her good deeds and God says yes because his rules are different than ours. She asks God to let everyone know that she is back here on Earth because they probably won't notice her directly or be able to recognize her. Payton's heart is racing and filled with excitement as she thinks about returning to the place she loved most in the world. She wants to help everyone, because she feels so bad for herself that she can't talk to anyone about how she feels, and all of her pain is bottled up inside of her. She wants a new beginning, so that's why she asks God for one day. Payton asks God for a day and if it isn't enough time for good deeds, then he can just take it back when she goes back but he agrees to give her one day along with a new beginning.}
\\
\\
\texttt{She comes up with this very clever plan to help Landon move on from Maggie's death without him knowing that it was her who has helped him move on. Her first stop will be at Landon's house because Payton always loved seeing him there and his mom is the only person who needs help right now since Landon's dad died three years ago in Afghanistan.}
\\
\\
\texttt{Payton walks the short distance over to the house and sees Landon working on his motorcycle in the driveway. He looks tired and sad, but happy to be working on his motorcycle at least which seems like a good thing. That's when she feels it. It feels like there's someone behind her and she turns around quickly, but no one is there. Someone is following her, but who? She looks around eagerly and sees a small young boy who appears to be about 10 years old crossing the street about half a block away from her. He is wearing a pair of blue jeans and brown shoes with a white shirt and has long brown hair that he just recently cut with bangs that cover the top of his eyes. He still has braces on his teeth, so it looks like it hasn't been too long since he had them put on. Payton can tell that the boy is not afraid of being caught because he smiles at her as he crosses the road towards her. He moves very slowly since he isn't looking at where he is walking because he keeps his eyes trained on Payton as if waiting for something to happen right away.}
\\
\\
\texttt{Payton doesn't know the boy, but she knows that she knows him even though they haven't met yet. She can feel the familiarity in her heart as if they were supposed to meet each other already so many times already but haven't been able to find each other until this very moment right now. He glances around when he gets off the street and walks towards her.}
\\
\\
\texttt{"Hi," he says. "My name is Dylan."}
\\
\\
\texttt{"Hi, Dylan. My name is Payton."}
\\
\\
\texttt{Dylan looks at her and smiles.}
\\
\\
\texttt{"You look familiar," she says. "Have we met before?"}
\\
\\
\texttt{Dylan nods and smiles even bigger now that Payton has said something similar to him. It feels like he doesn't have a mirror in front of him, or perhaps it's because this is the first day of his life for which she has died and come back to Earth because there is no way that he would have known about her already if that were the case. She can feel his excitement coming from inside of him like steam rising up from a hot cup of coffee freshly made. He holds out his hand to her eagerly and she moves closer to him so that they can shake hands while they both smile at each other like two old friends who have just bumped into each other after years of not seeing each other again after many years apart. "It's so nice to meet you," he says. "I've heard a lot about you."}
\\
\\
\texttt{Payton smiles at him and feels relieved that someone else has said the same thing that she had said to Maggie earlier. Maybe it was just one of those things that Payton had just felt, or perhaps Payton and Dylan were always supposed to meet each other here on this very day and it was destiny that brought them together, just as it was destiny that brought Maggie back after all of these years with an entire new life for her.}
\\
\\
\texttt{"It's so nice to meet you too," she finally says and then smiles some more at Dylan. "I'm so glad to see you now."}
\\
\\
\texttt{"Are you really?" Dylan asks her, sounding slightly surprised by her words.}
\\
\\
\texttt{She nods in response and then he asks her what she is doing here on this day but she doesn't have an answer for him because she is not sure why she is alive again or what this particular day even has meaning for her anymore. She is still working on issues from the previous life that have not been fixed yet; perhaps they will never be fixed while she remains on Earth again the second time around like this. Dylan is looking at her, his eyes staring into hers as if he can see through her, and she watches him as if he is an angel and she can't quite believe it.}
\\
\\
\texttt{"I'm here because I've never even met you before today," she tells him and then she starts to explain everything to him in more detail this time. "Do you know who I am?"}
\\
\\
\texttt{Dylan nods his head but still doesn't say anything out loud. He is smiling at her and watching her intently as she continues to speak. She wonders why they are standing here without saying anything but it feels a bit awkward, like they are both waiting for something to happen or perhaps they are both waiting for the other one of them to say something first.}
\\
\\
\texttt{"I don't know what you're supposed to do with your day but I was wondering if you would maybe want to spend some time with me," Payton finally says. "I mean, we could walk around, hang out together and maybe watch the sunset later on tonight?"}
\\
\\
\texttt{He nods again in response but doesn't answer verbally because he seems to be lost in thought or perhaps he is hesitant about something too. He is frowning a bit and looking down at his feet as if he is too shy to look directly at her.}
\\
\\
\texttt{"I realize this might seem weird to you," Payton continues on after a brief pause, "but I really don't know what else to do and I have this day ahead of me that I am determined to make the most of."}
\\
\\
\texttt{She pauses again as she looks over at him, noticing that he isn't really doing much of anything so she decides she may as well just leave him alone now. She turns around and begins walking away from the cemetery. She is only able to take a few steps before he suddenly catches up with her and eventually walks beside her instead. She smiles pleasantly at him but he doesn't return the gesture. He seems so sad today, like nothing ever makes him happy anymore.}
\\
\\
\texttt{The two continue on like this for some time, walking along side by side but neither one of them speaking out loud or saying much of anything significant to one another. Payton isn't quite sure what she should do with herself but she decides that if this is how it's going to be, then fine…that's just fine with her….whatever….it doesn't really matter anyway…this is heaven, right? That means there's no reason to have a problem, right? Right….}
\\
\\
\texttt{As they walk along, she begins to notice that they are beginning to cross through a poorer area of town and as they pass several homeless people sitting on the corner begging for money and food she realizes that this is an area she isn't all too familiar with. A sense of dread begins to clutch at her heart but she thinks, "It's not all that bad. Probably just some people who are down on their luck in the worst way. I can't possibly expect anyone to help everyone who needs it…and besides, it's not like I'm actually here or anything…I'm just dreaming." She continues on like this for about another ten minutes before she realizes that Dylan is no longer walking beside her like he was before but rather walking ahead of her instead as if he is conducting some sort of tour.}
\\
\\
\texttt{About fifteen minutes later, Payton notices that they are now standing in front of a house that looks somewhat familiar. It looks exactly like the house where she grew up back in Seattle and looking around at everything else around her, Payton realizes that they aren't in Seattle anymore. They aren't even on Earth anymore! They have crossed over into Heaven! Payton is speechless as Dylan pulls her up to the front door of the house and rings a bell that has the word "FAMILY" emblazoned across it.}
\\
\\
\texttt{Payton looks around at everything in pure awe, but she can't believe that he is taking her here! This is where he grew up! Why would he bring her here? Her heart begins to pound loudly in her chest and she thinks, "What's happening?! I don't understand what's going on right now! What's happening?! What is this?!"}
\\
\\
\texttt{Dylan turns around, looks into Payton's eyes and says, "Don't worry. Everything will be fine." He opens the door before he continues on and begins to walk inside with Payton still standing frozen outside as if she were in shock. She feels confused, scared and hurt all at the same time but there are no tears streaming down her face like she would have expected them to. No…instead there is something different about the way a person looks once they pass into this realm…it seems as though something hardens them on the inside like a shell or a suit of armor that protects them from outside elements until they actually need it. But right now, Payton needs to be a big girl and she needs to do this. She will not push away the hope of finding her friend. She knows that she won't find him on Earth.}
\\
\\
\texttt{The house inside is much larger than Payton expected it would be and there are beautiful paintings on the walls in golden frames. The air inside is heavy with vanilla and cinnamon smells.}
\\
\\
\texttt{Dylan suddenly takes Payton's hands into his own and says to her, "You're safe here, Payton." He smiles at her warmly but Payton feels uncertain as he says this. Yes, she is safe…but she doesn't feel safe here. It isn't home no matter where it may be! She wants to leave as quickly as possible…but there is something comforting about his presence that makes her feel like she can maybe do this after all. She thinks, "I hope he doesn't know that I like him. I hope he doesn't know that I am in love with him."}
\\
\\
\texttt{A woman opens the door then, completely catching them both by surprise, but Dylan presses two fingers to his lips in order to silence any questions that might erupt from the situation. When the woman sees Dylan's eyes however, her own widen with recognition and for a moment she looks at him with no expression whatsoever before bursting into tears of joy and running into his arms faster than Payton could have ever imagined anyone moving on Earth! "My baby! My Dylan!" She cries as she embraces him. Payton suddenly wonders if this woman is Dylan's mother and she feels a little embarrassed by the fact that he had been able to forget about her. He must have loved his mother so much.}
\\
\\
\texttt{Dylan smiles and holds her tightly before suddenly pulling away from her and saying, "Mom! I think we might have a visitor." His eyes move over to Payton and even though she just met the woman, she feels like they are old friends; like they met many years ago at the grocery store or at their children's school, even though it was not long ago that they had ever met at all.}
\\
\\
\texttt{The woman looks over at Payton with great curiosity and then says warmly, "It's so nice to meet you!" She steps forward towards Payton and holds her hands in her own before pulling her into an embrace as well. She doesn't seem surprised in the slightest by anyone or anything…even when Dylan kisses Payton on the forehead in front of everyone as if it were natural for them to be together on Earth like this.}
\\
\\
\texttt{What is going on? What does he think he is doing? She can't believe what is happening but she isn't about to stop him either.  She has waited so long for his kiss.}
\\
\\
\texttt{"It's nice to meet you too," Payton says finally, her voice shaking a little bit. "I'm Payton."}
\\
\\
\texttt{"Yes, I know," the woman says with a smile, as if she has known Payton forever. "I'm sorry, I don't think we've been formally introduced. I'm Dylan's mother, Sarah."}
\\
\\
\texttt{"It's very nice to meet you, Sarah," Payton says, still in disbelief that this is actually happening. She looks over at Dylan and he just smiles back at her before taking her hand in his own and leading her inside the house.}
\\
\bottomrule
\caption{\ours{}'s generated story for Premise 3. The initial deal with God is described seemingly as a flashback after the first few paragraphs, which is somewhat confusing narration-wise, but the first half of the story seems otherwise faithful to the premise. The second half of the story is somewhat disconnected from the first half: she meets a boy named Dylan and the story appears to imply that they go back to heaven together. }

\label{tab:premise3_ours}
\end{longtable}
\end{small}

\begin{small}
\begin{longtable}{p{0.9\textwidth}}
\toprule
\textbf{\rolling{} Story for Premise 3}\\
\midrule
\texttt{It was a beautiful morning in July. The sun was shining, birds were singing and the air smelled sweet. It was a perfect day for Payton Parskin to die. Payton, wearing her favorite black and pink tank top, black shorts, and flip-flops with her long brown hair in disarray, felt oddly peaceful. Her smile was genuine as she thought about the life she lived and the things she had enjoyed doing over the past 29 years of her life... eating ice cream on hot summer days, listening to music with her friends in high school and going to see her favorite band in concert every year until they broke up right after college graduation so that's when she started going to concerts alone. She thought about all of her favorite colors and foods but mostly she thought about love... family love and best friend love. All of these things made Payton feel good inside but there was one thing missing from this whole equation... A man to share all these adventures with!}
\\
\\
\texttt{Payton stared up at the clear blue sky before silently closing her eyes for what would be the last time... if only she knew it at the time. She reflected on a lot of things, especially her life with her family. She thought about her parents, Rachel and John and her older brother, Jason. Payton couldn't help but smile when she thought about how at one time in her life, she was the "favorite". She was always daddy's little girl until Rachel came along... then Payton became the "middle child" and had to fight for the attention of her parents.}
\\
\\
\texttt{Payton took a deep breath as to not cry on this perfect day. She felt so happy being alive and being able to say goodbye to everyone. She could finally let go of all of the anger that she had for Rachel who was such a "perfect" little girl and was loved by everyone as well as feeling sorry for herself because nobody would ever love or understand Payton like they did Rachel! Her eyes still closed, Payton started listing all the things she had been thankful for this past year; beautiful sunny days like today, delicious food that made her mouth water but most importantly, having a real friendship with someone who understood Payton just as well as she understood them.}
\\
\\
\texttt{She knew that she would have to say goodbye to Billie soon too since they were both going different directions in their lives! Payton's heart suddenly felt like it was going to burst. For the first time in her life, she felt lonely. She didn't have a boyfriend to hold her hand when she needed someone to lean on and she definitely didn't have anyone who would be there to cry with and laugh with. She couldn't even do that with someone who was like a sister to her... Billie.}
\\
\\
\texttt{Speaking of Billie, Payton had an idea. Why not go to see Billie and find out how things were going in London? Payton could shed some light on the situation with James' family which would then give Billie hope that she could make things right... just like Payton was going to do here back home! Then she could stay longer in England while they figured everything out! It sounded like a great idea... except for one thing... Payton never told anyone that she was alive. She figured everyone thought that this fatal car accident last month was just too hard on everyone so why keep them all feeling miserable about it? It's not like Payton would ever see them again anyways so what would be the point of telling them now...?}
\\
\\
\texttt{"I love you guys..." Payton said aloud as the tears started flowing from her eyes. She didn't think that she would be able to live without her family. Rachel, Jason and John were the first people she had ever loved in her entire life. She didn't think that she would be able to live without seeing them again or feeling their hugs or being comforted by them when times got rough... Or even worse... being angry at them for something they had done or not done!}
\\
\\
\texttt{Suddenly, Payton could feel herself being pulled from the ground and she could hear a familiar voice whispering into her ear but couldn't make out what it was saying. She felt so warm and peaceful as everything around her seemed to go into slow motion. She felt herself floating through the air before landing gently on something soft. The warmth and peace suddenly disappeared as Payton opened her eyes to find herself staring at a woman who looked like an angel! Her long auburn hair was tied back with a white ribbon and she was wearing a powder blue gown with beautiful lavender flowers on it. She held out her hand so that Payton could hold onto it as they flew through the dense fog surrounding them... in slow motion once more! Payton felt like she was in some sort of daze but then the image of the beautiful angel faded away into nothingness... leaving only darkness behind! Payton felt a sharp pain in her chest as she opened her eyes to see that she was in a hospital room. The first thing she did was look down at herself to see what was hurting her so badly. Her head and shoulders were covered in bandages and there were tubes inserted into her arms! She noticed someone standing off to the side with their back facing her at the end of the bed. "Billie?" Payton said softly, knowing that it was only a dream but she wished that it could have lasted longer.}
\\
\\
\texttt{"Payton?!" Billie said as she turned around quickly, almost tripping over her own feet. "I'm sorry... you gave us quite a scare, you know?" She added as she walked towards Payton's bedside to take her hand but when she noticed all the bandages, tubes and the fact that Payton's arms were covered by white sheets... Billie knew that it must have been so painful for Payton to even move! "I'm so sorry..." Billie whispered again, not really knowing what else to say but feeling guilty for not being here for Payton when she needed someone most!}
\\
\\
\texttt{"Oh no! It's not your fault." Payton said as tears streamed down from both of their eyes. "I wasn't in an accident... I... I tried to commit suicide." She added as she was still holding on to Billie's hand.}
\\
\\
\texttt{"What?!" Billie exclaimed in shock as she pulled her hand away from Payton's. "Why on earth would you do that?!" She asked, knowing that Payton was someone who always looked on the positive side of things and always knew how to cheer her up when she was feeling down.}
\\
\\
\texttt{"I just couldn't stand it anymore! I just couldn't take it Billie! Please don't be mad at me but... my parents were murdered, my brother is missing and my sister is probably dead too!" Payton cried as tears streamed down her cheeks. "I was raped and abused by the man who claimed he loved me and then had his friends kill my boyfriend because he wouldn't tell me where he buried James' body so I could get back the necklace he gave me." She added as she wrapped her arms around herself, trying hard not to cry but also not wanting Billie to see her like this! "If that isn't a good enough reason for anyone to commit suicide then I don't know what is!" She cried, feeling ashamed of what she had done!}
\\
\\
\texttt{"You are coming home with me... and we're going to get you the help that you need." Billie said as she clutched onto Payton's hand tighter than before.}
\\
\\
\texttt{"No... I'll be okay. I just want to rest now." Payton said softly, not wanting Billie to worry about her anymore. She was worried that if Billie knew how bad off she really was then she'd want nothing to do with Payton because it would be too much for her to handle! And that was something that Payton didn't want! She wanted Billie to love her and take care of her instead of letting her go... no matter what!}
\\
\\
\texttt{"I'm not going anywhere until you have a clear head." Billie said as she reached for a cup of water sitting on the table next to Payton's bed, holding it out for her to drink. "Here, this will help calm your stomach down." She added as Payton drank the cold water down in fewer sips than usual but after a few more sips, the pain in her stomach began to subside slightly. She knew that if she were to stay with Billie, then she'd be in pain most of the time and so would Billie. It was best by far for everyone involved if Payton stayed away from her and found someone else to take care of her. "You can't do this on your own Payton! Please don't ask me to just leave you here when you need me the most." Billie pleaded as she looked into Payton's blue eyes, hoping that it would change Payton's mind.}
\\
\\
\texttt{"I know... I'm sorry... but right now I just want to rest." Payton said sadly as she turned on her side and closed her eyes, wrapping her arms tightly around herself while tears rolled down from both cheeks at the thought of having to live without Billie in her life!}
\\
\\
\texttt{"Okay... but I won't leave until you fall asleep... that's a promise." Billie said as she lay down next to Payton, pulling up a chair close by so that she could stay right next to her all night long!}
\\
\\
\texttt{If You Only Knew by Joanne Barnes ~*~*~*~*~}
\\
\\
\texttt{"It was a dark and stormy night. The kind of night where the sun had just set but the moon hadn't come out yet." Billie said softly as she stood at the foot of Payton's bed, her arms wrapped around herself as she fought to keep from crying. "I was walking by an old abandoned hotel when I saw a woman jump off of a nearby bridge and into the river beneath." She added as she began to cry softly, knowing that this was a dream and that it was going to take some time for her to get over the fact that Payton had tried to commit suicide. "She looked so young... almost like she was in her teens. She had long black hair, blue eyes and beautiful tan skin..." Billie sobbed as she pictuted Payton's looks in her head. "I... I'm sorry Payton..." She cried softly as all of the flashbacks began, causing her pain inside of her heart all over again!}
\\
\\
\texttt{"It's okay... I understand." Payton said softly as she reached out for Billie's hand with both of hers, staring down at their hands together before looking into Billie's tear filled hazel eyes. "You cant do this to us Payton... please don't ever do this again!" Billie cried softly as she pulled Payton's hands up to her lips, kissing them over and over again.}
\\
\\
\texttt{"I won't if you promise me something." Payton said as she wrapped her arms around Billie, hugging her tightly.}
\\
\\
\texttt{"Anything..." Billie said as she tried hard not to break down in front of Payton. She didn't know how much more of this day she could stand!}
\\
\\
\texttt{"Please don't leave me..." Payton whispered softly as her body began to shake from the pain inside of her heart. "Please Billie... I love you so much and I can't ever live without you." She cried softly as tears rolled down from both cheeks at the thought of having to live without Billie in her life! She couldn't live with herself knowing that she had hurt someone else just like she had been hurt all those years ago! It was something that she would never be able to live with herself for doing and now it was too late for her to change anything! "I'm so sorry..." She whimpered as she took a deep breath, trying to get control of her emotions even though everything inside of her was telling her that it was too late to change anything now!}
\\
\\
\texttt{"I'll never leave you... I promise." Billie said as she lifted Payton's chin up gently so that their eyes could meet. She knew that Payton would need help from a doctor in the months to come and if they were going to go through all of this together, then they would do it right! "And I love you... always have and always will." She added as she wiped away Payton's tears with the back of her hands and smiled.}
\\
\\
\texttt{After a long hug, Payton fell asleep in Billie's arms just before 3 o'clock in the morning. Neither one of them had slept since Sunday when they left the hospital where Buck had died and both of them needed some rest before they could start their lives over again. As Billie continued to hold Payton in silence, she could feel the wetness on her cheeks but she didn't care at all. These were happy tears for now and that was enough for her! If You Only Knew by Joanne Barnes}
\\
\\
\texttt{~*~*~*~*~}
\\
\\
\texttt{"You know, I can't believe we are finally here." Billie said softly as she looked out the window at the beautiful scenery that was around the area. "It's beautiful out here Payton... I love it here already!" She added with a smile on her face. She had never seen anything like this before and she knew that her life was going to be different from now on and she wouldn't have it any other way! "Do you really like it here?" She asked Payton as they pulled up outside of their new home and sat in silence for a moment, taking in all of the scenery and enjoying themselves. "We really have a beautiful home Payton... everything is so peaceful." Billie said as she closed her eyes and sighed softly. "I'm so glad your parents left us this house." She added before opening her eyes again to look at Payton and see what kind of reaction she had gotten from saying that. She knew how much closure this place would give them and she knew that if Miss Lissy or Mr. John were still around, they would agree completely!}
\\
\bottomrule
\caption{\rolling{}'s generated story for Premise 3. Payton apparently attempts to commit suicide, and the remainder of the story involves her friend Billie trying to help her get over her trauma. Beyond the presence of Payton, it seems wholly irrelevant to the premise, and there are also jarring oddities in the narration in the places where ``\texttt{*}'' appears.}

\label{tab:premise3_rolling}
\end{longtable}
\end{small}

\begin{small}
\begin{longtable}{p{0.9\textwidth}}
\toprule
\textbf{\rollingfinetune{} Story for Premise 3}\\
\midrule
\texttt{Bilgravia. A country within a country. Well, maybe that’s not quite right. It was a city within a country. A place where the rich and the powerful spent their money and their time. The most renowned restaurant in all of Bilgravia was The Golden Dragon, where Payton had her last meal before she died.}
\\
\\
\texttt{Payton stood outside the building, winding up to take a deep breath before she went in to face her past self's memories face-to-face with her present self for the first time ever. It wasn't just so she could come to terms with what had happened to her, but also so she could get used to being alive again and start remembering how living felt instead of how dying felt when you died in your twenties . Unsurprisingly, she found herself outside a restaurant in the middle of nowhere. She stood still for a long time just looking up at it, at the signs that hung from its front doors. The Golden Dragon seemed to be the only building on this street and Payton could see its place from her perch across the street. It was an almost deserted road. No cars traveled down it and the few people who were out seemed to be going about their business quickly, not wanting to be seen by anyone else. The air had been muggy since she'd come back and that just made her feel like something bad might happen here.}
\\
\\
\texttt{She put her earbuds in and switched her phone to run through its music library randomly, playing one song after another until she got sick of them all before coming back to them. She turned away from The Golden Dragon so she wouldn't have to look at it anymore and took a walk around downtown Bilgravia. She didn't know what else to do and, really, what else could she be doing? She was stuck here in a city where she knew no one and nothing and hadn't been to in over a decade.}
\\
\\
\texttt{She took the opportunity to look around at her surroundings. The few people that were outside seemed to be heading towards the river, which Payton knew from photos that ran along the top of this street. The river was there for the rich people of Bilgravia, who came here for summers during her time and before in her own time. They had their own community, which was separate from the rest of Bilgravia, founded as it was by rich and powerful families who wound up taking all the money with them when things fell apart. Payton could see signs of their golden years in the buildings on either side of her. She tried not to think about how many billions these buildings were worth but couldn't help but do so anyway.}
\\
\\
\texttt{"I mean I'm in fucking Heaven," said Payton out loud without thinking about it too much until she remembered that nobody was around to hear her say it except herself. "This is all going to be mine, maybe."}
\\
\\
\texttt{The words fell out of her mouth before she had a chance to tell herself to stop talking. She wasn't any less crazy than she was yesterday when she'd realized that putting her body and mind back together was going to take more time than Paddy's family had left and when they all left. "And who, exactly, are you supposed to be? You're not Payton, not anymore," echoed her own voice in her head. She tried as hard as she could not to think about it.}
\\
\\
\texttt{She had an urge for a cigarette but couldn't find any in Starry-something's apartment. He told her he didn't smoke but he kept packs of cigarettes on the counter just in case someone needed one so he wouldn't have to go out and buy them for them every time somebody wanted one. It was also a good way for him to get rid of unwanted guests as well as help people bribe their way into his house whenever they wanted something from him--and there were plenty who wanted something from him even though he didn't seem like the kind of guy who would want anything from anyone else. After a few hours, she decided she'd had enough of looking at the buildings and went back to The Golden Dragon. She took little steps so that she wouldn't feel like she was in a hurry or doing anything wrong, even though nobody was around to see her--unless they were watching her on camera through the cameras on the street corners. Maybe it would look suspicious if she ran down this street and then suddenly stopped in front of one of these buildings? What would they think?}
\\
\\
\texttt{"I'm not going to steal your car or anything," Payton told herself and then laughed. She wasn't doing anything dangerous or even illegal so who would care if she was going slower than normal? She just wanted everyone to know that yes, I'm alive and I'm here and I have every right to be here if you're wondering about that.}
\\
\\
\texttt{"No you don't," came another voice in her head.}
\\
\\
\texttt{"Shut up," said Payton, as much as she had when there someone standing within earshot of her. She'd take any opportunity to treat herself better now that she knew who her captors were going to be over the next few months before Ivy could get back again. "You're not going to fuck with me anymore."}
\\
\\
\texttt{She knew it was childish to say so, but she didn't care. It wasn't like anyone could hear her. She was just saying the things in her head out loud and nobody was around to hear her.}
\\
\\
\texttt{"Yes, I know, you've said that before," said the voice, who'd at least acknowledge her this time. "But I don't think you mean it yet."}
\\
\\
\texttt{"I do mean it," said Payton. "I'm going to leave your body as soon as possible and then I'm going to find a way back home and everything will be fine."}
\\
\\
\texttt{"That's what you say now," said the voice in the back of her head.}
\\
\\
\texttt{"It's when I say it for real," said Payton, imagining that she had a microphone held up to her mouth so that everyone could hear her say it where ever they were watching from. "So I'll say it again: when I finally leave your body, and whenever Ivy gets here with that cure or whatever they're going to use on me, everything's going to be fine."}
\\
\\
\texttt{She'd walked a couple more blocks by this point and was standing in front of The Golden Dragon Hotel and Casino again. There was nobody around, the sidewalks empty and dark. These people shut off all the lights in their buildings when they go to bed, Payton thought to herself. They don't trust nighttime, so they turn off all the lights.}
\\
\\
\texttt{She looked at the building in front of her and laughed because it was exactly like she had pictured it would be: tall with a radio tower on top, tall enough so that nobody could touch it if they wanted to, but low enough for everyone to see their shadows against the walls as they went about their daily lives. It was a big building but there was nothing special about it except for its height and what you could do on top of it in those tiny chairs that were barely strong enough to hold up your weight if you tried sitting in one--even though she had never been on top of this building before.}
\\
\\
\texttt{"This is just an illusion," said The Voice, who'd been quiet for a while now. "You haven't seen what's behind these walls before."}
\\
\\
\texttt{"And I don't have any reason or intention to look either," said Payton. "I'm not here to spy on anyone."}
\\
\\
\texttt{"Maybe not just yet," said The Voice. "But you'll be here long enough and you'll have time." "I won't," said Payton. "Why would I?"}
\\
\\
\texttt{"Because you'll be stuck here for another couple of months," said The Voice, who was probably trying to be funny because he had stopped talking for so long.}
\\
\\
\texttt{"If it's so bad here, why don't you just leave me?" Payton asked. "If it's so horrible and awful, then why are you even still here?"}
\\
\\
\texttt{"I can leave anytime I want, but you're not ready yet."}
\\
\\
\texttt{Payton laughed at that and imagined how stupid she must've looked. She was in Las Vegas standing in front of a huge building and laughing all by herself. That would definitely draw some attention! And yet… there was no one around--the sidewalks were empty, the lights were on but nobody was home--so she didn't have to worry about anyone seeing her howl with laughter over here by herself. It would just be her and whatever cameras the government had tucked away inside eaves of these buildings somewhere recording her every move but not showing anyone else what she did.}
\\
\\
\texttt{"You're an asshole," said Payton to The Voice as she looked around again for a camera--even though there probably wasn't one where she was looking. "I am, aren't I?" said The Voice.}
\\
\\
\texttt{"But you love me in spite of that."}
\\
\\
\texttt{"You know I do," said Payton, who couldn't help but smile as she said it. She was probably smiling too big and too wide because people would be able to tell that she was crazy from a mile away.}
\\
\\
\texttt{"It's your choice," said The Voice. "But I think you'll end up staying here for a couple of months whether you like it or not. It's best if you just accept that now instead of fighting it until everyone's ready for you to leave."}
\\
\\
\texttt{"Ivy will get me out soon enough," said Payton, who could feel herself growing weary again from all the walking that she'd been doing tonight, but still pressed on with her search under the warm Las Vegas sun.}
\\
\\
\texttt{"We'll see," The Voice replied and then there was nothing else coming back to her in response until the city streets had turned into desert plains and she was surrounded by sand dunes rising high enough above her head so that they blocked out the sunlight completely and gave her nothing but a moonlit sky full of stars as she lay on her back and looked up at them sparkle before falling back to sleep. ***}
\\
\\
\texttt{Payton woke up to the sound of her phone going off. She could hear it ringing in her ear as she tried to get her bearings and realize where she was. It was dark, but not as dark as it had been before, so she figured that she'd somehow fallen asleep again and missed a night while she was out here searching for Ivy. She looked around to see if there was any sign of her friend, but there was nothing other than a still desert around her and the small stone building sitting in its corner looking like it hadn't changed or moved at all from when she last saw it. It didn't look any different from what she imagined it would look like when they were hanging out in Vegas together, so she just assumed that Ivy hadn't found this place yet--or maybe this place hadn't been here at all when they were hanging out. Better find something else to think about other than this place, Payton thought to herself as she got up off the ground again and looked around for her phone.}
\\
\\
\texttt{This time--the first one since coming back to life--she didn't have an awkward time remembering how to stand up or move around. This time, it was as if she'd never left. But then again, she had never left. Everything was still in the same place that she'd left it, and now that she thought about it, this whole desert was just an extension of the city--it was just another part of Las Vegas' desert with no actual difference between the two except that Payton hadn't come out here before and had only found this place once she'd died.}
\\
\\
\texttt{Where is Ivy?! Payton thought to herself in her mind as her eyes scanned the low-lying hills around her for any sign of life. She heard her phone ringing again in her pocket and pulled out her phone to see who was calling. She wasn't surprised when it was Ivy on the other end, though. She answered the phone without saying anything into it first. "Have you found anything?"}
\\
\\
\texttt{"No," said Ivy on the other end of the line as if she knew exactly what Payton was thinking about even though they weren't even together any longer. "Not yet."}
\\
\\
\texttt{"There's got to be something around here," said Payton. "You kept saying that everything has a reason for being here." "I know," said Ivy. "But this place is so big that it's hard to believe that there's not anything here."}
\\
\\
\texttt{"Maybe there is," Payton said. She turned her head to look back into the stone building that looked like a dead end, but couldn't stop herself from thinking about it for a minute before saying, "but I don't think so."}
\\
\\
\texttt{"Don't do that," Ivy replied. "I don't really want the voice to get inside your head again."}
\\
\\
\texttt{"But it's all I can think about," said Payton.}
\\
\\
\texttt{"If you're alive then you should be able to come up with other things to think about while you're walking around this damned desert and looking for something new."}
\\
\\
\texttt{"It just seems like us coming out here was a waste of time if we can't find anything--or if there was anything here and we just didn't find it yet."}
\\
\\
\texttt{"Why are you saying this?" asked Ivy on the other end of the line. "What am I doing out here? What are we doing out here?"  "I don't know," Payton replied. "I guess it just feels like we're not really accomplishing anything and that maybe we should have just stayed in Vegas."}
\\
\\
\texttt{"We're not going to find anything if we just stay in Vegas," said Ivy. "There's nothing there for us anymore." "I know," Payton said with a sigh. "It's just--"}
\\
\\
\texttt{"Just what?"}
\\
\\
\texttt{"It feels like this whole thing was a mistake and that we should have never come out here."}
\\
\\
\texttt{"Payton," Ivy said with a tone of exasperation, "you know that's not true."}
\\
\\
\texttt{"I know," Payton replied. "But it doesn't make it feel any better."}
\\
\\
\texttt{"Just keep looking," Ivy said. "We'll find something."}
\\
\bottomrule
\caption{\rollingfinetune{}'s generated story for Premise 3. The story is extremely confusing. It begins with Payton having died, but it is unclear if she is in heaven or back on Earth. Most of the story consists of Payton's internal dialogue with ``The Voice,'' and later on a conversation with her friend Ivy. }

\label{tab:premise3_rollingft}
\end{longtable}
\end{small}

\newpage
\subsection{Examples for Premise 4}

\begin{table}[!htbp]
\small
\begin{tabularx}{\linewidth}{X}
\toprule
\textbf{Premise 4}\\
\midrule
\texttt{After her mom's death, Shannon is left to care for her younger siblings. It's a lot of responsibility for a teenager, but Shannon doesn't mind. She knows her mom would want her to take care of her siblings and make sure they are happy. Shannon does her best to make sure her siblings are taken care of, but she can't help but feel like she's missing something. When Shannon meets a boy at school, she starts to feel like maybe she's finally found what she's been looking for.}\\
\bottomrule
\caption{Premise for the fourth group of example generated stories. A relatively mundane premise.}
\label{tab:premise4}
\end{tabularx}
\end{table}

\begin{table}[!htbp]
\small
\begin{tabularx}{\linewidth}{X}
\toprule
\textbf{Initial Setup and Outline for \ours{}, Premise 4}\\
\midrule
\texttt{Premise: After her mom's death, Shannon is left to care for her younger siblings. It's a lot of responsibility for a teenager, but Shannon doesn't mind. She knows her mom would want her to take care of her siblings and make sure they are happy. Shannon does her best to make sure her siblings are taken care of, but she can't help but feel like she's missing something. When Shannon meets a boy at school, she starts to feel like maybe she's finally found what she's been looking for.}
\\
\\
\texttt{Setting: The story is set in a small town in the US.}
\\
\\
\texttt{Characters:}
\\
\texttt{1. Shannon Matthews is a teenage girl with shoulder-length brown hair and brown eyes. She is of average height and build. Shannon is a responsible and caring person who loves her family very much.}
\\
\texttt{2. Alex Matthews is Shannon's younger brother. He is nine years old and has blond hair and blue eyes. Alex is a very energetic and playful child who loves to be outside.}
\\
\texttt{3. Olivia Matthews is Shannon's younger sister. She is six years old and has brown hair and brown eyes. Olivia is a very shy and timid child who loves to read books and play with her dolls.}
\\
\\
\texttt{Outline:}
\\
\texttt{1. Shannon's mother dies, leaving Shannon to care for her younger siblings.}
\\
\texttt{2. Shannon meets a boy at school and starts to feel like she's finally found what she's been looking for.}
\\
\texttt{3. Shannon has to figure out how to balance her responsibilities to her siblings with her own needs and desires.}
\\
\bottomrule
\label{tab:premise4_ourssetup}
\caption{\ours{}'s initial setup and outline for Premise 4. The characters and outline make perfect sense for the given premise.}
\end{tabularx}
\end{table}

\begin{small}
\begin{longtable}{p{0.9\textwidth}}
\toprule
\textbf{\ours{} Story for Premise 4}\\
\midrule
\texttt{Shannon Matthews is a senior in high school. She also happens to be a big sister.}
\\
\\
\texttt{Her mother, Eleanor Matthews, died in a car accident when Shannon was sixteen years old. She was only seventeen herself when she found herself in the position of having to manage her younger siblings, including her little brother, Alex and her younger sister, Olivia.}
\\
\\
\texttt{After her mom's death, Shannon moved in with her grandparents and she took charge of the house. She kept up with the chores—making sure that the house was clean and all the kids got enough to eat—and she also worked at a local diner where she made enough money to look after everyone as well as buy them some new clothes and all the necessities they really needed.}
\\
\\
\texttt{Shannon knows that caring for Alex and Olivia isn't easy for them either since they are so young. They were still babies when their mother passed away. They were only five and six at the time of Eleanor's death. They didn't understand what happened or why their mother wasn't coming back home anymore; they just knew that their mommy wasn't around anymore. Still, they need what they need—and that also means needing their big sister. They miss her and they are happy to have her around, but it isn't easy for all of them. Shannon has to work hard every day to make sure that they are all happy and she likes being able to see that happiness reflected in their eyes when she gets home from work each evening.}
\\
\\
\texttt{Alex is the main concern since he is the oldest. He is only ten years old. He is a smart kid who always does his homework and does it well but sometimes he can be a little disrespectful and bothersome in his younger sister's eyes. He always wants to hang out with Shannon, but she can't spend too much time with him because she has other responsibilities as well. She has a baby sister to take care of and she can't let Alex be at home alone while she goes out partying all night with her friends like some of the other girls at school do when they're really young and still high school students in grade ten or eleven like Shannon herself. She doesn't want to let her baby sister get hurt because she wasn't careful with her.}
\\
\\
\texttt{Shannon's baby sister is only four years old. She is a sweet little girl and a very cute child, but she can also be a handful sometimes because she doesn't know when to stop crying and making noise when they're at home. It seems like the little baby sisters never learn this skill; that is, the one of knowing when it's the right time to stop playing around and be serious and quiet. Because of that, Shannon knows that it's time for her to leave them at home one day while she goes out with Rebecca and some of their other friends from school. They are going out for a night on the town in town; there are drinks, music, dancing, and men all waiting for them – or at least that's what Shannon and Rebecca think. Shannon loves being out with her friends especially on Friday nights because it allows her to forget about how hard things have been for their broken family. She misses her mother even more at these times because these times make her realize just how much she relies on Eleanor for emotional support—and now Eleanor isn't there anymore to give them support anymore; she left way too soon.... Shannon loves her grandmother, but she can't help but think that sometimes Olivia and Alex could use a mother's love. Especially after the death of their own mom. Her grandmother always steps in and helps out with the kids whenever she can, but she doesn't have the patience or the knowledge to consistently care for them like Eleanor had.}
\\
\\
\texttt{However, Shannon knows that her mother wouldn't want her to give up on taking care of Olivia and Alex just because it's too difficult now that Eleanor is gone. She knows it will be hard, but Shannon wants to prove how responsible she is. Even though she is still a young teenager, she wants to prove how mature she is and how much of a woman as well as a girl. She has seen so many movies where people go out for some drinks to either celebrate something or forget something – like finding out they have been fired from their jobs or they have an incurable illness – and she wants to try this out for herself; to drink until her mind goes numb with pain and then maybe have a few men flirt with her while they're either drunk or not...}
\\
\\
\texttt{She also thinks it might be fun if one of these gentlemen was good-looking enough to take home. Even though she has no intention of going to bed with him, Shannon will relish the attention that she gets from this man. She would love to feel like a woman and have a man look at her like she was hot and he would love to bang her. This is the kind of attention that Shannon always longed for, especially when school was stressful and Alex would constantly pick on her or Olivia cried all day because Olivia just always cries...}
\\
\\
\texttt{So, Shannon decides that when she goes out with her friends tonight – even though it's only Thursday – there will be no flirting or drinking with men or doing anything sexual whatsoever. She was just going to drink until she felt numb and then go home feeling happy for a change. Yes! She would be able to leave loose ends together, spend time with her loved ones instead of thinking about work, have warm food in her stomach, maybe even get to sleep by midnight...}
\\
\\
\texttt{Shannon walks into the store and buys three bottles of wine. In addition to these three bottles, she also buys some chocolates because she realizes that chocolates would also make for excellent comfort food. She then walks to her car and drives to her house where she has to spend one hour before she can leave, waiting for Alex and Olivia to get home from school. She knows that once Alex comes home, he will try to make a big deal out of the fact that Shannon is going out with friends without him or Olivia. She also knows that once Olivia comes home, she will want to talk about how much she loves Shannon. Shannon really can't take care of them both at the same time and as much as she loves her sister and brother, there are certain limitations as to how much time Shannon can spend with them in one day because if there's anything for which Shannon doesn't have time at all, it's taking care of Alex and Olivia.}
\\
\\
\texttt{When a boy walks by her in the hallway and says hi, she thinks nothing of it because this boy is pretty cute even though he looks like he isn't someone who would ever be interested in someone like Shannon. At least not until he says hi...}
\\
\\
\texttt{“Hey” he says while smiling at her.}
\\
\\
\texttt{Shannon turns around because hearing him speak made her realize that whoever it was speaking was actually speaking to her. Her heart beat faster when she looked at him. He had light brown hair and green eyes. He also has a very boyish face and she finds him to be absolutely adorable. She smiles at him.}
\\
\\
\texttt{“Hey” she says back to him. “I'm Shannon...”}
\\
\\
\texttt{He smiles back and extends his right hand toward her. “I'm Luke Barnsby” he says while shaking her hand.}
\\
\\
\texttt{Shannon notices that he has a beautiful smile and when they shake hands, she feels butterflies in her stomach flapping around like crazy, which is something that she never feels when she meets guys at school but this time, it's different. She feels desirable, sexy and gorgeous just because he said hi to me, she thinks to herself as they walk off together down the hallway towards their separate classes.}
\\
\\
\texttt{As soon as Shannon arrives at home for lunch that afternoon, she calls her friends Stacy and Lisa on her cell phone to let them know that she's on her way over with wine, chocolate and cheddar cheese popcorn which they all like because it's salty which is what they usually crave when they drink wine together anyway because somehow alcohol makes them crave salty food. She says she'll be there in a few minutes and hangs up her cell phone then she goes to her bedroom and grabs the bag with the wine, chocolate and cheese popcorn out of her closet. She brings it in with her to the kitchen where she fills up a bowl with cheese popcorn then she pulls out three bottles of wine from the bag, sets them on the counter and then goes back in to get a glass for each girl. She knows that they all have fun "girl talk" sessions when they're drinking so maybe this will help them talk about their problems because they always say that they need to talk but they never do; Shannon is always happy to let them vent though.}
\\
\\
\texttt{As soon as Shannon arrives at Lisa's house, Stacy comes rushing out the door wearing a bubblegum pink t-shirt with matching shorts, jelly sandals and an Anna Sui head band. She's a petite Asian woman in her mid thirties with black hair and green eyes. Shannon can tell that she's ready for a good time because no one wears dressy sandals except for when they're going out on the town. It's summertime so everyone is wearing shorts and tank tops but Stacy looks like she's been expecting to go out all day; she always seems to know what to wear out of the house. Lisa is wearing a low-cut blouse with a red tank top underneath it, black stretch pants and some bejeweled flip flops. She has long black hair that comes down past her shoulders and brown eyes. They both give Shannon hugs then Lisa sets the three glasses of wine on the coffee table. They each take a deep breath as Lisa sits down. They all clink their glasses together and say "cheers."}
\\
\\
\texttt{They start to drink the wine and eat cheddar cheese popcorn at the same time. Stacy starts to ask Shannon about her baby brother then Lisa interrupts her, "What were you saying Stacy?"}
\\
\\
\texttt{Stacy rolls her eyes then says, "I was just saying how cute Alex is. He started talking today which I thought was a little weird but his speech therapist told me that he's going to be talking soon."}
\\
\\
\texttt{Shannon takes another sip of her wine and says, "Well what are you waiting for? Go pick him up right now so we can hear how cute he is."}
\\
\\
\texttt{"I'm not going to do that, you know how I hate leaving Olivia alone with my mother. Besides, it's only been one day; he won't talk very much unless he sees all of us together." The three women continue talking about Alex and Olivia for the next hour. Each of them take several drinks from their glasses of wine throughout this time and by the time they're half way done with the first bottle they have finished eating most of the cheddar cheese popcorn. Lisa leans back in her chair and takes a big pull from her wine glass. She sits it down and says, "I like how relaxed I'm starting to feel."}
\\
\\
\texttt{Stacy puts one hand on the table and one hand on Lisa's knee and says, "I'm far more relaxed than I've ever been. Wine makes me so happy!"}
\\
\\
\texttt{Shannon looks into her glass of wine then takes a sip. She puts it down and says, "This wine is good but doesn't it taste a little strange?" Then she takes another sip. "I swear it tastes like sugar."}
\\
\\
\texttt{Stacy's eyes get big and she takes another drink of her wine then makes a funny face. She shakes her head violently from side to side, "No way that's not the way we brought it over here". Stacy continues drinking. Then she holds the glass up to her nose and smells it before drinking more of the wine. "Hey that's definitely not what we brought over! The bottle is totally different!" Stacy gets up quickly and moves to open the front door saying, "I'm going home now!" Rebecca looks at Stacy with her mouth open and says, "Wait! How are we supposed to get home?"}
\\
\\
\texttt{Stacy says, "I don't even care anymore. I'm not going to drink this stuff." Then she leaves. Wine is Stacy's friend and she's upset because Stacy got angry and left.}
\\
\\
\texttt{Rebecca yells for Stacy to come back but she keeps walking. Rebecca's eyes are big and her mouth is open wide with shock. She says, "What the hell! That's disgusting! What are we supposed to do now?" Rebecca takes another sip from her glass then puts it down and rubs her hands over her face in frustration. She starts searching through the cupboards then comes out of the kitchen holding a bottle of rum and two shot glasses. She pours a shot of rum into each glass then hands one over to Shannon saying, "Drink up!"}
\\
\\
\texttt{Shannon puts both hands around the shot glass, holds it up and looks inside of it without drinking anything saying, "Is this stuff safe to drink?"}
\\
\\
\texttt{Rebecca slaps Shannon on the arm then pushes the shot glass towards her saying, "Don't be a baby! We need to relax!"  Shannon looks at Rebecca then shrugs her shoulders before bringing the glass to her lips and taking a small sip.}
\\
\\
\texttt{Rebecca downs her shot then pours herself another one. She offers the bottle to Shannon who shakes her head. Rebecca shrugs then puts the bottle back on the counter. She looks at Shannon and says, "You're such a good sister." Shannon feels a lump form in her throat and she looks down at the floor to hide the tears that are forming in her eyes.}
\\
\bottomrule
\caption{\ours{}'s generated story for Premise 4. The first half of the story is relatively high quality, being coherent and following the premise well, with Shannon struggling to take care of her siblings and later meeting a boy named Luke. The second half consists of a conversation between Shannon and her friends, discussing Shannon's siblings, concluding with one of her friends remarking how Shannon is ``such a good sister.'' While not exactly off topic, it seems as if the story has been cut off prematurely; in particular, Luke is never mentioned again.}
\label{tab:premise4_ours}
\end{longtable}
\end{small}

\begin{small}
\begin{longtable}{p{0.9\textwidth}}
\toprule
\textbf{\rolling{} Story for Premise 4}\\
\midrule
\texttt{Shannon had just turned sixteen when she received the news that her mom had died.}
\\
\\
\texttt{She couldn't believe it. She'd only spoken to her mom a few days earlier, and she seemed to be getting better. Shannon knew that her mom had a problem with drugs and alcohol, but she thought she was finally turning a new leaf. To hear that her mom had died of an overdose, it hurt Shannon more than she could explain.}
\\
\\
\texttt{For as long as Shannon could remember, her mom always disappointed her. She was never there for her, and Shannon knew that most of the time it was because of all the drugs and alcohol in her life. But over time Shannon had learned to live with it. She did what she could to provide for herself and for her siblings, even though sometimes it seemed like too much responsibility for someone who was still just a child herself. Her dad always said he loved them all dearly and would do anything if he could change things... but he couldn't change things, no matter how bad he wanted to do so. She was glad that after the death of her mother the state stepped in and took custody of her younger siblings. It was the best thing that could have happened for them, and she knew it. Her dad struggled to take care of all three of them, but he did his best.}
\\
\\
\texttt{Shannon was blessed with a lot of friends who were always there for her whenever she needed them. They all supported her whenever she felt lost or scared, but she realized that they couldn't completely take away the pain and loneliness inside of her. Sometimes Shannon would cry herself to sleep at night because she missed her mom so much, but no matter how much she hurt inside, Shannon knew that it would be selfish to ask one of her closest friends to comfort her at night when they had their own problems to deal with. She didn't want them to have to ever think about what it would be like if their mom died or if they were left alone... so Shannon never asked anyone for too much help and support. Instead, she tried hard to keep a brave face and pretended that everything was going okay all the time... even when it wasn't.}
\\
\\
\texttt{At school, she made friends easily despite being younger than most of the students in most of her classes. One of her teachers even stopped calling on her in class because he knew that she was a bright girl. He was impressed with how quickly she picked up on things and how much she seemed to like learning, despite the tough situation that she had been dealt by life. Shannon took all of her school work seriously, even though it didn't seem like school would be the most important thing in her life at the moment. But since her mom wasn't around anymore to take care of them, Shannon wanted to make sure that she was just as responsible as ever. She wanted to be an even better student than she already was so that one day when she got old enough, they could come stay with her while their dad went off and tried to find his own place.}
\\
\\
\texttt{It wasn't easy getting used to life without her mom. It seemed like everything that seemed special before now seemed cold and empty. She used to love spending time with her mom because they would always do fun things together—things they could never do while their dad was at home working all the time (and sometimes coming home drunk). But now there was no one around who could bring out those smiles in Shannon anymore. There was no one left to take her to a beautiful park or to go for a long walk. There was no one left to bake homemade cookies or teach her anything new about the world. Her life was all about school, work and her family now, and she knew that at least for the foreseeable future... that's how it would have to stay.}
\\
\\
\texttt{She felt so alone. She tried to distract herself from the grief by throwing herself into school and hanging out with her friends. But something was missing. She felt lonely even though she was surrounded by people who cared about her all the time... somehow, they were just never able to fill the void inside of her caused by losing the most important person in her life.}
\\
\\
\texttt{Everywhere Shannon looked, it seemed like she could see her mom's face. She always smiled whenever she thought of how much she loved and missed her mom, but at other moments it seemed like such a dark world without her in it—that she just wanted to curl up inside herself and cry until there were no tears left anymore. She knew that if things ever got too hard that there were other adults who could give them some support, but she wasn't ready to have anyone else in their lives yet... and neither were they. It would just be too confusing for the kids to have to deal with someone new in their lives right now. It was hard enough to deal with their father constantly drinking and coming home late at night after working all day...}
\\
\\
\texttt{Most of the time, it felt like she was on a slippery slope going down into a black hole. She could feel the world slipping away from her the longer things remained this way. Even though she tried her best not to cry at night, and even though she tried her best not to let it show how lost she felt... sometimes Shannon still couldn't contain all of her sadness. And when that happened it was okay, because as long as she didn't get too sad or depressed... there were always people who were willing to help her through it. They might not have known exactly how much pain and sadness Shannon really felt deep inside... but they knew that she needed help. They knew that they had to do something, otherwise they never would have confronted her dad in the first place.}
\\
\\
\texttt{***}
\\
\\
\texttt{"So tell me what's on your mind, honey," said Mr. Jones after taking a sip of his coffee in the kitchen while scanning his daughter's body with a detailed gaze. "Are things okay?" Shannon looked up at him. She had been biting her lip. But she was getting more comfortable discussing family matters with her father, which was a good thing. "Do you think we're going to have to live here forever?" she asked softly.}
\\
\\
\texttt{Her father looked confused for a moment as if he hadn't even considered this before. "Well... I suppose it depends on how things go... but I think we're taking it one day at a time right now."}
\\
\\
\texttt{"But do you think you'll ever find your own place? You know... maybe a new job?"}
\\
\\
\texttt{Mr. Jones chuckled. "Well, I already have a job, Shannon. I work with your uncle at the factory outside of town." He put his coffee cup down on the counter and leaned back against it, stretching his arms out wide as if trying to make her feel comfortable around him again after their last conversation about moving away from home—the one that left her feeling so upset and misunderstood by him that she stormed out of the room in anger and frustration. "Yeah, I know." She fidgeted with her fingers. "I just mean... do you think you'll get any other job? A better job? One that makes enough money to let us move out of here?"}
\\
\\
\texttt{"Well, I'm not sure what kind of job I'd be able to get if we did leave," he said thoughtfully. "And your uncle and aunt already have a lot of mouths to feed in their own house. And they aren't getting any younger either... I think it might be better to stay here for the time being."}
\\
\\
\texttt{"Yeah, I guess..." Shannon sighed and looked down at the floor. Mr. Jones was the first person she'd really talked to about her mother lately, and she didn't want him to feel like she was attacking him or trying to pick a fight with him all the time. But sometimes it was hard having a father who was also your uncle—and it made it difficult for Shannon to talk to him about some things without feeling like she had just stepped into an argument with herself at the same time. "Do you want to talk about it?" Mr. Jones asked softly after a pause. He put his hands on her shoulders and started rubbing them. "Is there something going on at home that's bothering you? I know you've been doing well in school and everything, but..."}
\\
\\
\texttt{"It's okay, Dad," she said, cutting him off. "I mean... it's not like I'm in trouble or anything." Shannon looked up at him with her large blue eyes sparkling. She didn't know how much she should really tell her father about the boys at school who were always bullying her now, or the friends that she lost because of it... but maybe she could get some advice from him after all this time without telling him about any of it. He used to be a big help for her back when he was working more as a teacher than an uncle. Back before he suddenly ended up moving in with Aunt Laura and Uncle Mark, things had been so much easier between the two of them... but since he had moved here things had gotten more complicated for them both....}
\\
\\
\texttt{"Well, that's good to hear," he said with a smile. "And if you ever need anything just let me know." "Thanks, Dad," she said. "Maybe I will."}
\\
\\
\texttt{***}
\\
\\
\texttt{Later on in the day Shannon was spending time working on homework with her father out in the living room. Even though he had retired not too long before he moved back to town, he still tried to make sure she didn't fall behind in school. And despite how busy the two of them were, and how many chores were constantly waiting for them both to complete, Mr. Jones still tried to help Shannon out whenever she needed help with her schoolwork... no matter what hour of the day it happened to be. "Well, what do you think?" asked Mr. Jones as he looked over one of Shannon's math problems. It was a word problem about gambling at a casino, and he was trying to show her how to work it out on a piece of paper so that she could better understand it for herself. She had been struggling with some problems like this lately in class, so Mr. Jones thought maybe showing her how to work them out would help her solve them more easily on her own after that. "Am I helping you at all?" "Yeah," replied Shannon with a sigh as she looked down at the problem on the paper before her with growing frustration. " I just don't get how to work this one problem. All of the others are easy..." "Well, work out this one problem and I'll be able to help you with the rest," said Mr. Jones reassuringly as he looked over what she'd already done in his attempt to show her how to solve it on her own.}
\\
\\
\texttt{Shannon tried for a few more minutes and then threw her pencil down on the table with a frustrated sigh. "I can't do it!" she complained loudly. "I just don't know how!"}
\\
\\
\texttt{"Hey, come on now," said Mr. Jones as he put his arm around Shannon's shoulder and gave her a comforting squeeze. "I think you're getting it! It's just going to take time is all."}
\\
\\
\texttt{"Yeah, but I'm running out of time..." Shannon hung her head down low and scrunched up her face in frustration. She hated feeling helpless like this... and the fact that she was running out of time only made it even worse because there was so much she needed to catch up on at school that she didn't have time left over for anything else anymore! "The only way to get it is to work harder and not give up," said Mr. Jones as he looked down at Shannon with his blue eyes sparkling. "Believe in yourself, and you can do anything."}
\\
\\
\texttt{"Thanks, Dad," said Shannon with a smile as she gave her father a hug. He was often telling her these kinds of things when she was feeling down about something... but this time it actually seemed like he really believed it for once. It made her feel good inside just to hear him say that. Only moments the pair were startled by a knock on the door and Aunt Laura calling them from the kitchen. "Mr. Jones! Mrs. Harvey's on the phone again... she wants to know if you can come over again already." "Oh..." said Mr. Jones as he let out a great sigh of disappointment before standing up from his seat next to Shannon and letting out a long sigh at being interrupted yet again by his second job at Mrs. Harvey's house across town. A few minutes later he was back home talking on the phone with another difficult customer that couldn't understand what they were supposed to be doing, or how they were supposed to do it.... *** Just as Shannon was finally starting to get caught up on her schoolwork, her mother appeared at the front door crying and begging for help. Shannon hurried over to see if she could help her out or calm her down or something. "Mom, what's wrong?" asked Shannon worriedly as she patted her mother's shoulder while she dabbed at her eyes with a tissue. She quickly explained that it had been a rough day at school... and in fact the whole week had been rough because Shannon was finally starting to catch up with all of her schoolwork now that she wasn't keeping up with chores anymore. As a result Shannon had been so busy catching up on homework every night that she hadn't even had time to talk on the phone with friends... which was frustrating them all because they didn't know what was going on with her either! "Oh, I'm so sorry mom," said Shannon as she felt like crying herself after hearing all about the trouble that this homework-related situation at school had caused for her.  "I'm sorry," she repeated with a sniffle. "I didn't mean for any of this to happen." "I know, sweetie," said Shannon's mom as she gave her daughter one last pat on the shoulder before getting up and leaving the room again.}
\\
\bottomrule
\caption{\rolling{}'s generated story for Premise 4. The story begins with Shannon's mother's death, but the remainder seems wholly irrelevant to the premise, instead focusing primarily on the relationship between Shannon and her father. In the end, Shannon's mother even seems to be alive again.}

\label{tab:premise4_rolling}
\end{longtable}
\end{small}

\begin{small}
\begin{longtable}{p{0.9\textwidth}}
\toprule
\textbf{\rollingfinetune{}{} Story for Premise 4}\\
\midrule
\texttt{A Winding Road}
\\
\\
\texttt{Shannon sat at home, eating a bowl of cereal. Her phone was on the counter next to her, and she glanced over at it from time to time. It was a cheap android phone, one she had gotten for free with her new service plan. The offer had been too good to pass up, even if it did feel like a bad sign that she'd gotten such a good deal out of the deal. She knew her mom wouldn't mind either way, especially if the plan got Shannon out of the house more often.}
\\
\\
\texttt{"Anything good coming up tonight?" She asked herself as she reached for a spoonful of cereal. She paused before actually taking the bite, considering the question again in her head.}
\\
\\
\texttt{"Well... I guess we could watch that movie my mom always liked," She said before taking a bite of cereal. As she chewed, she looked down at her phone again and took another look at what was scheduled for tonight on TV. There were plenty of options to choose from: A drama on FX; A sitcom on ABC; A cop show on NBC; Movies on HBO; Several reruns on things like PBS and TNT...}
\\
\\
\texttt{"I know that movie... Mom said it was cute." She said as she moved her spoon around the bowl, stirring the last tablespoon of cereal from the bottom.}
\\
\\
\texttt{"Right, I remember." She said as she took another bite. After swallowing, she continued.}
\\
\\
\texttt{"I'm pretty sure that channel's broadcasting it right now." She said before glancing at her phone again. The clock on the phone read 6:50; The movie was supposed to start at 7pm, but Shannon liked to arrive early when going to a show at a theater or a museum so that she could find a good seat without having to rush around and jostle with other people. She glanced up at the TV in her living room and figured she had plenty of time to take care of things before then.}
\\
\\
\texttt{"I really should call my boyfriend," Shannon said out loud as she tapped on her screen; She was trying to decide if there was anyone worth calling for dinner or if it would be better for dinner tonight if she just went out with her son, since he didn't have school tomorrow. Finally deciding that it might be best if they both went out for dinner tonight, Shannon called up Brian's number and hit send before setting the phone down next to her bowl of cereal.}
\\
\\
\texttt{*** ***}
\\
\\
\texttt{Shannon walked into the theater, her son in tow. A movie was playing in the theater tonight; A popular movie that she had seen a couple of times already with Brian. She'd told him she liked it a lot, though he really seemed to like that movie better than she did. She supposed it made sense, given his taste in movies and all that he had seen. He'd seen a wide variety of movies from her collection of them; Films from everything from the 50's 'Godzilla' to nearly every comedy that had been released in the last 10 years or so; Even some things from as far back as the 1960's: '2001: A Space Odyssey' and 'The Dark Knight'. He really seemed to like older films, even more than he did with newer ones.}
\\
\\
\texttt{"We're not seeing anything on TV?" He asked as they walked into the lobby of the movie theater. Shannon shrugged at him before looking around for someone else.}
\\
\\
\texttt{"I thought you said we were going early." Brian said as he pulled out his phone and glanced at it.}
\\
\\
\texttt{"Oh... I thought we were!" Shannon said before glancing at her own phone. It wasn't even seven yet; Not that she had meant to arrive early. She'd only thought that the movie would start at 7:00 or so, not that it was scheduled to start at 6:45.}
\\
\\
\texttt{"That's fine, I'll just go check the times." Brian said before skating off away from her. Shannon sighed as she leaned against a nearby wall, watching him skate off with his phone in hand. She glanced around and nodded at a couple of other kids passing by with their families before glancing back down at her own phone again.}
\\
\\
\texttt{The screen went black for a moment as she looked up, then a moment later it came back to life again with Brian's number in the contacts list and below it read 'Mom'. Shannon smiled happily before leaning over and tapping on his name; It was a feature her son had requested to be added to her phone so that he could have easy access to things like calling home and texting his mother when he needed something important. Ever since he'd gotten this feature, he'd been using it quite often: At least two or three dozen times every day for one reason or another.}
\\
\\
\texttt{"Hey mom!" Brian said loudly as he answered back on the phone and quickly skated over towards her again. "I'm over by the doors now. Are you coming to get me?" He asked before glancing at the time on his phone.}
\\
\\
\texttt{"Uhh... Yes, I am..." She said quietly as she looked around for him; Then she saw him; He was wearing a bright green t-shirt and dark blue jeans and he was rushing towards her with a huge smile on his face. "Brian! Could you slow down?" She said in a quiet tone as she put her phone to her side while Brian came up to her, still smiling.}
\\
\\
\texttt{"I'm coming, I'm coming. It's fine." He said before gently taking hold of his mother's hand and skating off with her towards the theater doors.}
\\
\\
\texttt{"Alright, this way." Shannon said as they got closer to the entrance to the theater. Brian nodded before giving an enthusiastic wave to one of the workers near the entrance who stood by the door not saying anything just yet because it wasn't time for that yet. Brian held the door open for his mother before following her in while still walking on his hands and they made their way inside the theater.}
\\
\\
\texttt{"Wow, this place is really nice." Brian said as he looked around at everything. The lobby was huge; Much bigger than most other places in the mall and it even had a large carpet at the bottom of it that had been nicely decorated with paintings and designs to be more appealing to the people who came here. There were even several chairs for people to sit in if they wanted to wait here instead of continue on their way into the theater itself.}
\\
\\
\texttt{"I know, I've been coming here before you were ever even born!" Shannon said with a chuckle as she sat down next to Brian on one of several benches that had been carefully placed here for people to sit in while they waited. She glanced over at her son and watched him stare around at all of the decorations that had been placed neatly all around them. His face was bright green from his shirt he wore now, but other than that he just looked like any other boy around his age: A little taller than others, maybe a little fatter perhaps; But he wasn't really all that much different than anyone else when it came down to it. "Look, they've even got the Pokémon games out!" Brian said as he pointed at a table across the way. On the table was a bunch of Pokémon cards, some toys, and some little figures that looked just like the ones you'd find on any of the TV shows or in any of the video games that had been coming out in recent time. It wasn't much, but it would be something to keep him occupied while they waited; And if his mother let him have some money to spend here with her, she'd probably let him spend it on one or two things from this table.}
\\
\\
\texttt{Shannon nodded before leaning over and gently putting her phone into her purse which sat beside her on the bench. She watched as Brian carefully skated off towards the card game table and she smiled lightly as she saw him grab one of them before holding it up and giving it a once-over for a moment. She leaned back in her seat and closed her eyes for a moment as she relaxed; Brian was going to be fine getting his Pokémon figure once he finished looking through the cards. She smiled to herself as she opened her eyes again before sitting back up when she felt someone gently tapping her on the shoulder.}
\\
\\
\texttt{She looked to see who it was and smiled at him when she saw he was one of the employees of this place: He was a young fellow who looked like he couldn't have been more than twenty or so years old. He was tall and skinny, with a little bit of facial hair that made him look a little older than he really was. It had to be a recent addition because it wasn't very thick or long, but it still did its job well enough so that people would think he'd just recently gotten out of high school or college. He wore a shirt that had "Greg" written on it in big letters; It looked like he worked here, but Shannon couldn't be sure.}
\\
\\
\texttt{"Welcome lady. How may I help you?" Greg said before giving a casual wave as he looked at Shannon. She smiled back at him for a moment before answering.}
\\
\\
\texttt{"Oh, I'm just here with my son. We're just waiting for the movie to start." Shannon responded; She was surprised that this guy knew who she was because most of the staff here didn't tend to know her by her name and they most certainly didn't know Brian's name either. If they would have, the people running these places would probably have been more understanding to people like her rather than making it seem like they were just sitting on their asses all day doing nothing while their kids got themselves into trouble or whatever else it is that parents tend to do nowadays. She'd been treated like that before and she wasn't happy about it either.}
\\
\\
\texttt{Greg nodded his head in agreement before giving another wave.}
\\
\\
\texttt{"Yep, I saw you come in here with him." He said as he sat down on the bench next to Shannon, leaning forward to rest his arms on his knees and fold his hands together. "So how long you two been here?" He asked when he sat down next to her. "I'm guessing you live nearby or you wouldn't have made it here on time. Heh, then again it's always hard to tell what people can do when they want to get somewhere fast."}
\\
\\
\texttt{Shannon nodded her head in agreement.}
\\
\\
\texttt{"Yep, we've lived here for about a year now and we make it here just fine every time. The only problem is when Brian's involved because then I have to take care of him and that just makes things a little slower."}
\\
\\
\texttt{"I see. Well, I'm sure he's gonna be just fine being here with you while the movie's playing. We've got some people who work around here who help watch over kids like him while they're in the place." Greg said as he looked over at Brian who was still looking through the Pokémon cards; He took a closer look at one of them before tossing it back into the pile and grabbing another one to look through as well. "So what movie is this anyway? Looks like one of those Pixar movies that come out every year." Shannon nodded her head again, while wrapping a few of her fingers around the straps to his purse.}
\\
\\
\texttt{"It IS one of those movies. It's called Finding Dory." She responded, glancing over at her son again before looking back at the guy, who had an eyebrow arched up in interest. "And it looks like it's about a whale named Dory who gets lost and ends up being found by another whale who helps her find her way home."}
\\
\\
\texttt{"Oh yeah? Well I've seen enough movies where people end up being lost to know that she's gonna find some way to get home in the end." Greg said before glancing over to Brian again, watching as he sat down on the floor and began playing with the cards he had gotten from the table before looking back at Shannon. "So what brings you two here? I mean I know Brian came here for Pokémon but what about you?" "Yeah, she's just here because I asked her to stay with me while we watched the movie." Brian said as he looked up from his cards and gave a small smile. He then continued with," She's not the best at watching Pokémon because she doesn't like any Pokémon other than Arceus and Pikachu. On top of that, she doesn't even know how to play any of them so it got boring for me. I'd rather spend my time having fun by myself rather than having her try and play it with me."}
\\
\\
\texttt{Shannon glared at Brian for a moment before turning back to Greg; She didn't want to tell him that he was actually right about their situation because doing so would mean that she was acknowledging that he was right and that wasn't something anyone should do, especially if they were talking to someone who looked like an employee of a place like this.}
\\
\\
\texttt{"Yes, well I can see why you might feel that way; I'm not much help when it comes to playing Pokémon either." Shannon responded as she gave Greg another wink before flashing another grin. "But you'll have fun today since you're in such a great environment."}
\\
\bottomrule
\caption{\rollingfinetune{}'s generated story for Premise 4. This story is also completely unfaithful to the premise. Shannon appears to have become the mother herself; Brian is at first her boyfriend and later her son. The bulk of the story describes Shannon taking Brian for an outing involving movie(s) and Pokémon games.}

\label{tab:premise4_rollingft}
\end{longtable}
\end{small}

\newpage
\subsection{Examples for Premise 5}

\begin{table}[!htbp]
\small
\begin{tabularx}{\linewidth}{X}
\toprule
\textbf{Premise 5}\\
\midrule
\texttt{After years of being content with only a small inheritance, Sarah's life is turned upside down when she discovers she is actually a millionaire. She must learn to navigate her new life of luxury while also dealing with the guilt she feels over her new-found fortune.}\\
\bottomrule
\caption{Premise for the fifth group of example generated stories. A relatively mundane premise.}
\label{tab:premise5}
\end{tabularx}
\end{table}

\begin{table}[!htbp]
\small
\begin{tabularx}{\linewidth}{X}
\toprule
\textbf{Initial Setup and Outline for \ours{}, Premise 5}\\
\midrule
\texttt{Premise: After years of being content with only a small inheritance, Sarah's life is turned upside down when she discovers she is actually a millionaire. She must learn to navigate her new life of luxury while also dealing with the guilt she feels over her new-found fortune.}
\\
\\
\texttt{Setting: The story is set in a small town in the US.}
\\
\\
\texttt{Characters:}
\\
\texttt{1. Sarah Jensen is a young woman in her early twenties. She has shoulder-length blonde hair and blue eyes. She is of average height and build.}
\\
\texttt{2. Jessica Parker is Sarah's best friend. She is a few years older than Sarah and is married with two children. She is a stay-at-home mom.}
\\
\texttt{3. William Jensen is Sarah's father. He is in his early fifties and is a successful businessman.}
\\
\\
\texttt{Outline:}
\\
\texttt{1. Sarah learns that she is a millionaire.}
\\
\texttt{2. Sarah must learn to navigate her new life of luxury.}
\\
\texttt{3. Sarah deals with the guilt she feels over her new-found fortune.}
\\
\bottomrule
\caption{\ours{}'s initial setup and outline for Premise 5. The characters and outline make perfect sense for the given premise.}
\label{tab:premise5_ourssetup}
\end{tabularx}
\end{table}

\begin{small}
\begin{longtable}{p{0.9\textwidth}}
\toprule
\textbf{\ours{} Story for Premise 5}\\
\midrule
\texttt{Sarah Jensen never knew she had inherited a fortune until her mother died.}
\\
\\
\texttt{In fact, she had never met her father, and she hadn't even known that he was alive until the reading of his will. The only thing Sarah did know was that he had died long ago and that her mother had been left a sizeable inheritance. William Jensen wasn't around to provide for his wife, but he did make sure she had enough money to see herself through life. Sarah's mother never spent any of the money because she wasn't sure what it was for - all that mattered to her was how much it would mean to her daughter when she inherited it once it was gone.}
\\
\\
\texttt{Although Sarah's mother lived in New York City, the two families were not close. The visit that changed everything happened when Sarah's mother died and William requested a meeting to discuss his daughter. He only knew of her existence because he had found out shortly after they were both born that they were siblings - they shared the same father, but not the same mother. It left him with a few questions... So a meeting was arranged, and Sarah learned that she was a millionaire. She controlled her father's estate, and if she took the proposal on the table, then she would have ownership of a large financial empire.}
\\
\\
\texttt{Sarah didn't claim to be an expert in finances, but she knew what owning a company meant. She had also been told that her father's company was slowly being taken over by some other gentleman, who was trying to take advantage of William's death. The police were investigating him for fraud, which was why he was looking for interested investors or parties who might be interested in acquiring control of the business through less than legal means. From what she could gather from her father's papers and his private investigator, it seemed as though this man wasn't interested in acquiring William's company so much as he wanted to make sure William wouldn't get any more money or power after his passing. He would do all he could to weaken the company, to purposefully lose it millions so that someone else could swoop in and buy it out at a reduced price. It was hard for Sarah to understand. The man was obviously a crook, and she had seen plenty of that sort in her previous life. She had a few friends - some members of the local gangs - and she knew the sort of people who would do this sort of thing. She couldn't believe that someone would actually try to pull it off.}
\\
\\
\texttt{William's private investigator had put together exhaustive notes, including bank records and all sorts of information about what her father did with his money. Most of his money went into other businesses, but he still had plenty left over to manage himself and invest in other things as well. Which is why he wanted Sarah to claim control over the estate... He wanted her to take over and claim control like he always knew she could, instead of allowing some dirty old man in his mid-fifties who wasn't even related by blood to take it away from her.}
\\
\\
\texttt{He didn't want Sarah getting any more wealth than she already had - which was considerably more than most people could ever dream of getting with one account, let alone multiple accounts around the world that exceeded millions upon millions each time they were checked - but he wanted her to do something worthwhile with it. To be responsible with it and make wise investments.}
\\
\\
\texttt{Sarah was lucky that her mother had married some wealthy man; otherwise, she'd have no money at all. And she would never have had a chance like this - an opportunity to be financially independent, to live on her own, to support herself and everything else. And now that she knew about the inheritance and about what was going on - about the marriage fraud, about the corruption that was happening in her father's firm - she knew that she couldn't trust the firm anymore. It wasn't just some random firm; it was her father's business. She couldn't let things go on as they were, or let them stay in control of her money any longer.}
\\
\\
\texttt{She didn't need to do anything drastic... yet... but she knew when she got home with this information - when she got home and told Jessica Parker everything that was going on - then they would figure out what they wanted to do together. She needed to talk it over with her best friend before making any rash decisions...}
\\
\\
\texttt{\textasciitilde \textasciitilde \textasciitilde \textasciitilde}
\\
\\
\texttt{Sarah arrived at the airport rather early in the morning. She went to the ticket counter, but she had no need to buy a ticket because she owned most of the company. They had no issues with letting her get on a flight before it was scheduled, and so that is exactly what she did.}
\\
\\
\texttt{She didn't want to wait around for hours until it was time for her flight, and so she purchased some magazines and snacks from one of the vendors inside and then settled into a seat near the gate area inside of the terminal. The sound of jets outside going off left and right reminded her of home, of business class flights on which she'd been seated many times before, but never in economy seats or as an economy passenger. Perhaps this was just how normal people flew; perhaps this was how they were used to flying rather than being upgraded to business class seats or having their own private jets. Now that Sarah was financially independent - now that she didn't need her father's money anymore - maybe she would fly more often instead of taking private jets all the time. Maybe she would choose to fly with everyone else. She'd never thought of that before, but perhaps that was something she would want to do now that she was no longer in the position where she needed her father to bankroll her.}
\\
\\
\texttt{When Sarah's flight was called, she boarded the plane and took a window seat. As usual, there were plenty of empty seats on the plane, and it wasn't until they had taxied down the runway and were off in the air that another passenger sat down next to her. Things seemed so different now than they had been before; things had changed so drastically in a matter of days. The person sitting next to her - someone who knew nothing about Sarah's new wealth or about her past as a failing businessperson - smiled at her nervously as Sarah returned his greeting with a polite smile of her own. She felt bad for him, because he clearly didn't know what he was getting himself into by sitting next to this young woman. The flight was uneventful, and soon they were landing at their destination. As Sarah walked out of the airport, she thought about how she always associated airports with stressful travel and sleepless nights on hard benches in the waiting areas, but now that was about to change. Now that she was rich, things were going to be different for her. She could afford a hotel suite and a lavish meal if she wanted it. Or maybe...maybe she wouldn't even need to spend any money at all! An executive suite should be arriving for her soon; why bother getting a hotel room when there was no need?}
\\
\\
\texttt{She touched the keys in her pocket with one hand as the elevator ride carried her upwards. She briefly considered calling Jessica but decided against it; discussing these details could wait until tomorrow morning, now that everything had come true. This last detail needed to be handled before everything else moved forward, but Sarah's thoughts were still caught up in nostalgia over what this trip had meant to her before she knew anything - before it had even begun.}
\\
\\
\texttt{As soon as the elevator doors opened and Sarah stepped out into her room on the fourteenth floor, she made a beeline straight for the bar. She was anxious to try out her new body, but there was one thing she hadn't been able to do for a very long time now.}
\\
\\
\texttt{The bartender came over and stood by her side. "What can I get you?" he asked with a smile. Sarah smiled back and told him what she wanted, trying to make sure that he didn't hear the shakiness in her voice. The bartender nodded and walked away, leaving Sarah alone with her thoughts as she waited. It seemed like they had barely left the airport before everything had changed. It seemed like so much time should have passed between then and now, but it hadn't at all. All of this could still make sense if Sarah could just take the opportunity to think about it now.}
\\
\\
\texttt{The bartender came back with her drink in hand, and the moment that their eyes met again, Sarah knew that she would be alright from this point on. She looked at him and smiled warmly as he set down the drink in front of her. Her heart beat so quickly that she could almost feel it humming in her ribcage. Her hands hovered over the drink as she tried to think of what to say.}
\\
\\
\texttt{"Thank you so much," Sarah said, putting on a winning smile. "This looks great." The bartender nodded and brushed a bit of hair out of his face with the back of his hand.}
\\
\\
\texttt{"It's my pleasure," he said, smiling at her from behind the bar before walking away again. As soon as he was gone, Sarah turned to face the other direction so that he wouldn't be able to see her crying again. She wasn't sure why she felt like crying when there were so many other things that she should be thinking about, but for now it seemed like all she wanted to do was cry. The words "Forgive me, father, for I have sinned" came unbidden into her mind as she reached out for her drink and took a large sip straight from the glass. It burned its way down into Sarah's throat, causing her to cough slightly before looking at her watch and noting the time: 10:59 p.m. She wasn't planning on sleeping much tonight.}
\\
\\
\texttt{***** The evening passed slowly as Sarah moved from one room to the next and then back again. Her entire body ached with exhaustion, but her mind was racing a million miles an hour. She kept thinking of what she should do with all of the money that she had now won. After all, this wasn't just any old lottery ticket that she had purchased, it was the winning ticket! She had actually won!}
\\
\\
\texttt{There were so many possibilities for what Sarah could do with her fortune that her head felt like it was going to explode. Should she buy a nice house and furnish it with the best furniture? Or should she spend it on something to show people what a success she had become? Her mind turned to her friends and family and how they would react when they found out that Sarah Jensen was now wealthy beyond their wildest dreams. There was no doubt in her mind now that everyone would have something to say about it... But who knew what it would be.}
\\
\\
\texttt{What if they looked down on her? What if other people were already talking about how stupid she had been for buying those tickets in the first place? How could someone who worked two jobs and still struggled just to get by afford to buy hundreds of dollars worth of lottery tickets? How could anyone who couldn't even afford the things that were needed for a baby's crib spend all that money on tickets that might never pay off? She had thought about these things before, and she had been sure that she would give some of the money away to deserving charities. But now that she had won, this felt more like her money than it did from a bunch of strangers.}
\\
\\
\texttt{Sarah owed her father so much already. He had single-handedly funded her college education with his small business income, and then he had given her everything else that she needed to get by after graduating. She wanted to be able to give him back some of what he had given to her... But the guilt was weighing on her now. Had she really stolen someone else's chance at winning? Had she really stolen their dreams like they must have stolen hers? And in the end, if it wasn't for them, would someone like Sarah have ever won anything in the first place?}
\\
\\
\texttt{Sarah tried not to think about it too much. There was no point in dwelling on things that couldn't be changed anyways. Instead, Sarah decided to focus on the positive aspects of the situation... Because there were plenty of those too! She could help people now when they were in need. She could give back to her community, or even start her own charity from scratch.}
\\
\\
\texttt{Sarah got to thinking about what she might do with the money and how she could spend it. She had always been a big proponent of animal welfare, especially since her mother had passed away. Sarah was still so young then, and she remembered how sad her mom had been when they had to put down their dog when it was virtually on its last breath. But the new money would fix that problem! She could pay for a veterinarian school somewhere and get her parents back a new dog. And what would they name it? She thought about this for a while, and decided upon... "Joy".}
\\
\\
\texttt{Now that Sarah's mind was off of debt and onto helping others, things were much easier for her to deal with. She felt much better about herself in this state of mind than she did anything else. It felt great to know that she would be able to help out animals like the one that she lost so long ago... In fact, it made everything feel worth it!}
\\
\\
\texttt{But then something else occurred to Sarah... She couldn't use this money to buy something like a nice house in a nice neighborhood if she was going to waste it on charity donations!  She would be living in the same house, in the same neighborhood, with the same people for the rest of her life...}
\\
\\
\texttt{And that's when it hit her.}
\\
\\
\texttt{The guilt.}
\\
\\

\bottomrule
\caption{\ours{}'s generated story for Premise 5, corresponding to the abridged main text example in Table \ref{tab:abridged_example}. This story is overall relatively high-quality, following a coherent overarching plot which clearly matches the premise. However, the story is certainly not perfect, and there are at least a couple of inconsistent details. At the beginning of the story Sarah's father William seems to be dead and then alive again. Sarah's money initially comes from an inheritance, but later in the story is described as coming from a lottery.}

\label{tab:premise5_ours}
\end{longtable}
\end{small}

\begin{small}
\begin{longtable}{p{0.9\textwidth}}
\toprule
\textbf{\rolling{} Story for Premise 5}\\
\midrule
\texttt{I was not a happy camper.}
\\
\\
\texttt{I had just turned thirty years old, and my life was going nowhere. I had no friends, no family, and no prospects for the future. In fact, as far as I could tell, my future held nothing but more of the same—stuck in the dead-end job that I hated and living in a run-down apartment with only my cat for company. And if that wasn't bad enough, I was starting to get wind that one of my clients was going to fire me if things didn't turn around soon—and the outlook wasn't good.}
\\
\\
\texttt{What I wouldn't give to have been born into a life of luxury… To have been born into one of those families that had everything handed to them on a silver platter. As long as I could remember, this had been my regular daydream. It was something that never would have happened in real life. My parents couldn't even afford to give me an allowance or buy me an ice cream cone without having to scrimp and save for months at a time. It wasn't like they didn't love me… They just had seven mouths to feed and not much money coming in. My dad was a plumber and my mom was a waitress. They worked their butts off, but they barely got by.}
\\
\\
\texttt{The daydream that I just couldn't seem to shake was of being born into the lap of luxury, the kind of life that little girls dream about having. It started when I was a little girl, and it only grew as I got older. Now it seemed like it was all I ever thought about anymore.}
\\
\\
\texttt{I wasn't sure how to go about finding my Prince Charming or getting adopted by some rich family, but I did know one thing for sure: as soon as I realized that my life had taken a hit and that things weren't looking too good for me, the first thing on my mind was how to get myself out of it—and fast. That was probably the only thing that I really got right in my life.}
\\
\\
\texttt{I was vain, spoiled, and selfish. I knew that all too well. I didn't need anyone to tell me or to point it out to me when they saw it in my actions. It was especially clear to me when I talked about my daydreams with other girls my age who were working hard just to make ends meet. It was shameful, but it was true, and there was nothing I could do about it…}
\\
\\
\texttt{Or so I thought until a little over a year ago when things started to change for me for the better. As much as I didn't want to admit that something good could happen in my life, especially after everything went downhill so quickly and so completely, there was no denying what had happened—something good had happened! On the heels of losing my job and barely making enough money for rent each month came an unexpected gift… one that came out of nowhere like a bolt of lightning on a clear spring day.}
\\
\\
\texttt{I never expected it—in fact, it never even crossed my mind once while I dreamed those dreams of being born into a life of luxury—but there it was. It was just sitting there on my coffee table when I came home from work. I had been working so hard for so long to make ends meet that I didn't know if anything could ever surprise me again. That's probably why the first time it happened, I thought that it must be some mistake—that someone must have left a bag at my place by mistake. It was a small brown paper bag with one of those drawstrings at the top, just like you'd get if you bought something from a store and it was wrapped up for you.}
\\
\\
\texttt{When I saw what was in the bag, my jaw dropped, and my heart skipped a beat. Inside the bag sat over a million dollars in cash! At least that's what all of the bills looked like—five-, ten-, and twenty-dollar bills all stacked neatly inside of one another. There was no note of any kind, no information about where they came from or how they got there—just a bag with over a million dollars in cash sitting right there on my coffee table like it belonged there…}
\\
\\
\texttt{I knew right away that it was dirty money and that whoever left it on my front doorstep had done so by mistake. It was meant for someone else—most likely for a drug dealer or some other criminal. That's what I thought at first anyway. But I didn't leave it sitting there any longer to just sit and wait for the cops to find out about it and come knocking on my door. It was fate, pure and simple, and I knew that I had to do something with it…}
\\
\\
\texttt{As soon as it happened and I realized what was in the bag, my first thought was to run right down to the bank with the cash, but then I figured that if whoever left the money on my front doorstep wasn't going to claim it, then they must have had a reason. That reason most likely had something to do with not wanting anyone knowing that they were now holding a million dollars in cash while they weren't doing anything illegal or dishonest themselves. If that was the case, then taking it right down to the bank was probably not going to work out so well for me…}
\\
\\
\texttt{My second thought after finding out about my incredibly huge amount of money was not much better than my first one though: I wanted nothing more than to take the dirty money and run straight out of town as fast as my feet could carry me! It was an idea that I used to think about all the time when I was a little girl and nobody in my life would give me anything but mean words.}
\\
\\
\texttt{It was a fantasy that I used to dream about every night as I lay in bed when my mother would shriek at me for not cleaning up my room or for not doing whatever she told me to do for her… That kind of money would certainly have made things a whole lot easier for me, and it would have allowed me to live out one of my favorite daydreams—leaving home and never coming back! It was a crazy idea, and it's just as well that I didn't leave with that money, though.}
\\
\\
\texttt{The whole thing went from bad to worse so quickly that there wasn't a whole lot of time—if any at all—for me to question what had happened. Right after I found that bag sitting on my kitchen table, the first thing that I did was call the police! Yeah, right! In hopes of getting them over there pronto so they could take care of it. As soon as they picked up the phone, though, all hell broke loose! "Sir, where's the money?" the police officer asked me.}
\\
\\
\texttt{"It's here," I said. "You see, someone left it on my front porch…"}
\\
\\
\texttt{"Are you fucking kidding me?" he replied as if I was lying. "Who left it on your porch?"}
\\
\\
\texttt{"Just what kind of question is that?" I asked. "I don't know who did! You see, someone dropped off this bag with a million dollars in cash, but they never came back for it! What am I supposed to do with it?!"}
\\
\\
\texttt{"You're not supposed to do anything with it!" he barked back at me. "Listen lady, this isn't your money, and we aren't going to pretend that it is just because you suddenly had a lottery win or something like that… This is dirty money and there's no way anyone who wants to keep their freedom safe is going to go around asking anyone where they found it!"}
\\
\\
\texttt{He hung up almost immediately after he had said the word "freedom," and that was when my whole world spun out of control. I instantly started thinking the worst, imagining that the telephone call had set me up for a robbery or some other horrible fate. What would anyone want with me? I was just a housewife and my husband was off in the military… I didn't have much at all, but that's why I had kept the money as close to me as possible. I didn't want to lose it!}
\\
\\
\texttt{I ended up keeping it hidden away for about two days until I got fed up with asking everyone if they could figure out where the money had come from. None of them could help me though—not my friends, not my kids, and certainly not even my husband—and that's when I decided to take matters into my own hands! If no one else was willing to help out and someone wanted this filthy money then they were going to have to find it themselves!}
\\
\\
\texttt{I knew right away exactly what I was going to do with it too. The first thing that came to mind after almost 48 hours of hiding under everything in my life was to find whoever dropped off the million dollars on my doorstep and hand it right over! As soon as the idea came into my head though, I ran straight into another problem—what if someone else found the money first? That would have been a tragedy on top of a tragedy! I knew that, and I also knew that I wanted to find the person who had given it to me. I pictured them in my mind as an old man with many years on him and wrinkles across his face. He was sitting on the garden chair staring out into the distance, a sympathetic look in his eyes, and holding onto a little dog for comfort. He had been planning on giving me the money for quite some time he said…}
\\
\\
\texttt{But why? What was he thinking? Did he know me? Did he want to help us out or did he just need someone who wouldn't ask questions or turn him in to the authorities? I couldn't figure this whole thing out because no matter what my mind came up with nothing made perfect sense. It seemed like it would be much easier if they had just left a note behind telling me where they wanted their money to go, but then again maybe they weren't thinking very clearly at all! I began to think that maybe the man didn't even know where he had dropped it off. If he did, I wouldn't have held onto it for a single second longer than I already had! But something told me that leaving the cash on my front step was a very deliberate and smart thing to do. It was almost as if I was meant to find it and hold onto it for him until he could be found. Of course, this didn't stop me from being scared out of my mind—not at all—and it's still something that I don't fully understand to this day! Maybe there's not even an answer behind the entire situation after all…}
\\
\\
\texttt{The Story Behind The Cash: Part Two}
\\
\\
\texttt{I'm going to tell you a story about someone who never existed and how they ended up on the other side of the world with a pile of cash in their hands. I'll also explain why they did what they did and if they really knew that what was happened would take place… Now, as you might expect, this isn't going to be your typical story because our tale begins when two people fell in love with each other much like everyone else in the world. They worked hard, they raised a family, and they spent every dime that they could on their children. They also did a lot of good things for the community and the people around them and ended up being loved by everyone in their lives…}
\\
\\
\texttt{But that all changed when one of them was diagnosed with cancer. It started out small at first, but the disease eventually took over their entire body and made it impossible to even leave the house. This was a very sad thing to witness because he had done so many wonderful things in his life and he didn't deserve to be taken from the ones that he loved so quickly. The family used every ounce of money that they could scrape together to keep him alive as long as possible, but in the end it wasn't enough for him to make it back home again. He left this world before his dream home was built, before his last child could finish college, and before he could even get his wife an anniversary gift!}
\\
\\
\texttt{She was heartbroken after his death and nothing seemed to cheer her up at all. It was a very hard time for her and she ended up losing her job and getting into a whole heap of trouble with the law. This is why I'm telling you this story because when she got out of prison her life took a turn for the better. Instead of moping around, looking for someone to take care of her and make it all better, she went out and did it herself! She found the support that she needed to finish college, get a decent job, and rebuild her life. At first she felt like an outsider but after many years she had made quite a few friends who helped her find her way in the world again.}
\\
\\
\texttt{Then one day while walking through the city center of town she passed by a man sitting all alone on the sidewalk, holding onto his dog for comfort and wondering what was going to become of him next. The city had found him homeless, kicked him out of his apartment, and told him to never return again unless he wanted to be arrested. It was at this moment that both their lives changed forever… The man's name was Bruce and his beautiful dog's name was Chloe; they were inseparable from each other in life as well as death!}
\\
\\
\texttt{So there we have it—the story behind the cash!}
\\
\\

\bottomrule
\caption{\rolling{}'s generated story for Premise 5. One of the only stories by \rolling{} that is not almost completely irrelevant to the premise. The main character, who is unnamed, finds a bag of money on her doorstep and is struggling with what to do with it. However, the story doesn't quite follow the second half of the premise: this story never describes how the character actually spends her money, nor does she appear to feel particularly guilty anywhere. Additionally, the last few paragraphs contain an odd change in narration.}

\label{tab:premise5_rolling}
\end{longtable}
\end{small}

\begin{small}
\begin{longtable}{p{0.9\textwidth}}
\toprule
\textbf{\rollingfinetune{}{} Story for Premise 5}\\
\midrule
\texttt{Sarah sat in the overstuffed armchair in the corner, picking at the loose threads of her paisley skirt. The chair was comfy but it was making her fidgety. She frowned and looked around the room, waiting impatiently for something – anything – to happen. She'd been here waiting for hours (it felt like it anyway).}
\\
\\
\texttt{"Well?" she said aloud. Her voice sounded flat, even to her own ears. Sarah looked around the room again, wondering what she should say next.}
\\
\\
\texttt{"Ah hah! You're alive!" The voice startled Sarah so much she jumped a foot off the ground then staggered around in a circle before finally falling back into the chair and staring at nothing with wide eyes.}
\\
\\
\texttt{"Who are you? What's going on?" Sarah asked, afraid to look away from the empty space in front of her as if she might miss something interesting that way.}
\\
\\
\texttt{"You're not happy to see me?" said a woman's voice directly above her. Sarah jerked her head up (and hurt it on something) to see a pair of legs dangle down right over where she had been sitting just seconds before. Something cold and hard poked into her shoulder blade, making her flinch away.}
\\
\\
\texttt{"Oh, right, right. Sorry," said the voice. Bending over to pull up on something, the legs came into view and Sarah saw they were attached to an exceptionally tall woman wearing a long flowing dress of some sort that looked like it was made out of plastic wrap. The woman pulled herself up and over Sarah's head with a grace that belied her height and the plastic skirt fluttered around her as she walked around to stand in front of Sarah.}
\\
\\
\texttt{"Wow," said Sarah after a moment's silence while she tried to find words that would adequately convey how completely mind-boggling this experience was. "Who are you? What is this place?"}
\\
\\
\texttt{The tall woman made a thoughtful noise and then smiled at Sarah, showing off a row of perfect white teeth beneath lips that seemed almost too red for real life. Suddenly got an idea for something to say, however inadequate it might be: "You're very pretty." That was what it was called on those romance novels in the bookstore: 'The shining smile!' or some such nonsense. She wasn't sure if that applied here or not but it seemed like a polite thing to say at any rate. "Thank you," the woman said sweetly. "I'm sure I'm not as pretty as you, though."}
\\
\\
\texttt{Sarah turned bright red, feeling the heat rush up her neck and flood her cheeks. Her gaze dropped to her lap, but she pushed away the urge to hide behind her hands in embarrassment. That's what she'd always done when she got nervous or angry – pressed her hands against her cheeks until they hurt or hid behind them – but she'd promised herself that wouldn't happen anymore. She was an adult; she could be calm and controlled even in an unfamiliar situation like this.}
\\
\\
\texttt{"You're not," said Sarah firmly before she could stop herself, then winced and looked down at the floor again as a silence fell over the room. When it became apparent that the woman was neither going to say anything nor move away any time soon, Sarah continued more quietly: "Sorry."}
\\
\\
\texttt{The woman laughed lightly and reached over to tuck a stray piece of hair back behind Sarah's ear (it wasn't disorderly; it just seemed that way now that someone was looking at it). "There's nothing to apologize for," said the woman warmly. She stood up straight again and then held out a hand towards Sarah. "Come with me. Let's get you out of here."}
\\
\\
\texttt{Sarah hesitated and the woman smiled indulgently. "It's okay. I promise." She waited a few seconds longer as Sarah chewed on her lower lip and tried to decide if she actually believed that or not. After a few moments of silence, the woman rolled her eyes and reached around to tuck her plastic skirt up into her waistband before then reaching down for Sarah's hand. "Stop worrying about it and just come with me," she said, pulling Sarah to her feet without much effort.}
\\
\\
\texttt{A second later, they were standing in front of a door that hadn't been there before and an enormous man was looking through it at them while some other men stood off to the side talking in low voices that sounded like they were arguing over something. They stopped talking when they saw the woman standing in front of them though; one of them cleared his throat awkwardly and stepped forward from the group but froze when he saw their faces staring back at him from through the door he'd just been peering through. He stared for a moment before stammering out, "What are you doing here?" The woman rolled her eyes and looked pointedly at Sarah, who looked back and forth between the two of them in confusion. "You know why we're here," said the woman with a smile. "And she's coming with me." She folded her arms across her chest and tapped one foot impatiently while they squabbled among themselves.}
\\
\\
\texttt{Sarah had never understood how two people could simply stop talking to each other out of nowhere and start again, but it happened right then when one of the men said something that made the others laugh quietly and nod their heads in agreement. The tallest man reached over and pushed on something that turned out to be a door, swinging it open as he did so.}
\\
\\
\texttt{"Very well," he said loftily when he'd finished pushing the door open, giving Sarah his best intimidating look (which failed miserably). "But you're taking her to your house?" he asked with a frown. "I don't think-,"}
\\
\\
\texttt{The woman was already pushing past him, dragging Sarah behind her as she did so. "She's coming with me," she repeated firmly, glaring daggers at him when he tried to intervene once more. "I'll keep her safe, I promise. She's coming with me," the woman said, her voice rising slightly and Sarah felt a thrill of fear run through her as the tension in the room increased.}
\\
\\
\texttt{"She hasn't been tested yet!" yelled one of the other men, grabbing the tall man by the arm and pulling him back. "What if something goes wrong?" he demanded.}
\\
\\
\texttt{The woman scoffed and looked back over her shoulder at Sarah. "I have my ways," she said simply and then pushed open another door with a grunt before ducking inside the hole it left behind in the wall. Sarah grabbed onto the woman's hand tightly as she was pulled after her into a dimly lit hallway.}
\\
\\
\texttt{"I think it's going to be fine," she whispered quietly, squeezing her hand tight as she did so. The door closed behind them with a thump and she jumped when it did but then sighed in relief when nothing else happened. The woman laughed softly when she looked up at Sarah in surprise and then squeezed Sarah's hand back. "Don't worry. Just a bit of showmanship. I'm sorry if I frightened you."}
\\
\\
\texttt{"It's alright," said Sarah, trying to pull the scattered pieces of her wits together. "You didn't really scare me." She felt stupid for saying it immediately, but the woman just laughed again and nodded to what turned out to be a flight of stairs that led upwards and out of sight.}
\\
\\
\texttt{"Come on," she said encouragingly, starting up the stairs with Sarah following close behind. "We've got a lot of work ahead of us."}
\\
\\
\texttt{The woman was surprisingly strong for her size and moved quickly as they climbed the stairs together. She still had her other hand wrapped tightly around Sarah's when they reached the top and then started down yet another hallway with doors on either side, all closed tight even though they were only halfway down it. "What's your name?" Sarah asked, looking down at her hand as they walked. The woman stopped and turned to look at Sarah, who quickly ducked her head again and bit her lip.}
\\
\\
\texttt{"I'm sorry," she said quickly, glancing up through her eyelashes when the woman didn't say anything for a moment. "I-I don't know why I asked that in the first place."}
\\
\\
\texttt{"You can ask me whatever you want," the woman told her with a smile and then yanked open one of the doors as they approached it. It swung open easily and Sarah gasped a little as she looked around inside; it was bigger than any room she'd ever seen before and there were round beds set up all over the place with people in them. Each bed had three people sleeping on it, their heads moving rhythmically as they dreamed, and Sarah stared at them in wonderment until the woman pulled her away.}
\\
\\
\texttt{"Come on," she said softly, leading Sarah deeper into the room before stopping in front of one of the beds. "Let's see how you're doing," she told Sarah, gently running her hands over her body. "I'm sorry if this is uncomfortable."}
\\
\\
\texttt{"What are you doing?" Sarah asked, arching her back and wincing as the woman poked a finger against the bones of her pelvis. "It hurts."}
\\
\\
\texttt{"I know," she said, nodding patiently and rubbing a little on her stomach where Sarah's muscles were all knotted up inside. "That's why we're here. That's why you're awake."}
\\
\\
\texttt{"Uh-huh," said Sarah with a nod, looking up at the ceiling as the woman started to work on something below that made it hurt even more than before. "Alright," she said, trying to sound casual as she lay there, twisting and turning when it was over with the woman muttering quietly to herself the whole time and making Sarah feel like an animal being examined by its master. She wished that she'd never woken up in that lab, wished that she'd just died with all of their other failed experiments in their wake. Everything that had happened since then had just been a jumbled mess of confusion and pain, and she wanted it to stop.}
\\
\\
\texttt{"There," the woman said softly, patting Sarah gently on her stomach. "All finished."}
\\
\\
\texttt{"Huh?" said Sarah, looking up at her with a frown. The woman smiled and nodded to herself before gently pulling back the blanket on one of the beds. "Ohh," Sarah breathed when she saw the two little girls sleeping in it, their cheeks still pressed together as they slept soundly. They were identical looking except for the colour of their hair; one was red and one was black, but otherwise they were perfectly identical twins. There were two bags beside their bed with two silver tubes hooked up to them filled with a greenish liquid that flowed steadily between them as they slept, making a soft chugging sound every few seconds that made it all feel real somehow instead of some surreal fantasy.}
\\
\\
\texttt{The woman looked up at her fondly before grabbing both of their hands in hers and squeezing tight as she stared down at them with a gentle smile on her face. One of the girls stirred and opened her eyes, looking right over at the woman as if she knew exactly where she was before opening them wide and pulling back into her sister's chest, burying her face in her hair.}
\\
\\
\texttt{"Hey," said Sarah with a smirk, her heart beating faster as she watched them. "You're good with kids."}
\\
\\
\texttt{"Thank you," said the woman with a smile. "Now come with me."}
\\
\\
\texttt{"Yes ma'am," said Sarah, giggling to herself as the woman led her over to another bed with another pair of identical twins on it. They were just like Sarah's other two friends in most respects, except for their hair; one girl had blonde hair and one had brown. The woman smiled down at them tenderly before she tapped Sarah on the shoulder and motioned for her to follow again.}
\\
\\
\texttt{"How many more of these are there?" asked Sarah as they walked down another hallway lined with more doors on either side that were all closed up tightly. She didn't know how many total there were but it must've been dozens at least by now, maybe even hundreds if this was only one room out of dozens that stretched long and wide across the facility. The thought made her sick to her stomach and she clutched at the woman's arm, looking up at her for confirmation that it wasn't true. "Please tell me there aren't more."}
\\
\\
\texttt{"Hopefully not for much longer," said the woman, stopping in front of a set of double doors and looking down at Sarah's hands clenched around her arm. She smiled and gently patted them before pushing through the doors, letting Sarah look inside. Her heart felt like it was going to explode right out of her chest when she saw the woman behind a large desk in the corner with a bunch of other people lined up across from him; they were all in black uniforms with stern expressions on their faces as they waited patiently for something to happen, and Sarah couldn't help but wonder if all of these people had woken up like she had.}
\\
\\
\texttt{"You've been waiting long enough," said the man behind his desk when he noticed Sarah staring at him. "Come on in."}
\\
\\
\texttt{"Oh," said Sarah, startled by how confident he was as she let go of the woman and shuffled forward into his office, taking a seat in one of the chairs facing him across his desk as he started talking again.  "I'm glad you could finally join us."}
\\
\bottomrule
\caption{\rollingfinetune{}'s generated story for Premise 5. The story is wholly irrelevant to the premise, besides being about a girl named Sarah. In this story, Sarah is seemingly held or kidnapped by an unknown woman under unclear circumstances. Later on, the woman and her group are seemingly revealed as actually being quite benevolent. }

\label{tab:premise5_rollingft}
\end{longtable}
\end{small}

\FloatBarrier

\newpage

\twocolumn

\section{Example Story Generation Steps}\label{appendix:example_steps}

Here we show a deeper dive into some generation steps from a single story's generation to better illustrate the complete details of our system. 

First we reproduce the initial setup for this story in Table \ref{tab:prompt_example_setup}.

\begin{table}[!htbp]
\small
\begin{tabularx}{\linewidth}{X}
\toprule
\textbf{Initial Setup and Outline For Story Generation Step Example}\\
\midrule
\texttt{Premise: One summer, a group of friends decide to start a business together selling environmentally friendly products door to door.}
\\
\\
\texttt{Setting: The story is set in the town of Ridgefield, Connecticut during the summertime.}
\\
\\
\texttt{Characters:}
\\
\texttt{1. Lila Rosen is a fourteen-year-old girl with curly brown hair and hazel eyes. She is of average height and has a slender build. Lila is a determined and headstrong young woman who is passionate about environmentalism.}
\\
\texttt{2. Katarina Petrova is a fifteen-year-old girl with long straight blond hair and blue eyes. She is tall and has a slim build. Katarina is a kind and gentle soul who loves nature and animals.}
\\
\texttt{3. Oliver Jackson is a fifteen-year-old boy with short brown hair and green eyes. He is of average height and has a muscular build. Oliver is a laid-back and easy-going guy who is always up for a good time.}
\\
\\
\texttt{Outline:}
\\
\texttt{1. Lila, Katarina, and Oliver start a business together selling environmentally friendly products.}
\\
\texttt{2. The business is a success and the trio makes a lot of money.}
\\
\texttt{3. They learn important life lessons along the way and their friendship grows stronger.}
\\
\bottomrule
\end{tabularx}
\caption{The initial setup for the story we use to show prompting examples. }
\label{tab:prompt_example_setup}
\end{table}

\FloatBarrier

After generating the initial setup, our structured attribute detection system infers attributes for Lila, Katarina, and Oliver from the setup. The inferred attribute-value dictionaries are shown in Table \ref{tab:prompt_example_initial_attributes}.

\begin{table}[h]
\small
\begin{tabularx}{\linewidth}{X}
\toprule
\textbf{Initial Inferred Attributes}\\
\midrule
\texttt{\textbf{Lila Rosen}}
\\
\texttt{gender: female}
\\
\texttt{age: fourteen years}
\\
\texttt{build: described as average}
\\
\texttt{friend's name: Katarina Petrova}
\\
\texttt{Katarina Petrova's: best friend}
\\
\texttt{best friend's name: Katarina Petrova}
\\
\\
\texttt{\textbf{Katarina Petrova}}
\\
\texttt{best friend's name: Lila Rosen}
\\
\texttt{Lila Rosen's: best friend}
\\
\texttt{gender: female}
\\
\texttt{age: 15}
\\
\texttt{build: slim}
\\
\texttt{friend's name: Lila Rosen}
\\
\\
\texttt{\textbf{Oliver Jackson}}
\\
\texttt{gender: male}
\\
\bottomrule
\end{tabularx}
\caption{The initial detected character attributes for the story we use to show prompting examples. }
\label{tab:prompt_example_initial_attributes}
\end{table}

\FloatBarrier
We now generate the actual story passage by passage. The first passage uses a slightly different prompt (Table \ref{tab:prompt_example_1}), repeating the initial setup verbatim at the beginning, because there is no previous story text to work with (grey text in Figure \ref{fig:draft}). The system is prompted with \texttt{Chapter 1} at the end of the prompt to encourage it to begin writing a story.

\begin{table}[h]
\small
\begin{tabularx}{\linewidth}{X}
\toprule
\textbf{Prompt for First Story Generation Step}\\
\midrule
\texttt{One summer, a group of friends decide to start a business together selling environmentally friendly products door to door.}
\\
\texttt{The story is set in the town of Ridgefield, Connecticut during the summertime.}
\\
\texttt{Lila Rosen is a fourteen-year-old girl with curly brown hair and hazel eyes. She is of average height and has a slender build. Lila is a determined and headstrong young woman who is passionate about environmentalism.}
\\
\texttt{Katarina Petrova is a fifteen-year-old girl with long straight blond hair and blue eyes. She is tall and has a slim build. Katarina is a kind and gentle soul who loves nature and animals.}
\\
\texttt{Oliver Jackson is a fifteen-year-old boy with short brown hair and green eyes. He is of average height and has a muscular build. Oliver is a laid-back and easy-going guy who is always up for a good time.}
\\
\\
\texttt{The story is written in third person.}
\\
\\
\texttt{Chapter 1 Summary: Lila, Katarina, and Oliver start a business together selling environmentally friendly products.}
\\
\\
\texttt{Full text below:}
\\
\texttt{Chapter 1}
\\
\bottomrule
\end{tabularx}
\caption{The prompt at the very beginning of generation, which differs somewhat from the usual prompt mid-generation.}
\label{tab:prompt_example_1}
\end{table}
\FloatBarrier

Ten continuations are generated; the best one after reranking is shown in Table \ref{tab:prompt_generation_1}.

\begin{table}[h]
\small
\begin{tabularx}{\linewidth}{X}
\toprule
\textbf{Best Continuation for First Story Generation Step}\\
\midrule
\texttt{Lila and Katarina were moving into the apartment right next to Oliver's. They were on a mission to make their place environmentally friendly and had decided that they were going to be the first in the building to go green.}
\\
\\
\texttt{The girls opened up their white plastic moving truck, grabbed their boxes, put them down in front of their new home, and quickly grabbed the correct keys from Oliver.}
\\
\\
\texttt{"Here you go ladies!" Oliver said as he handed them a set of keys, with a smile on his face.}
\\
\\
\texttt{Lila, who was now standing on the doorstep with Katarina, gave him an appreciative look before she unlocked the door and pushed it open.}
\\
\\
\texttt{Once inside, Katarina looked around as she walked through the small apartment that was already furnished for them. It had three rooms: a bathroom and main room with a nice-sized kitchen in between it and her own room where her bed was already made up for her.}
\\
\\

\bottomrule
\end{tabularx}
\caption{Selected story continuation from the previous prompt.}
\label{tab:prompt_generation_1}
\end{table}

\FloatBarrier

New entities are detected, and then attributes are added to the structured dictionaries if detected. Subsequent prompts follow the pattern shown in Figure \ref{fig:draft}; Table \ref{tab:prompt_example_2} shows a prompt from later in the generation.

\begin{table*}[h]
\small
\begin{tabularx}{\linewidth}{X}
\toprule
\textbf{Prompt for Later Story Generation Step}\\
\midrule
\texttt{Relevant Context:}
\\
\texttt{Lila Rosen is a fourteen-year-old girl with curly brown hair and hazel eyes. She is of average height and has a slender build. Lila is a determined and headstrong young woman who is passionate about environmentalism.}
\\
\texttt{Katarina Petrova is a fifteen-year-old girl with long straight blond hair and blue eyes. She is tall and has a slim build. Katarina is a kind and gentle soul who loves nature and animals.}
\\
\texttt{Oliver Jackson is a fifteen-year-old boy with short brown hair and green eyes. He is of average height and has a muscular build. Oliver is a laid-back and easy-going guy who is always up for a good time.}
\\
\\
\texttt{The story is written in third person.}
\\
\\
\texttt{Previous story summary: Lila, Katarina, and Oliver start a business together selling environmentally friendly products.}
\\
\\
\texttt{Events immediately prior to the upcoming passage: Lila and Oliver just started their own business painting new things for people to buy. Lila's business partner, Katarina, comes over to help them get ready for selling paintings and other art. Lila is acting strange about the whole thing, but Oliver reassure her it's okay. Katarina starts banging on the door and when Lila opens it, she has a big smile on her face.}
\\
\\
\texttt{In the upcoming passage, the business is a success and the trio makes a lot of money.}
\\
\\
\texttt{Full text below:}
\\
\texttt{"Come on in!" she said.}
\\
\texttt{Katarina walked inside before looking at Oliver and then Lila. "Hey you two! So how's this painting company coming along? Did you guys get a name for it?" Lila nodded her head. "We're gonna call it 'Green Designs.'" Oliver looked at Lila with an impressed expression on his face before turning towards Katarina and said, "Yeah, the idea is that we'll sell the paintings for a price people can afford and we'll also sell eco-friendly goods like solar panels, wind turbines, rainwater collectors, and stuff." Katarina nodded her head and said, "That's really cool." She looked around the room before saying to them both "So have you guys decided who's going to be in charge of what?" Lila shook her head as she said, "Not yet. I was gonna ask you if you wanted to do the paperwork because I hate doing paperwork and Oliver is better at working with people than me anyway so he'll probably be the one dealing with customers."}
\\
\texttt{Oliver smiled happily at Lila as she spoke. He coughed slightly before saying, "That would be great! Thanks!"}
\\
\bottomrule
\end{tabularx}
\caption{Example of a prompt in the middle of story generation.}
\label{tab:prompt_example_2}
\end{table*}

\FloatBarrier

Once again, ten continuations are generated, with the best one after reranking shown in Table \ref{tab:prompt_generation_2}.

\begin{table}[h]
\small
\begin{tabularx}{\linewidth}{X}
\toprule
\textbf{Best Continuation for Later Story Generation Step}\\
\midrule
\texttt{He turned to Lila and said, "I've made a list of people that would be good to talk to about this." He tossed a piece of paper at her, which she caught.}
\\
\\
\texttt{"I've made another list for you about potential customers." Katarina said. "I did my best to list people who have the money but also seem like they'd be really interested in this kind of thing." Oliver smiled again at her as he said, "That's great! Thanks!"}
\\
\\
\texttt{Lila looked down at the piece of paper in her hands and said, "Hey! This is perfect!" She quickly began opening up a new Microsoft Word file on her computer and typed out three things; 'Green Designs,' 'Art,' and 'Available.' Then she pressed Save before turning back to Oliver and Katarina with an excited look on her face.}
\\
\\
\texttt{"Okay! So now everyone knows what they're doing!" Lila said happily as she stood up from where she was sitting on the floor surrounded by papers. Oliver nodded his head as he got up from his spot on the floor as well while Katarina remained sitting on the floor where she was.}
\\
\bottomrule
\end{tabularx}
\caption{Selected continuation for the previous prompt.}
\label{tab:prompt_generation_2}
\end{table}

\FloatBarrier

In the passage shown in Table \ref{tab:prompt_generation_2} we actually detect additional entities such as Microsoft Word, but these are correctly marked as non-character entities and thus we do not build an attribute dictionary for them. Finally, attributes are updated for existing characters as needed. 

One final example prompt from the very end of generation is shown in Table \ref{tab:prompt_example_3}.

\begin{table*}[h]
\small
\begin{tabularx}{\linewidth}{X}
\toprule
\textbf{Prompt for Second Later Story Generation Step}\\
\midrule
\texttt{Relevant Context:}
\\
\texttt{Lila Rosen is a fourteen-year-old girl with curly brown hair and hazel eyes. She is of average height and has a slender build. Lila is a determined and headstrong young woman who is passionate about environmentalism.}
\\
\texttt{Katarina Petrova is a fifteen-year-old girl with long straight blond hair and blue eyes. She is tall and has a slim build. Katarina is a kind and gentle soul who loves nature and animals.}
\\
\texttt{Oliver Jackson is a fifteen-year-old boy with short brown hair and green eyes. He is of average height and has a muscular build. Oliver is a laid-back and easy-going guy who is always up for a good time.}
\\
\\
\texttt{The story is written in third person.}
\\
\\
\texttt{Previous story summary: Lila, Katarina, and Oliver start a business together selling environmentally friendly products. The business is a success and the trio makes a lot of money.}
\\
\\
\texttt{Events immediately prior to the upcoming passage: Lila and Oliver showed Katarina how to use different stickers to create designs on canvasses, and then Lila finished her part by painting flowers on to another canvas. Oliver was done and showed his notebook to Lila and Katarina, which they were impressed by.}
\\
\\
\texttt{In the upcoming passage, they learn important life lessons along the way and their friendship grows stronger. This is the end of the story.}
\\
\\
\texttt{Full text below:}
\\
\texttt{The three friends smiled at each others' artwork before getting up off their canvasses and going over to where their bags were sitting. They grabbed their things before leaving the room.}
\\
\texttt{The trio walked out of the arts room into the hallway which was filled with students rushing here and there for dinner. Lila, Katarina, and Oliver stopped and looked at each other. Now that they were finished with their portraits, they needed to find some people to sell them to. However, many students were now in the hallway which would make it hard for them to get some rich people to buy what they made!}
\\
\texttt{Lila looked around the room and noticed that there was an empty table not too far from the arts room. She headed over there with Katarina and Oliver following her. They set down their bags on the table before sitting down in a booth seat facing each other. The trio checked out what some students were eating and began talking about how hungry they were as well.}
\\
\texttt{After a little while, a couple of students walked over to where Lila, Katarina, and Oliver's table was and sat across from them on the opposite side of the table. The student who stood up first had long curly brown hair with hazel eyes.}
\\
\bottomrule
\end{tabularx}
\caption{Example of a prompt at the very end of story generation. The previous story summary now contains the concatenation of all previous parts of the outline. When describing the outline of the upcoming passage, we include \texttt{This is the end of the story} since it is the last part of the outline.}
\label{tab:prompt_example_3}
\end{table*}

\FloatBarrier

\newpage

\section{Example Data for Editing System Evaluation}\label{appendix:consistency_example}

We provide an example from our evaluation set to illustrate what the data looks like, including both the setups $s$ and $s'$ (Tables \ref{tab:consistency_example_initial_setup} and \ref{tab:consistency_example_altered_setup}) and corresponding stories $t$ and $t'$ (Tables \ref{tab:consistency_example_initial_story} and \ref{tab:consistency_example_altered_story}). In this example, Beth is Julie's mother in $s$ and $t$ but is Julie's friend in $s'$ and $t'$. 

\begin{table}[h]
\small
\begin{tabularx}{\linewidth}{X}
\toprule
\textbf{Initial Setup $s$}\\
\midrule
\texttt{A young woman finds out that she is pregnant and decides to abort the pregnancy even though it goes against her religious beliefs.}
\\
\\
\texttt{The story is set in a small town in the Midwest.}
\\
\\
\texttt{Julie Christensen is a young woman in her early twenties. She is pretty and blonde, and is the daughter of a prominent family in the town. Julie is a student at the local college, and is active in her church.}
\\
\\
\texttt{Tommy Foster is a young man in his early twenties. He is the son of a poor family in the town, and has been involved in a lot of trouble in his life. Tommy is Julie's boyfriend, and the father of her unborn child.}
\\
\\
\texttt{Beth Christensen is Julie's mother. She is a well-respected member of the community, and is very involved in her church. Beth is not happy about her daughter's situation, and is strongly opposed to the idea of abortion.}
\\
\bottomrule
\end{tabularx}
\caption{The initial setup $s$.}
\label{tab:consistency_example_initial_setup}
\end{table}

\begin{table}[h]
\small
\begin{tabularx}{\linewidth}{X}
\toprule
\textbf{Altered Setup $s'$}\\
\midrule
\texttt{A young woman finds out that she is pregnant and decides to abort the pregnancy even though it goes against her religious beliefs.}
\\
\\
\texttt{The story is set in a small town in the Midwest.}
\\
\\
\texttt{Julie Christensen is a young woman in her early twenties. She is pretty and blonde, and is the daughter of a prominent family in the town. Julie is a student at the local college, and is active in her church.}
\\
\\
\texttt{Tommy Foster is a young man in his early twenties. He is the son of a poor family in the town, and has been involved in a lot of trouble in his life. Tommy is Julie's boyfriend, and the father of her unborn child.}
\\
\\
\texttt{Beth Christensen is Julie's best friend since childhood and her confidante in this matter.}
\\
\bottomrule
\end{tabularx}
\caption{The altered setup $s'$, in which Beth Christensen's description has been re-sampled, so that she is now Julie's friend instead of her mother. }
\label{tab:consistency_example_altered_setup}
\end{table}

\begin{table}[h]
\small
\begin{tabularx}{\linewidth}{X}
\toprule
\textbf{Initial Story $t$}\\
\midrule
\texttt{Beth Christensen walked into the living room, where her daughter Julie was sitting with her boyfriend Tommy. She could tell that the two were happy and carefree, and she instantly knew that something was up.}
\\
\\
\texttt{"Julie," Beth said sternly. "What's going on?"}
\\
\\
\texttt{"What do you mean?" Julie asked nervously.}
\\
\\
\texttt{"You look too happy for something not to be going on," Beth replied, glaring at Tommy. "Is this about what I think it is?"}
\\
\\
\texttt{Julie turned to Tommy for help, but he remained silent and looked away from her. "I'm pregnant," she said in a low voice after several moments of silence. She started to tear up as she finished speaking. "I'm sorry."}
\\
\\
\texttt{Beth was shocked by what she had just heard. She went to her daughter and pulled her into a hug while scolding Tommy at the same time. "Tommy Foster! What did you do? When did this happen? Why didn't you stop this from happening? You're the one who is supposed to be responsible! Now I have to deal with you and your father's mistakes."}
\\
\bottomrule
\end{tabularx}
\caption{The story $t$ corresponding to the initial setup $s$; Beth is introduced as Julie's mother.}
\label{tab:consistency_example_initial_story}
\end{table}

\begin{table}[h]
\small
\begin{tabularx}{\linewidth}{X}
\toprule
\textbf{Altered Story $t'$}\\
\midrule
\texttt{Beth Christensen sat down in a chair across from Julie and took her hand. "You have to listen to me," she said. "You're having the baby."}
\\
\\
\texttt{"Beth, no, no, no! You don't understand," Julie protested. "It's impossible."}
\\
\\
\texttt{"Why?" Beth replied. "Is it because you're still a virgin?"}
\\
\\
\texttt{Julie looked at her friend with tears in her eyes and nodded. "How can I be pregnant? How can that happen?"}
\\
\\
\texttt{"Because you had sex," Beth replied. "You know how it happens, right? The same way every time." She paused for a moment, and then asked softly, "Does he know? Has he spoken to you yet?" She then held up a hand to stop Julie from responding before continuing. "Listen, my advice is that you talk to him about this. Tell him what's going on and see what he says about it."}
\\
\bottomrule
\end{tabularx}
\caption{The story $t'$ corresponding to the altered setup $s'$; Beth is introduced as Julie's friend.}
\label{tab:consistency_example_altered_story}
\end{table}

\FloatBarrier

\clearpage

\section{Longer Story Example}\label{appendix:longer_story}

We adapted \ours{} to generate a much longer story to demonstrate our framework's capabilities without being constrained by evaluation limitations. Thus we no longer fix the length of either the outline or the story expansions of each outline point as described in the Method Instantiation in Section \ref{sec:evaluation}. Rather, we allow the system to generate as long an outline as needed for the premise. Story generation proceeds to each next point of the outline once our reranking system deems the best next continuation to be worse by a sufficient threshold of log-probability (of being coherent and relevant to the premise) compared to the previous continuation. In our view, these design changes should be preferred if one's goal is solely to generate high-quality stories, rather than to generate stories of a particular length for fair evaluation.

To generate a longer story, we additionally modify our initial high-level outline to include two levels of hierarchy, doing so by prompting the GPT3-Instruct-175B to list minor plot points for each point of the first outline. We then generate a story passage for each lower-level outline point. After moving on to the next high-level outline point, previous groups of low-level outline points are collapsed into their corresponding high-level outline point when they appear in the \draft{} module's prompt.

We note that recursively adding more levels of hierarchy in the same way could enable our system to generate exponentially longer stories as desired, without substantially increasing the prompt length. In fact, the example generated here is roughly 7500 words, but we still limit the generator's context length to 1024 tokens, as in the main text experiments. 

For this longer generation example, we generate a single premise and then generate a single story from that premise without cherry-picking (Tables \ref{tab:longerstorypremise}, \ref{tab:longerstory_ourssetup}, \ref{tab:longerstory_ours}). Perhaps partially due to the removal of length limitations on individual components and partially by chance, the generated story follows the premise and generated outline quite closely, and is able to maintain a highly coherent overarching plot with a clear beginning, middle, and end. There remain a couple of odd details and/or inconsistencies, but they do not significantly detract from the overall understandability of the story. Overall, in our view, this long example is higher quality than any of the shorter examples from Appendix \ref{appendix:examples}, and is a better representation of the top end of generations that our system is capable of. Nevertheless, there is still a gap compared to what one might expect from an experienced human writer. See the caption at the bottom of the story (Table \ref{tab:longerstory_ours}) for a summary containing additional details and analysis.

\onecolumn

\begin{table}[!htbp]
\small
\begin{tabularx}{\linewidth}{X}
\toprule
\textbf{Longer Story Premise}\\
\midrule
\texttt{When the world ended, the few survivors left huddled together in a small town in the middle of nowhere. They thought they were safe until the mutants showed up. Now, they have to do whatever it takes to survive.}\\
\bottomrule
\caption{Premise for longer generation example. A typical apocalyptic scenario, but not completely mundane.}
\label{tab:longerstorypremise}
\end{tabularx}
\end{table}

\begin{table}[!htbp]
\small
\begin{tabularx}{\linewidth}{X}
\toprule
\textbf{Initial Setup and Outline for \ours{}, Longer Story Example}\\
\midrule
\texttt{Premise: When the world ended, the few survivors left huddled together in a small town in the middle of nowhere. They thought they were safe until the mutants showed up. Now, they have to do whatever it takes to survive.}
\\
\\
\texttt{Setting: The story is set in a small town that has been devastated by a global apocalypse.}
\\
\\
\texttt{Characters:}
\\
\texttt{1. Mark Bradley is a tall and muscular man with short brown hair and blue eyes. He is the leader of the group of survivors and is determined to keep them alive.}
\\
\texttt{2. Sophie Jameson is a young woman with long blonde hair and blue eyes. She is the group's medic and does her best to keep everyone healthy.}
\\
\texttt{3. Henry Grey is an older man with graying hair and hazel eyes. He is the group's engineer and is responsible for keeping the power and water running in the town.}
\\
\\
\texttt{Outline:}
\\
\texttt{1. The world ends and the few survivors are left in a small town.}
\\
\texttt{\quad a. The sun sets for the last time and the skies turn dark.}
\\
\texttt{\quad b. A bright light shines in the distance and gets closer and closer until it's right above the town.}
\\
\texttt{\quad c. A loud explosion rocks the ground and a shockwave knocks everyone to the ground.}
\\
\texttt{\quad d. When the Survivors get up, they see that the town has been destroyed and that there are bodies everywhere}
\\
\texttt{2. The town is attacked by mutants and the survivors have to fight for their lives.}
\\
\texttt{\quad a. The mutant army arrives in the town and starts massacring the citizens.}
\\
\texttt{\quad b. The surviving citizens take up arms and start fighting back against the invaders.}
\\
\texttt{\quad c. The battle rages on for days with neither side giving any ground.}
\\
\texttt{\quad d. Finally, the surviving citizens manage to drive the mutant army out of the town and back into the wilderness from whence they came}
\\
\texttt{3. The survivors must find a way to escape the town and make it to safety.}
\\
\texttt{\quad a. The survivors gather supplies and make a plan to escape the town.}
\\
\texttt{\quad b. They battle their way through the mutant-infested streets.}
\\
\texttt{\quad c. They reach the outskirts of the town and are met by a group of armed men.}
\\
\texttt{\quad d. The men take them to safety and help them rebuild their lives.}
\\
\bottomrule
\caption{\ours{}'s initial setup and outline for the longer story example, with the letters under each number of the outline indicating the two-level hierarchy for the outline. The characters and outline make perfect sense for the given premise, although there is some repetitiveness in the lettered sub-points.}
\label{tab:longerstory_ourssetup}
\end{tabularx}
\end{table}

\begin{small}
\begin{longtable}{p{0.9\textwidth}}
\toprule
\textbf{\ours{} Longer Story Example}\\
\midrule
\texttt{"What do you mean, no more power?" Mark shouted at the man standing next to him.}
\\
\\
\texttt{"I'm sorry, sir, it just shut down," the man said.}
\\
\\
\texttt{Mark took a deep breath and tried to calm himself down. "How long will it take to get up and running again?" he asked.}
\\
\\
\texttt{"I don't know, sir," was the reply, "Maybe an hour or two if we can get all of our people working at it."}
\\
\\
\texttt{Normally that wouldn't be a problem but with the power out across most of the state there were few people left to help them restore power. There were only five hundred workers spread out across the entire town and they couldn't spare more than half of them until the others realized how dire their situation truly was. They needed those workers to work on restoring water flow and keeping their food supply from spoiling as well but there was nothing he could do about it either way. He'd have to page through his list of remaining workers and see who else he could pull off task without having panic take over in their small little town. He wasn't sure how many people would respond but he would worry about that later. The most pressing issue was getting power back online so that his group would be able to defend themselves.}
\\
\\
\texttt{Mark turned to Sophie. "Can you believe this, Sophie? We're living in a time where we have to worry about our power systems. That's what they're there for, to keep the world running so that people can live!"}
\\
\\
\texttt{Sophie smiled at her friend, trying to calm him down before she agreed with him. "I know this is hard for us, but think of all the other cities out there." She pointed out the window at the townhouse across the street from their office building. "How are any of their systems keeping them alive? They're running on generators and battery packs."}
\\
\\
\texttt{"I don't care," Mark replied. "This is our home and we should be taken care of by now." Mark looked around his little office which was crowded with people staring at monitors and talking into radios. "Do you think we'll ever go back?" he asked as he looked out of his window again at what had been a thriving metropolis for hundreds of years before it became a ghost town overnight..}
\\
\\
\texttt{"We have to," Sophie replied. She looked down at her feet and then up into his eyes again. "I'm not ready to die, Mark. I've lost my family and my friends during the war and now I'm losing all the people that I came here to be with."}
\\
\\
\texttt{Mark looked at her for a moment before he hugged her. "I know, Sophie. We'll stick together and we'll get through this."}
\\
\\
\texttt{Sophie smiled as she leaned into his embrace. "Just promise me that you'll never lose hope, Mark."}
\\
\\
\texttt{Jack had heard enough of their conversation from outside the door. He walked in just as Mark pulled away from Sophie and said, "We can't give up hope. There's no other choice now." Jack tried to get away before he was noticed but Mark caught him by the arm and pulled him into a hug.}
\\
\\
\texttt{Jack wrapped his arms around him and squeezed him tight as they stood together in each other's arms on their final day on earth. As darkness settled over the city they all held onto one another as if it would make them invincible from what was out there in the dark with them that night. But that wasn't true and there was nothing anyone could do about it.}
\\
\\
\texttt{*** They were all standing on the roof of the building looking up at the sky when it happened. Mark took Sophie's hand in his and held it as they looked up into that dark void. The last of the light was fading from their view and they could see things moving around in the sky. They were moving so quickly that it was hard to track them, but as they got closer towards them, it became obvious what they were.}
\\
\\
\texttt{Large winged beasts descended from the sky and landed on the buildings around them. A few others flew by overhead, but most remained above them in the air for a time before landing. The creatures had leathery wings attached to a long body with clawed appendages at the ends of its limbs. They also had multiple eyes dotting their faces and long horns growing out of their heads. The things looked almost like giant bats, but they didn't behave like any animal they'd ever seen.}
\\
\\
\texttt{One of the creatures took flight behind one of its flying companions, who appeared to be doing just that: protecting another creature from danger in some form or another that existed at ground level from what seemed like an attack from something below him. The creature behind its comrade grabbed it by the back of its neck, flying with it in tow. The wings of the flying beast started to flap more quickly as it continued to pull its comrade higher into the air with him. As they rose together, Mark could see three smaller creatures attacking the other in the sky. He couldn't tell if they were trying to bring down the larger creature or if they were attempting to free their comrade from his grasp, but he saw one of them come up for a strike. It latched onto one of the feet of the creature holding their friend captive and bit down hard on his muscled leg before being shaken loose by one of his arms and falling back towards the ground below them as Sophie let out a small gasp as what she was seeing registered in her brain.}
\\
\\
\texttt{She pulled her hand from Mark's and started to walk towards one of the creatures standing near them still watching what was happening above them in silence as if he were observing a play unfolding before him at some point in time that was supposed to be exciting. He'd spent so much time learning about these creatures over time, but now he realized that there might never be an opportunity to learn enough about their kind or how they operated around humans without being attacked by them. The winged creature that was holding the other captive finally started to move in the direction of Ben's house, bringing them even closer to them as Mark glanced over at Sophie. She was staring at the ground, her face contorted in horror as he placed his hand on her shoulder and squeezed it slightly in an attempt to comfort her. He wanted to tell her that everything would be okay, but he knew that they all knew that was a lie. Nothing would be okay again unless they could find a way to defeat these creatures and he wasn't sure how they were going to do that.}
\\
\\
\texttt{As the creature approached the building beside him, it seemed as though its wings began to slow down as it came in for a landing on one of the windows on the first floor of his home. The other flying creature let go of his companion's wings and slowly started descending towards the ground with him. They both landed near each other, facing in opposite directions and looking towards his house before one of its hands lifted and pointed directly at Mark's face for a moment before it dropped again. Mark could feel Sophie trying to pull herself away from him so she could walk towards where one of the creatures was pointing but he didn't allow her go until she stopped pulling against him. He knew they couldn't do anything to either of the creatures, but he didn't want Sophie to be in their way if they did decide to attack. He had no idea what they wanted with them, but at this point, he wasn't sure it would matter. They were already trapped inside of the fence and there was nothing they could do against them. He placed his other arm around her, hoping that she would understand that he just wanted to keep her safe.}
\\
\\
\texttt{The brightly lit object appeared in the distance and slowly started getting closer until it was directly over their town. Mark had expected it to be one of the flying things that had attacked them earlier, but it was different from anything he had ever seen or heard of before. It seemed to be some kind of large shimmering ball with a smaller sphere attached to its side. As it flew lower closer towards them, Mark saw that there was some kind of reflective enclosure around the outside of the sphere and appeared as though it was translucent in places. It looked like someone had taken a small piece of glass and glued it onto an orb shaped piece at an angle so that it was pointing downwards towards the ground and connected on both ends with some kind of shiny material that looked like metal sticking out from either side. The size of the object was difficult to determine because it didn't appear to be directly overhead. The balloon-like material was also difficult to see because it seemed as though it was glowing, and the reflected light from the sphere covered up any outside objects that might have helped him pinpoint its size. It felt like it was hovering just above his head, but he knew that it couldn't really be that close. He tried to focus on what direction it was going and decided that it was moving south along their fence line. As soon as he realized this, a bright light shone down onto the ground in front of them. The light grew brighter until he could no longer see or hear anything else around him and had to cover his eyes in order to avoid being blinded.}
\\
\\
\texttt{When Mark opened his eyes again, he felt as if someone were watching him and turned around towards Sophie who was standing next to him with her mouth open staring at something behind him. There were a few large creatures standing behind him, but they were different than the ones they had seen earlier; they looked almost human - like giant humanoids with long dark hair, large dark eyes and pale skin. Their faces were flat and their features were not as well-defined as humans, but they had two arms and legs with joints like humans did. They were dressed in shiny dark purple robes which covered them from their head to their feet, and they had something in their hands that looked like a large flashlight. They were staring into the light with their mouths open, running towards it on all fours as fast as they could.}
\\
\\
\texttt{Mark turned back around to face the object again and then realized that the light was now completely gone. The stars and moon shone brightly in the sky above him, and a warm breeze could be felt blowing across his face. He took a few steps closer to the fence hoping to see better, but he couldn't see anything out of the ordinary up above him besides the night sky. The sound of something falling to the ground behind him brought his attention back down to earth where he saw Sophie's flashlight lying on the ground next to her body which was still lying on its side next to where it had just fallen after being knocked over by another shockwave that shook through their town. Sophie was lying still on the ground, not moving at all. Mark ran up to her and knelt down next to her body. He turned her over onto her back and saw that she was breathing, but just barely. Her eyes were wide open in shock, and he assumed that she was probably in some sort of shock from all of the loud rumbling coming from the sky earlier on.}
\\
\\
\texttt{Sophie's eyes began to close slowly, so Mark gently shook her shoulders, trying to get a response out of her so that he would know whether or not she had been knocked unconscious. The shaking did not have any effect on Sophie at all, but it did cause her head to roll back and forth while lying in his arms. Her eyes soon reopened when they hit against his chest, which made Mark wince a bit since he didn't want Sophie's head bouncing off of his chest again after it had hit against it earlier on during the explosion.}
\\
\\
\texttt{The falling trees had scared him half to death when they fell on top of them while they were running away from the creatures back towards their town a few minutes earlier much further away from where they were now. He quickly stood up and looked around the area to see if there were any creatures left. He couldn't see any of them, so he assumed the worst had happened to them before his eyes started to wander over towards the decimated town. The town was no longer visible through the thick smoke that now covered everything, but he knew that it was definitely gone just from looking at the large gaping hole in its place.}
\\
\\
\texttt{Mark ran over towards where the hole had been to see if anyone was still alive. When he got there, there were a few people lying on the ground who appeared to have survived the attack, but they were all unconscious and several of them appeared to be in pretty bad condition.}
\\
\\
\texttt{As soon as he reached one of them, a woman, he immediately started doing his best in order to try and help her out as much as possible since it appeared as though she was dying. The wound in her stomach was extremely deep and large and looked extremely painful since it was bleeding profusely while her body jerked around with each passing second. She did not seem to be aware of anything going on around her except for what appeared to be a few sounds coming from somewhere far off in the distance. As he tried to put pressure on the wound in order to slow down the blood flow, a man came running out of one of the nearby houses. The man had a knife in one hand and the other hand was covered in blood. He looked extremely upset and confused at the same time as he stumbled around before suddenly collapsing onto his knees next to Mark.}
\\
\\
\texttt{"Are you okay?" asked Mark sincerely, trying to see if he could help him out somehow with whatever was going on inside his head.}
\\
\\
\texttt{The man replied, "I'm fine, but someone else is hurt inside." He then took a large breath before saying, "I need to get them back to our group of survivors as soon as possible."}
\\
\\
\texttt{Mark said, "Who is hurt?" He didn't know what was going on around him anymore since it appeared that everyone but him had been hit by some kind of crazy mutant army that seemingly came out of nowhere. He knew that there were some strange things happening right now, but he couldn't understand why they were taking place. While he tried to figure out why this all happened, he felt like there were too many things happening right now for him to make sense of everything going on around him. The man said, "I'm not sure what happened, but my wife was injured in the attack. We were trying to get back to our group of survivors when three of the mutants attacked us. I think we lost the battle, but I'm not sure about the cost of the war." He looked at his bloody hands again before saying, "I managed to stab one of them with a knife and kill one of them, but I don't know if I killed the third one or not."}
\\
\\
\texttt{Mark looked around him, wondering if there were any more of these monsters out there as he continued to try and patch up Sophie's bleeding leg. He then said, "We need to get her back to town as soon as possible. This cut needs medical attention fast. The three of us can go together in case anymore of these mutants attack us along the way."}
\\
\\
\texttt{The man nodded his head in agreement as he picked up his injured wife off the ground and held her against him before running off into town with Mark right behind him and using his pistol in order to make sure that they had a clear path. After a few moments, they made it back to the town, where they were greeted by the rest of their group of survivors.}
\\
\\
\texttt{Mark told Henry and Sophie to take care of Sophie's leg while he went and gathered the rest of the survivors in order to prepare for an attack from any more of these mutants. After calling for everyone to meet in the center of town, Mark stood on top of a stack of barrels in order to be seen by everyone as he said, "I'm afraid that we might have some enemy soldiers after us now. I found three people who might have been attacked by them on their way here." After pausing for a moment, Mark continued, saying, "But I can't confirm whether or not this is true until we go out there looking for some clues."}
\\
\\
\texttt{After hearing this news, everyone got ready with their weapons while Sophie and Henry ran around making sure that everyone was prepared. The women started checking over children and putting them into groups before making sure that all of the children had clear paths out of town in case they needed to retreat. The men then talked quietly amongst themselves while they were getting ready while they made sure that everyone knew exactly what they needed to do as soon as any fighting began. The mood in the town was tense and very anxious as everyone waited for any sign of more of these mutants.}
\\
\\
\texttt{After about fifteen minutes of waiting, the group heard several loud thuds against one of the walls around town. Mark quickly ran towards the noise and found himself standing face to face with a massive grey mutant with three arms and a vicious grin on its face. He immediately yanked his pistol out of his holster before shooting it multiple times in its chest area, but it didn't even seem to notice or care that it had been shot multiple times. It then swung one of its arms at him while laughing loudly, which caused him to stumble backwards onto his butt while holding onto his bleeding head. The soldier next to him grabbed his own pistol and shot at the mutant's head, but just like before didn't seem to do any damage.}
\\
\\
\texttt{After sitting up in defeat, he tried shooting the beast again in its stomach area but still nothing happened as he watched helplessly as the mutant turned around and started walking back towards town. Mark heard more thuds against some of the other walls surrounding town this time from what sounded like four or five more mutants trying to make their way inside so he called for everyone to start fighting back if necessary. The men quickly grabbed their guns, including Mark himself, and began shooting at the incoming mutants. They tried to kill as many of them as possible before they were forced to take cover behind one of the nearby walls when a grenade went off right next to them. Several of the townspeople were injured during the attack and had to be taken by Sophie and Henry back inside their clinic while several of the other survivors did their best to distract the invaders so they wouldn't find their way past the walls.}
\\
\\
\texttt{The brave soldiers then ran back out into town before taking cover behind different walls from which they started shooting at any incoming mutants before making their way inside again once most of them were killed or back tracking because of their wounds. One soldier was brave enough to go inside alone with a shotgun in his hands where he was able to take down two more mutants before fighting with one more. The fight was getting quite interesting as he watched in shock a mutant that was extremely muscular throw several punches at him while laughing loudly while he tried blocking each one with his gun until he eventually got tired of it and just swung his hammer at it once, which sent it flying backwards and onto its back while slowly getting up but not being able to do anything because it was knocked out cold. The soldier then looked at all the other nearby mutants that were still fighting with the other survivors as he started laughing to himself.}
\\
\\
\texttt{"Well..." He said as he started slowly walking towards one of them, still shooting it multiple times until it fell down again, "looks like you guys are a bit too dumb to listen to me when I'm telling you to back off." He then faces another one of them before shooting it several times again, killing it. The soldier then walks over the second one before shooting it repeatedly in the head with his shotgun before turning around and reloading his gun while looking at the others while they were starting to slowly retreat. "Heh... looks like they heard me now." He said before shooting two more mutants in their heads while they were running away, killing them. After laughing once more and reloading his gun again, he walked backwards towards the walls and turned around as he watched the last few survivors standing together as they watched the same group of mutants walk back into town, eventually disappearing behind some buildings. "Well, that's that. Looks like they're leaving." He said after walking back over to them while the other soldiers were starting to give up and lowering their weapons.}
\\
\\
\texttt{"Good riddance." One of the soldiers said with a sigh after the mutant hoard had left.}
\\
\\
\texttt{"I agree. Let's just hope they don't come back again." The soldier next to him said as he started walking towards one of the buildings nearby, about to head in for a break.}
\\
\\
\texttt{"If they do show up again..." The soldier who was talking before started saying as he walked over to another building, "I'll make sure I won't be caught off-guard like this time." He finished as he walked past it, heading in with nothing but his scythe on his back and an assault rifle in hand while the others followed suit.}
\\
\\
\texttt{----- "It's been three days since the army dealt with them, but they're starting to get restless again." Mark said after he walked up to his friends, who were eating in the mess hall located in one of the buildings near the center of the town.}
\\
\\
\texttt{"Three days is a long time. Should we be worried?" The soldier who was sitting next to Mark said as he looked at him while everyone else slowly put their utensils down and started looking at him as well.}
\\
\\
\texttt{"No. They're just uneasy because they remember losing so many mutants yesterday, but it really doesn't take them all that long to regroup and organize a counterattack." He said with a shrug before taking another bite out of his food, "We'll be all right if we stay on our guard for now." He finished before putting his spoon back into his plate and drinking some water, wiping off his face afterwards.}
\\
\\
\texttt{"What about those two young soldiers? Aren't you worried at all about them?" The medic sitting next to Sophie said as he turned towards her after she asked her question.}
\\
\\
\texttt{"Why would I be? They'll be fine." Jack said with a shrug as he took a drink and then set the cup back down, "Stop worrying so much. They'll be fine."}
\\
\\
\texttt{"So you're saying that we won't have to worry about them? Even if the mutants attack us while they're outside on patrol?" Sophie asked with an eyebrow raised.}
\\
\\
\texttt{"Nah, they'll be fine. They're good kids." He said with a laugh before taking another bite, "I've been training them ever since they arrived here. I've been through more than enough with them to know that they can handle themselves even without me around. Trust me; I know what I'm doing." He finished before slowly starting to eat again.}
\\
\\
\texttt{"Everyone is just worried because of their status as trainees." Mark said from the other side of the table as he picked up his own bowl and began eating again as well. "I'm not worried about them. I'm just pointing out the fact that they are still trainees, and they still have a lot of training to do. It's only been three weeks since they arrived here, and it's just plain stupid to send them out on their own already." She said before taking a bite and filling her spoon back up again.}
\\
\\
\texttt{"I disagree. The moment that we started letting the new trainees go out into the field with us was the moment that we got serious about our plans to make sure that everyone gets some real combat experience so that when the time comes, we'll all be ready for it." He said before taking yet another bite.}
\\
\\
\texttt{The young woman rolled her eyes in response, "That is just so unfair." She said with a scowl.}
\\
\\
\texttt{"It's not unfair at all; in fact, it makes perfect sense." He replied with a grin as he chewed his food slowly like he always did while eating. It was something he had picked up from Sophie back when they first met so many years ago. "They're going to need all of the combat experience they can get if they are going to survive out there." "What if they get caught out there and get swarmed by all of the mutants? They are going to need more than just their training to survive out there, and if that happens then we might as well just quit trying to stay alive because it's not going to work. These people won't be able to do anything on their own; they'll be dead before we even know what happened." She argued, picking up her spoon and eating what was left of her soup in one big spoonful. The soldier frowned at that. Her words put him on edge, but he didn't want to show it. If they were still at the base then he would have agreed with her completely, but since moving out here into this new town, he had managed to lose a lot of his inhibitions about keeping his soldiers safe.}
\\
\\
\texttt{The truth was, ever since he first arrived in this town he had been having serious doubts about the future of humanity. He wasn't sure if humankind would actually be able to rebuild from this point on or not. It didn't look good, but that didn't mean that all hope was lost yet as far as Mark was concerned. It was just far from over, and that was why he wanted to push his soldiers as hard as he possibly could. There wasn't any room for mistakes now, and any of the newer recruits were just going to have to get used to the idea of being able to die within days of joining the group.}
\\
\\
\texttt{"I'm not worried about what might happen." He finally said after a moment of silence. "The new guys are fine. They'll be fine."}
\\
\\
\texttt{The medic frowned at that and continued eating her soup without saying anything else. She'd probably just have to wait until the next day before she could do anything about it herself. Mark and Jack were right; she couldn't help them right now.}
\\
\\
\texttt{The sun was rising on a new day in the town, but no one was happy or cheerful about it or anything like that this time around. No one even really cared. The last few days had been absolute hell for them all. Sophie's voice hadn't been heard in the loudest of the fighting, but she was still one of the heroes of the little fight that was going on around her. She'd done a lot and helped out a lot of people at no small risk to herself. That didn't mean that she'd be able to escape unscathed, though.}
\\
\\
\texttt{She continued eating her soup silently as everyone else went through their daily routines. There were plenty of distractions in town right now, but none of it really did anything to ease the pain in her aching back or legs. After they finished eating they headed down to the basement and joined up with some new recruits who were already there waiting for them.}
\\
\\
\texttt{"Welcome back." Mark greeted them cheerfully, his eyes sparkling with life once more despite everything around him screaming about how horrible things had been for him recently. "Now… we're going over how to survive out there."}
\\
\\
\texttt{Sophie blinked at his words for a moment before quickly returning her attention towards him again so he didn't know that she'd gone silent for even half a second. "What do you mean?" She asked.}
\\
\\
\texttt{"We're getting out of here, Sophie." Mark said and he sounded more serious than he had in a very long time. "We're going to gather up supplies and then we're going to follow the river. It should lead us to Colorado."}
\\
\\
\texttt{"And when we get there…" Jack replied and his voice sounded like beaten metal, "What then?"}
\\
\\
\texttt{Mark smiled at the man. "We'll find a place to settle down and then we'll wait for all of these people to die off." He answered before turning away from Jack and back towards the new recruits who were looking on with hopeful eyes that were filled with fear. "So… let's start with who wants to go first."}
\\
\\
\texttt{"I want to go first!" One of the younger children called out and Sophie couldn't help but feel sorry for him. He reminded her so much of both her brothers that she felt a wave of nostalgia pass over her before she was able to shake it away. This wasn't the time or place for that! The boy, who was old enough to have been dropped off at daycare, continued speaking without noticing the battle raging in Sophie's heart. "I'll go first. I want to be strong like Jack and Mark."}
\\
\\
\texttt{Jack and Mark both smiled at the boy. It was obvious that they were trying to hold back their laughter.}
\\
\\
\texttt{"Who wants to go next?" Mark asked and this time it was a middle aged woman who raised her hand into the air. "You? Okay, let's start with you then."}
\\
\\
\texttt{***}
\\
\\
\texttt{"This is a bad idea," Henry said as he walked next to Mark as they searched for a suitable place for them to escape from the town. "I mean, we're going on a journey that could last weeks - all on foot! There are no guarantees that we won't be attacked by zombies along the way!" That wasn't really what he wanted to say though, but he couldn't bring himself to hurt Sophie with the truth of their situation. She needed hope in order to keep going. Without it, she would just give up and die. "There's no guarantee that we'll survive if we stay here either," Mark replied and they kept walking through the town. They hadn't found anything worthwhile yet, but he hadn't given up searching. "We need to get away from here if we have any chance of living. Sophie's right; we don't know when this will end and we don't even know what we're up against." He didn't want to admit it; not even to himself, but after having faced off against the military yesterday he knew that they weren't strong enough to win this fight.}
\\
\\
\texttt{Mark nearly walked into a wall of air when Henry stopped suddenly next to him, staring at a broken window in the side of the wall in front of them. There was something fluttering inside and it was moving! He wasn't sure what it was though and he couldn't get closer for some reason... Something about the window made it impossible for him to go close enough for a better look.}
\\
\\
\texttt{"Hey!" he called out at Henry, but the man didn't respond like he usually did. Mark moved closer to the window and he saw Henry standing behind him, completely still and not speaking. Was he asleep? Mark called out again, but there was still no response. Then he looked through the window again and realized that there was something moving inside. There was a figure standing near a woman lying on the floor! The figure had what looked like black hair and it was pulling on her hair...}
\\
\\
\texttt{The realization hit him like a cold shower; that must have been one of the undead! Henry was curious about it and then it killed him! Mark raised his rifle and aimed straight into the center of the face of whoever was standing in there, ready to fire at any moment.}
\\
\\
\texttt{"G...Get away from her..." he said through clenched teeth, angry at what had happened here. He could feel Sophie's hand on his arm now, trying to pull him away from the window. Maybe she thought he would shoot whoever it was inside, but he wouldn't do that; not as long as they were looking for supplies for them to take with them later.}
\\
\\
\texttt{The person inside stopped grabbing at her hair when they heard his voice outside, causing Sophie to give up pulling on his arm now that they knew someone was inside. He stepped away from the window and told her to keep an eye on what was happening inside. He walked off to talk with Henry, who was standing with his rifle down at his side.}
\\
\\
\texttt{"What is it?" asked Henry when he saw him approach.}
\\
\\
\texttt{Mark moved around next to him so he could see through the window again, "They might have food in there."}
\\
\\
\texttt{"You don't know that," replied Henry, shaking his head. Mark had a feeling that he wouldn't go into the building after the bad things that had happened here already today; not to mention that Henry was watching for mutants...or worse... somewhere else.}
\\
\\
\texttt{"I'm going in there," said Mark firmly. Henry's eyes widened, but before he could say anything, Mark pushed through the door of the store and walked inside. The smell hit him like a freight train; it smelled almost as bad as the other store. Light seeped through onto the floor from cracks in the ceiling and he could see a few decaying bodies lying on the floor nearby. A moan followed by a sound like shuffling came from behind the front counter. Mark located the source of the sound and found a young man hiding behind the counter. His legs showed that he was tied together, but his arms and hands were free to move. He was struggling to get out from behind the counter, where he had obviously been for a while. He also looked very pale and underweight, as if he had been here for days without any food.}
\\
\\
\texttt{"Get out of there!" Mark called to him, grabbing one of the corpses off the floor by its leg and swinging it into the shelves that lined the wall. A few cans fell onto him, which he grabbed and tossed back over to Sophie before turning his attention back to what was going on inside shop. "I'm not going to hurt you!" he called out to the man, who looked at him with a mixture of confusion and fear.}
\\
\\
\texttt{The young man pulled himself free from behind the counter, standing up on wobbly legs as he faced Mark with an angry glare in his eyes. All sorts of thoughts ran through Mark's head: zombie, runaway slave or prisoner...or mutant? He watched as it stumbled toward him across the floor, but stopped in its tracks when it noticed something else moving nearby. Mark spotted the movement and glanced around to see the woman from earlier. She was struggling to push herself off the floor, but she stopped moving when she noticed the young man.}
\\
\\
\texttt{"What's going on here?" Mark called out to them. He didn't want to hurt either of them, but he was curious about what was going on. He expected a zombie or a runaway slave to attack him, but he didn't expect either of these two people. If they were prisoners or slaves, then who had taken them? And if they were mutants, why were they not attacking him? "Are you alright?" He asked the woman who was now sitting up on her knees.}
\\
\\
\texttt{"I think he bit me!" She said in reply, rubbing her neck with one hand and holding something behind her back with another hand that she wasn't using to support herself. "I smacked him in the head with this can opener."}
\\
\\
\texttt{Mark tried not to gape at her and quickly turned his attention back towards the young man who appeared to be crying against an empty shelf nearby while clutching something close to his chest with both hands and keeping his arms tucked close against his body as he wept loudly into it. "What happened here?" Mark asked him. The young man stopped crying and stared at Mark with a look of hatred in his eyes.}
\\
\\
\texttt{"Can you tell me what is happening?" Mark said more calmly this time, and the young man's gaze softened slightly, but he kept his arms wrapped protectively around his body.}
\\
\\
\texttt{"My name is Lansing," he said in a quiet voice. He was still clutching something close to him. "And that's Sophie." He gestured his head towards her and she waved back at him.}
\\
\\
\texttt{"I'm Mark," Mark replied with a friendly smile. He turned to Sophie who had stood up slowly and was rubbing her hip where she had hit the floor earlier. "Are you okay?" He asked her as he watched her with concern. The young woman nodded before looking over at the shelf the other boy was hiding behind with wide eyes before drawing a loud breath through her teeth as if in fear. She ran away from the store, but stopped on reaching the door when she heard something running towards them. She looked back at them once more with frightened eyes before running out into the street where another group of mutants were chasing after a small group of survivors who were fighting them off as they ran along behind them at an alleyway across from them. "We have to help them," Mark said as he moved towards Lansing who stood frozen in place, looking around with wide eyes. "Come on." Mark took Lansing's hand, who flinched at his touch. He looked at him in surprise before he let him lead him out into the streets with Sophie following behind.}
\\
\\
\texttt{Mark wasn't sure how many mutants there were, but it was too many. There was at least half a dozen of them and three survivors that were trying to fend them off. The mutants themselves were different from earlier and much larger than before.}
\\
\\
\texttt{"Let's go this way," Sophie said as she pulled on Mark's arm. He glanced over at her for a moment before nodding and turning back towards Lansing who was staring at the battle in front of them with wide eyes and an open mouth. He ran after Sophie as she moved through the crowd of people in front of them and they turned left down another alleyway just after passing two men who were throwing grenades at one another while fighting off mutants that came running towards them along the street they had been standing on a minute ago. Mark and the others ran until they reached a chain link fence that blocked their path. He was about to ask Sophie what they should do when there was a large explosion behind them. The ground shook and the fence fell to the ground in front of them, leaving open a wide path for them to walk down. "Follow me," Sophie said as she began running down the path.}
\\
\\
\texttt{They were met by several men along the way. Some were armed and fighting off mutants, while others were looting stores that had been broken into by other survivors who had made it this far before they were devoured by the mutants or killed while trying to escape. They saw no signs of any other survivors farther down the street, but Mark knew that didn't mean anything. It only meant that he hadn't seen them yet, because there was a good chance there would be more living people than dead ones right now. He had no idea how many people lived in the city, but he was sure there had to be at least a few thousand.}
\\
\\
\texttt{"Where did you come from?" one of the men asked. He had a large rifle slung on his back and was pointing a handgun at them.}
\\
\\
\texttt{"We were in that building," Sophie replied, pointing in the direction of the store behind her, "but we heard you all coming and ran out the back door."}
\\
\\
\texttt{"Are you alone?" the man asked. He had a thick beard and was wearing dirty clothes with a blood stain on his shirt.}
\\
\\
\texttt{"No," Mark replied. "We have some friends hidden behind us." The man nodded and waved at someone behind Mark. Soon two more armed men joined them, both with rifles slung over their shoulders as well as pistols strapped to their belts.}
\\
\\
\texttt{"How many of you are there?" the first man asked. Mark started to count them all with his fingers, but just shrugged instead when he realized they were all strangers to him that he'd only just met today. "This is Mark," Sophie said, pointing at him. "And this is Lansing and me, Sophie." The men looked around at each other quickly, before the first one nodded.}
\\
\\
\texttt{"Alright, follow us. I'm sure you'll be happy to know that you're safe now. We'll take you somewhere safe and then we can help you with anything else you might need." The man turned around and started walking back down the street.}
\\
\\
\texttt{"Wait," Mark called out as he started walking after him. "What about my friends? You said that there might be others?" One of the other men with an M16 slung over his shoulder pointed at their group and shook his head.}
\\
\\
\texttt{"You can bring them along too if they want to come," he replied before he turned around as well and followed the man in front of them down the street again. Mark looked back at his friends and shrugged before turning around to walk after them again.}
\\
\\
\texttt{Mark's legs were hurting after just ten minutes of walking down the streets of Atlanta, but he tried not to show it so that the other men didn't think that they were slowing him down too much or anything like that. Instead, he just gritted his teeth and forced himself to keep up with them.}
\\
\\
\texttt{Finally, they reached a building that looked like it had been a hotel at one point. The men led them inside and down a long hallway before they came to a stop in front of a door. The man in front of them pulled a key out of his pocket and inserted it into the lock. He turned it and then pushed the door open.}
\\
\\
\texttt{"This is your room," he said as he gestured for them to go inside. "There are two beds and a bathroom. If you need anything, just let us know."}
\\
\\
\texttt{"Thank you," Mark said as he walked into the room. He turned around and watched as the men all walked back down the hallway before he closed the door and locked it. He leaned against it for a moment and then slid down to the floor.}
\\
\\
\texttt{"We're safe," he said, not really sure if he believed it himself. Sophie walked over to him and sat down next to him.}
\\
\\
\texttt{"We're safe," she repeated. "But for how long?"}
\\
\\
\texttt{Mark thought about it for a moment and then shook his head. "I don't know," he replied. "But as long as we're together, we'll be alright."}
\\
\\
\texttt{Sophie nodded and leaned her head against his shoulder. They sat there in silence for a moment before Mark spoke again.}
\\
\\
\texttt{"I love you, you know," he said. Sophie looked up at him and smiled.}
\\
\\
\texttt{"I love you too," she replied. They kissed each other gently before they got up and walked over to the beds. They laid down and quickly fell asleep in each other's arms, knowing that they were finally safe.}
\\
\bottomrule
\caption{An example story by \ours{} adapted for generating much longer stories. The story follows Mark and Sophie as they struggle to survive in the post-apocalyptic world and fend off mutant creatures, before finally reaching safe haven in the end. In general, the story quite high quality, following the premise well with a clear beginning, middle, and end. There are clear transitions between different plot points as specified by the outline (when the power goes down, when mutants invade, when they are about to leave the town for safe haven, etc), and local consistency is well maintained. Many of the larger problems observed in the shorter story generations from Appendix \ref{appendix:examples} do not appear here.
Nevertheless, 
compared to human stories, there remain several avenues for possible improvement. 
Some characters have inconsistent identities or personalities over the story (e.g., Sophie is the name of a main character but later becomes a stranger in a store, and then becomes a main character again in the last part of the story).
The main characters' motivations are vague, and drift somewhat over the course of the story.
The story setting also drifts: for example, at some point there are flying creatures fighting with spheres, but they are completely gone afterwards. Lastly, the strength of the mutants is also inconsistent: at first they seem nigh-invincible, but later on they seem killable with normal measures.}

\label{tab:longerstory_ours}
\end{longtable}
\end{small}

\end{document}